\documentclass[10pt,journal,compsoc]{IEEEtran}
\usepackage{comment}
\usepackage{amsmath,amssymb} 
\usepackage{amsmath}
\usepackage{amssymb}
\usepackage[bookmarks=false]{hyperref}
\usepackage{pifont}
\usepackage{tabularx}
\usepackage{tabu,multirow}
\usepackage{booktabs}
\usepackage{color}
\newcommand{\carla}{Carla}
\newcommand{\cityscapes}{Cityscapes}

\newcommand{\rev}[1]{\color{black} #1 \color{black}}

\newcommand{\domainA}{$\mathcal{A}$}
\newcommand{\domainB}{$\mathcal{B}$}
\newcommand{\atdt}{AT/DT}
\newcommand{\taskone}{$\mathcal{T}_1$}
\newcommand{\tasktwo}{$\mathcal{T}_2$}
\newcommand{\taskaux}{$\mathcal{T}_{aux}$}
\newcommand{\trnet}{$G_{1\rightarrow2}$}

\newcommand{\losstask}[1]{$\mathcal{L}_{\mathcal{T}_{#1}}$}

\newcommand{\semdep}{$Sem \rightarrow Dep$}
\newcommand{\depsem}{$Dep \rightarrow Sem$}

\usepackage{xspace}
\newcommand*{\eg}{e.g.\@\xspace}
\newcommand*{\ie}{i.e.\@\xspace}

\usepackage[table]{xcolor}
\definecolor{LightCyan}{rgb}{0.88,1,1}
\definecolor{chromeyellow}{rgb}{1.0, 0.65, 0.0}
\definecolor{grannysmithapple}{rgb}{0.66, 0.89, 0.63}
\ifCLASSOPTIONcompsoc
  \usepackage[nocompress]{cite}
\else
  \usepackage{cite}
\fi
\ifCLASSINFOpdf
   \usepackage[pdftex]{graphicx}
\else
\fi
\begin{document}
%
\title{Learning Good Features to Transfer\\Across Tasks and Domains}
%
%
%
%

\author{Pierluigi Zama Ramirez*\thanks{* Equal contribution.}$^1$, Adriano Cardace*$^1$, Luca De Luigi*$^1$,\\Alessio Tonioni$^2$, Samuele Salti$^1$, Luigi Di Stefano$^1$\\
$^1$University of Bologna, Italy\\
$^2$Google Inc. \\
{\tt\small\{pierluigi.zama,adriano.cardace2,luca.deluigi4,samuele.salti,luigi.distefano\}@unibo.it\\
\tt\small alessiot@google.com
}
}
\IEEEtitleabstractindextext{%
\begin{abstract}
Availability of labelled data is the major obstacle to the deployment of deep learning algorithms for computer vision tasks in new domains. 
The fact that many frameworks adopted to solve different tasks share the same architecture suggests that there should be a way of reusing the knowledge learned in a specific setting to solve novel tasks with limited or no additional supervision.
In this work, we first show that such knowledge can be shared across tasks by learning a mapping between task-specific deep features in a given domain. 
Then, we show that this mapping function, implemented by a neural network, is able to generalize to novel unseen domains.
Besides, we propose a set of strategies to constrain the learned feature spaces, to ease learning and increase the generalization capability of the mapping network, thereby considerably improving the final performance of our framework.
Our proposal obtains compelling results in challenging synthetic-to-real adaptation scenarios by transferring knowledge between monocular depth estimation and semantic segmentation tasks.
%



\end{abstract}

\begin{IEEEkeywords}
Domain Adaptation, Task Transfer, Semantic Segmentation, Depth estimation\end{IEEEkeywords}}

\maketitle

\IEEEdisplaynontitleabstractindextext

%
\IEEEpeerreviewmaketitle

\IEEEraisesectionheading{\section{Introduction}\label{sec:introduction}}

%
%
%
%

\IEEEPARstart{D}{eep} learning has revolutionized computer vision by providing an effective solution to address a wide range of tasks (\eg, classification, depth estimation, semantic segmentation, etc.).
The rise of a common framework has allowed incredible leaps forward for the whole research community thanks to the ability to reuse architectural and algorithmic improvements discovered to solve one task across many others. 
However, the real knowledge of a neural network is stored inside its trained parameters and we still have no simple way of sharing this knowledge across different tasks and domains (\ie, datasets). 
As such, the first step for every practitioner faced with a new problem or domain deals with  acquisition and labeling of a new training set, an extremely tedious, expensive and time consuming operation.
We argue that sharing the knowledge acquired by a neural network to solve a specific task in a specific domain across other tasks and domains could be a more straightforward and cost-effective way to tackle them.

Indeed, this is demonstrated by the widespread use and success of \emph{transfer learning}. Transfer learning concerns solving new tasks by initializing a network with pre-trained weights, thereby providing a basic approach to knowledge reuse. 
However, it still requires a new annotated dataset to fine tune the pretrained network on the the task at hand. 
A few works focused on the related \emph{task transfer} (TT) problem \cite{zamir2018taskonomy, zamir2020robust}, \ie, on exploiting supervised data to tackle multiple tasks in a single domain more effectively by leveraging on the relationships between the learned representations. As unlabeled domains are not considered in TT problem formulations, the proposed methodologies still rely on transfer learning and availability of a small annotated training set in order to address new datasets.
On the other hand, the unsupervised \emph{domain adaptation} literature (DA) \cite{Wang_2018} studies how the need for annotated data can be removed when leveraging on knowledge reuse to solve the same task across different domains, but it does not consider different tasks.

\begin{figure}
    \centering
    \includegraphics[width=\linewidth]{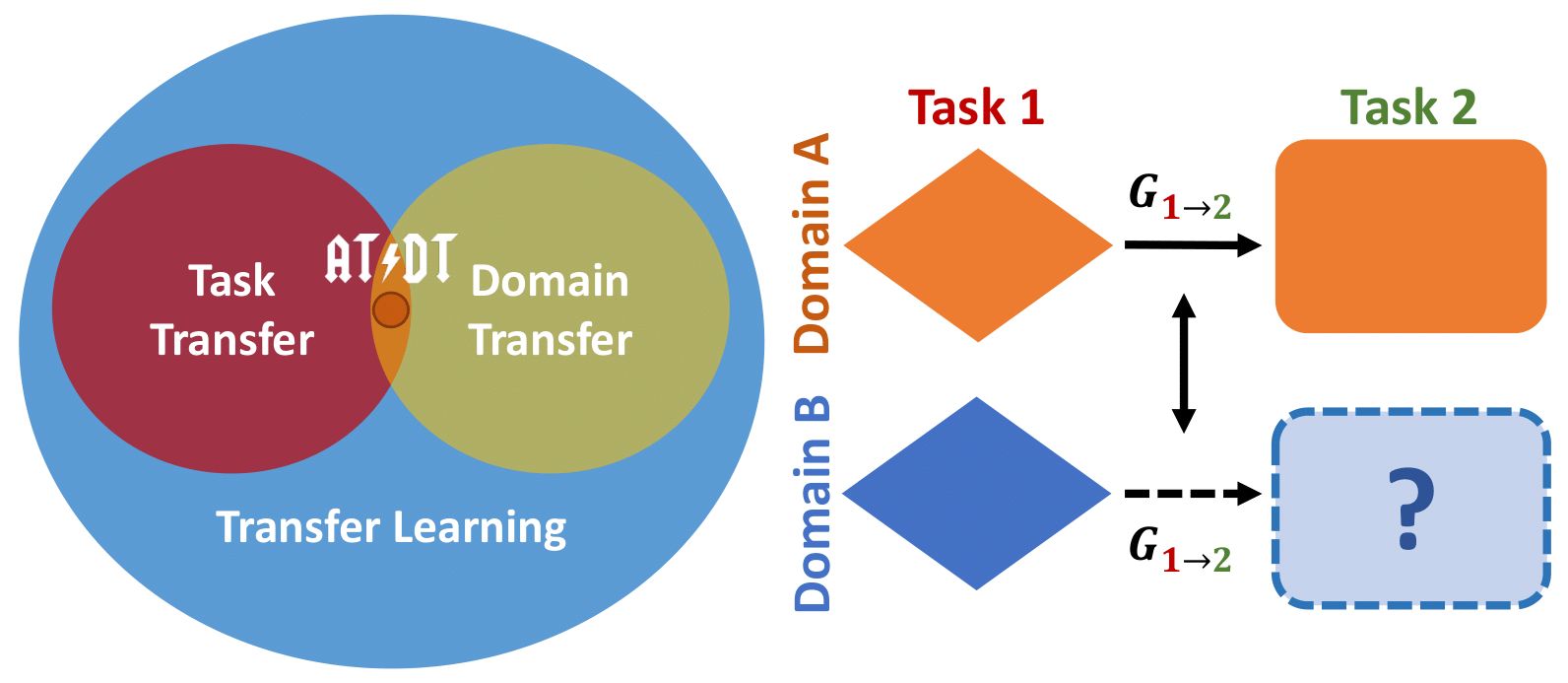}
    \caption{Our framework transfers knowledge across tasks and domains. Given two tasks (1 and 2) and two domains (A and B), with supervision for both tasks in A but only for one task in B, we learn the dependency between the tasks in A and exploit this in B in order to solve task 2 without the need of supervision.}
    \label{fig:teaser_main}
\end{figure}

Differently, we propose to merge DA and TT by explicitly addressing a cross domain and cross task problem where on one source domain (\eg, synthetic data) we have supervision for many tasks, while in another target one (\eg, real data) annotations are available only for a specific task while we wish to solve many.
A schematic representation of our problem formulation with two domains and two tasks is shown in the right part of \autoref{fig:teaser_main}.
Following this schematic representation we will consider a scenario with two domains (a source one and a target one, namely A and B) and two tasks (again a source one and a target one, namely task 1 and 2), but nothing prevents our method to be extended to more.
In domain A we use the available supervision to learn two models for the source and target tasks, while in the target domain B we can do the same for the source task only.
In domain A we use the trained task-specific models to learn a mapping function ($G_{1\rightarrow2}$ in \autoref{fig:teaser_main}) between deep features extracted to solve the source task and those extracted to solve the target task.
This mapping function is then applied in domain B to solve the target task by transforming the features extracted to solve the source task.

The key component of our framework is the mapping function between the two task-specific deep features.
In \cite{ramirez2019learning} we proposed a preliminary formulation  of our framework by modeling the mapping function as a deep convolutional neural network and optimizing its parameters by standard supervised learning in the source domain A.
In this work, we expand and improve upon our preliminary formulation by proposing two features alignment strategies aimed at learning the feature mapping function more effectively.  
Firstly, we align feature representations across domains using a novel norm discrepancy alignment (NDA) loss that constraints the feature space by penalizing features with very different norms in a spatially-aware manner. 
Secondly, we align feature representations across tasks by using them as inputs to solve a common auxiliary task. 
%
This pretext problem acts as a bridge between the source and the target tasks: in fact, if the deep features extracted to solve them independently can be used to address effectively an additional common task, we are pushed to believe that those features present the same semantic content and encode it in a similar manner.

We test the effectiveness of our proposal in a challenging autonomous driving scenario where we try to solve the two related dense prediction tasks of monocular depth estimation and semantic segmentation \cite{ramirez2018exploiting}.
We select edge detection as the auxiliary task since color edges provide oftentimes detailed key information related to both the semantic as well as the depth structure of the scene.
Many edge detectors have been proposed during the years, with recent deep learning based approaches outperforming classical hand-crafted methods even in the most challenging scenarios \cite{soria2020dexined, HED, Wang_2019}.
Interestingly, such deep models present good generalization capabilities, allowing us to use the state-of-the-art approach \cite{soria2020dexined} to generate proxy supervision for the auxiliary task without extra labels.
Thanks to our formulation, we can use a fully supervised and completely synthetic domain (\ie, the Carla simulator~\cite{Dosovitskiy17}) to improve the performance on a partially labeled real domain (\ie, Cityscapes~\cite{Cordts_2016_CVPR}).

The contributions of this paper can be summarized as follow:
\begin{itemize}
    \item We propose for the first time to study a cross domain and cross task problem where supervision for all tasks is available in one domain whilst only for a subset of them in the other. This is done by learning a mapping between deep representations. 

    \item We demonstrate how constraining explicitly deep features across domains with a novel norm discrepancy alignment loss improves the learning of the mapping function.
    
    \item We further show how the learning of the mapping function can be improved by deploying an auxiliary task.

    \item Considering the dense prediction tasks of monocular depth estimation and semantic segmentation, we achieve results close to the practical upper bound when transferring knowledge between a synthetic and a real domain.
\end{itemize}

\section{Related Works}
\subsection{Transfer Learning and Task Transfer}
Collecting training data is often expensive, time-consuming, or even unrealistic in many scenarios. 
Many works have tackled this problem by exploiting the existence of a relationship between the weights of CNNs trained for different tasks \cite{zhuang2019comprehensive}. 
In particular, \cite{yosinski2014transferable} showed that this strategy, referred to as transfer learning, can lead to better results than using random initialization even if applied on quite diverse  tasks. 
Transfer learning has become a common practice, for instance, in object detection, where networks are usually initialized with Imagenet \cite{deng2009imagenet} classification weights \cite{redmon2016you,Ren_2017,He_2017,liu2016ssd}. Additional insights on the transferability of learned representations between different visual tasks were provided in \cite{zamir2018taskonomy}, where the authors present Taskonomy, a computational approach to represent a taxonomy of relationships among visual tasks. 
Along similar lines, \cite{Pal_2019_CVPR} proposed to exploit the correlation between known supervised tasks and novel target tasks, in order to predict the parameters of models deployed to solve the target tasks starting from the parameters of networks trained on the known tasks.
While \cite{zamir2018taskonomy} and \cite{Pal_2019_CVPR} study the correlation between tasks in a given domain and assume either full or no supervision, we explicitly address a multi-domain scenario assuming full supervision in one domain and partial supervision in the target one. 

\subsection{Domain Adaptation}
Domain adaptation techniques aim at reducing the performance drop of a model deployed on a domain different from the one the model was trained on \cite{Wang_2018}. 
Throughout the years, adaptation has been performed at different levels. Early approaches tried to model a shared feature space relying on statistical metrics such as MMD \cite{gong2012geodesic,long15Learning}. Later, some works proposed to align domains by  adversarial training \cite{ganin2015unsupervised,ganin2016domain,tzeng2017adversarial}.
Recently \cite{xu2019larger} noticed that, for classification tasks, aligning feature norms to an arbitrarily large value results in better transferability across domains. Generative adversarial networks \cite{goodfellow2014generative} have also been employed to perform  image-to-image translation between different domains  \cite{Zhu_2017_ICCV,Bousmalis_2017,Isola_2017_CVPR}, and, in particular, to render cheaply labelled synthetic images similar to real images from a target domain. 
However, when dealing with dense tasks such as semantic segmentation, feature-based domain adaptation approaches tend to fail \rev{as deeply discussed in \cite{Tsai_2018}}. Thus, several approaches to address domain adaptation for dense tasks, such as  semantic segmentation  \cite{hong2018conditional,pizzati2020domain,hoffman2016fcns,zhang2017curriculum,ramirez2018exploiting,chang2019all,Tsai_2018,hoffman2018cycada,shrivastava2017learning,Zhang_2018,Sankaranarayanan_2018, pan2020unsupervised, kim2020learning} or depth estimation \cite{Tonioni_2017_ICCV,Zheng_2018_ECCV,Tonioni_2019_CVPR} have been proposed recently. \rev{Among them, SPIGAN \cite{lee2018spigan} uses extra supervision coming from synthetic depth of the source domain to improve the quality of an image-to-image translation network and consequently achieving better adaptation performances.}
Akin to DA methods, we learn from a labeled source domain to perform well on a different target domain. However, unlike the classical DA setting, we assume the existence of an additional task where supervision is available for both domains.

\subsection{Multi-task Learning}
The goal of multi-task learning is to solve many tasks simultaneously. By pursuing this rather than solving the tasks independently, a neural network may use more information to obtain more robust and reliable predictions. 
Many works try to tackle several tasks jointly \cite{Kokkinos_2017,ramirez2018geometry,Cipolla_2018, tosi2020distilled}. For example, \cite{Cipolla_2018} showed that by learning to correctly weigh each task loss,  multi-task learning methods can outperform separate models trained individually. \cite{ramirez2018exploiting, tosi2020distilled} show how learning multiple perception tasks jointly while enforcing geometrical consistency across them can lead to better performances for almost all tasks. Recently, \cite{zamir2020robust} proposes a method to improve the performances of multiple single-task networks by imposing consistency across them during training. 
Finally, Taskonomy \cite{zamir2018taskonomy} investigates the relationship between the deployed tasks to accomplish multi-task learning effectively.
However, multi-task learning approaches usually try to achieve the best balance between tasks in a single-domain scenario. We instead tackle a multi-task and multi-domain problem. Nevertheless, taking inspiration from multi-task learning, we show how jointly learning an auxiliary task while learning the two task networks helps the alignment of features across tasks.

\subsection{Task Transfer and Domain Adaptation}
Most existing approaches address independently either task transfer or domain adaptation. Yet, a few works have proposed to tackle these two problems jointly. \cite{tzeng2015simultaneous} was the first paper  to propose a cross-tasks and cross-domains adaptation approach,  considering as tasks different image classifications problems.
UM-Adapt \cite{kundu2019adapt}, instead, learns a cross-task distillation framework with full supervision on the source domain and deploys such framework on the target domain in a fully unsupervised manner, while minimizing adversarially the discrepancy between the two domains.
Differently, in a preliminary version of this work \cite{ramirez2019learning}, we introduced \atdt{} (Across Tasks and Domains Transfer) and set forth a novel learning framework, where the relationship between a set of tasks is learned on the source domain and it is later deployed to solve a specific task on the target domain without supervision thanks to the availability of ground-truth for all the tasks except the target one. In this work we will expand and improve this methodology.

\begin{figure}[t]
	\centering
	\includegraphics[width=0.45\textwidth]{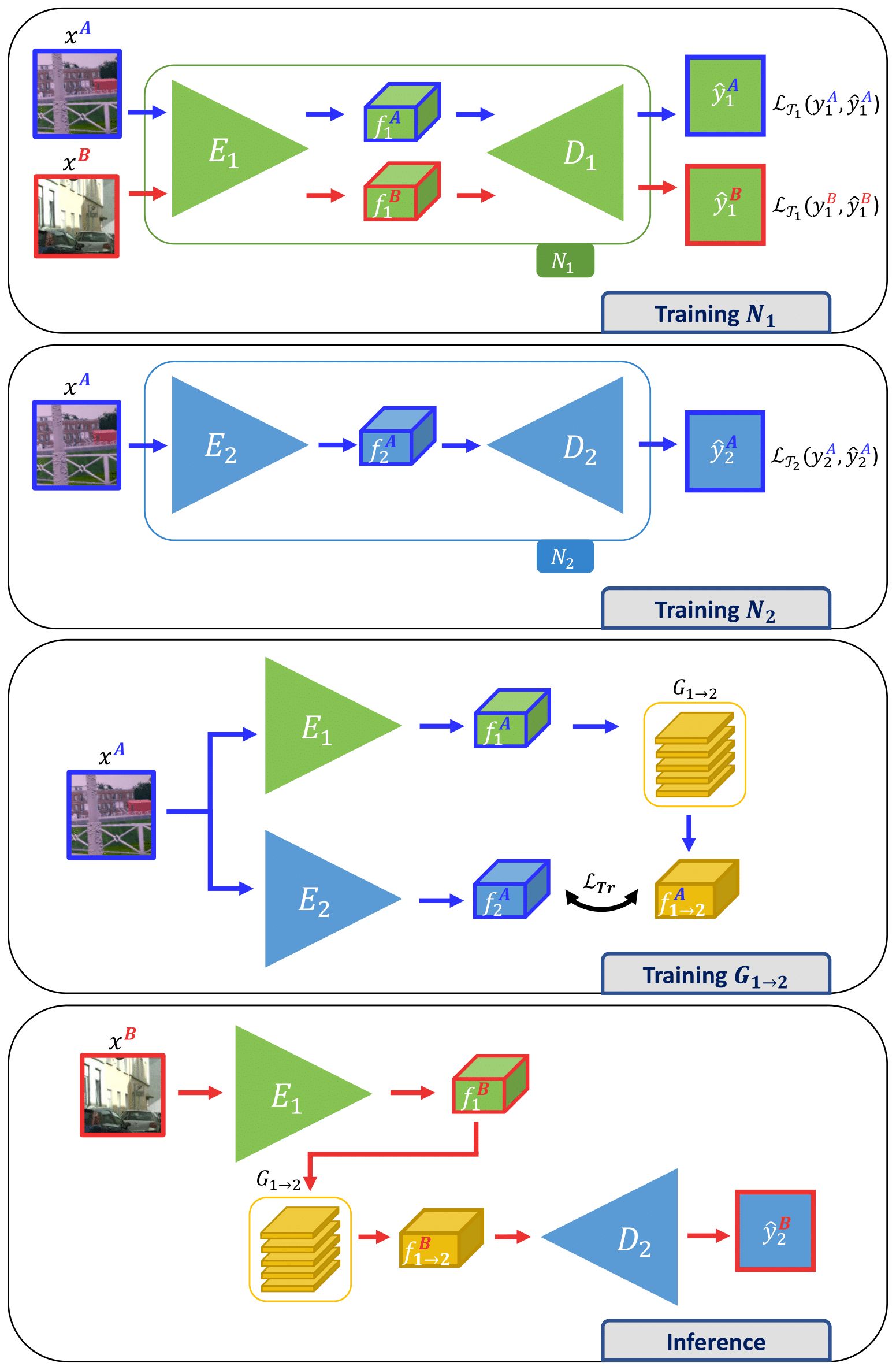}
	\caption{AT/DT framework: here $N_1$ and $N_2$ are trained separately to solve tasks \taskone{} and \tasktwo{}. While $N_2$ is trained only on images from domain \domainA{}, $N_1$ is trained jointly on both domain \domainA{} and domain \domainB{}, to enable the extraction of domain invariant features. Then, encoders from the two networks are frozen and used to learn the transfer function \trnet{}, which aims at transforming features extracted for \taskone{} in features that are good for \tasktwo{}. This step is performed only on domain \domainA{}, since we have no supervision for \tasktwo{} on domain \domainB{}. Finally, at inference time, features are extracted from $E_1$ starting from images of domain \domainB{}, transformed with the \trnet{} and fed to $D_2$ to produce the final predictions.}
	\label{fig:atdt_original}
\end{figure}

\section{Method}\label{sec:method}

\begin{figure*}[t]
	\centering
	\includegraphics[width=0.8\textwidth]{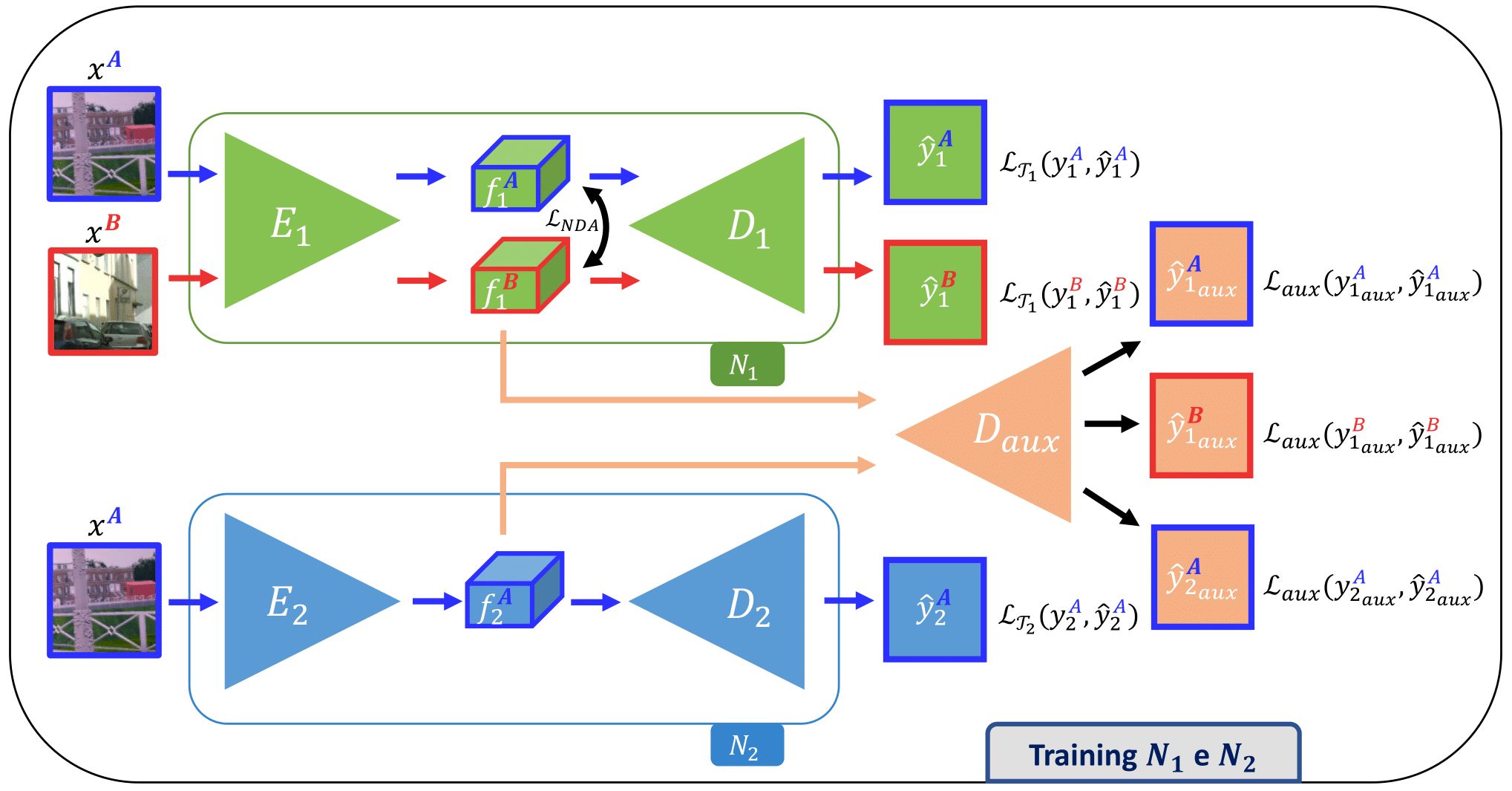}
	\caption{Features alignment strategies across tasks and domains. We train jointly the networks $N_1$, $N_2$ and a shared auxiliary decoder $D_{aux}$. We train $N_1$ to solve \taskone{} on images from domains \domainA{} and \domainB{} using a supervised loss \losstask{1} for $\mathcal{T}_1$ alongside a novel feature Norm Discrepancy Alignment loss $\mathcal{L}_{NDA}$ which helps better aligning the features computed by $N_1$ across the two domains. We train $N_2$ using a supervised loss \losstask{2} for $\mathcal{T}_2$ on images from \domainB{}. $D_{aux}$ is trained to solve an auxiliary task $\mathcal{T}_{aux}$ using the loss $\mathcal{L}_{aux}$ and based on the features computed by $E_1$ on images from \domainA{} and \domainB{} as well as by $E_2$ on images from \domainB{}.}
	\label{fig:framework}
\end{figure*}

We introduce the problem we are trying to solve with a practical example.
Imagine we aim to solve semantic segmentation in a real domain but we only have labels for a closely related task (\eg, depth estimation). 
Moreover, let us suppose to have access to a synthetic domain, where labels can be easily obtained for both tasks.
Unsupervised domain adaptation may be used in this synthetic to real scenario.  However, we wish to go one step further, trying to answer this question: can we exploit the depth estimation task to boost the performance of semantic segmentation in the real domain? The answer is yes, thanks to our novel framework \atdt{}.
In \atdt{} we first learn a mapping function in the synthetic domain between deep features of two networks trained for depth estimation and semantic segmentation. 
This mapping function captures  the relationship between the two tasks. 
Once learned, we use the mapping on depth features extracted from real samples to solve semantic segmentation in the real domain without the need of labels for it, thereby  transferring knowledge across tasks and domains.
To further improve performance, we propose two strategies aimed at increasing the transferability of the learned features, namely leveraging on a norm discrepancy alignment loss and an auxiliary task.

In the following sub-sections, we first describe the base \atdt{} framework and then delineate its improved formulation which includes the  norm discrepancy alignment loss and  auxiliary task.

\subsection{Notation}
We consider two tasks, \taskone{} and \tasktwo{}, as well as two domains, \domainA{} and \domainB{}.
We denote the images belonging to \domainA{} and \domainB{} as $x^{\mathcal{A}}$  and $x^{\mathcal{B}}$, respectively. 
We have labels for \taskone{} in \domainA{} and \domainB{}, denoted as $y^\mathcal{A}_1$ and $y^\mathcal{B}_1$, respectively. 
On the other hand, we have labels for \tasktwo{} only in \domainA{}, denoted as $y^\mathcal{A}_2$.
Our aim is to solve \tasktwo{} in \domainB{}, where we do not have supervision.
We assume  \taskone{} and \tasktwo{} to be both dense tasks, which can therefore be addressed by an encoder-decoder architecture. We denote as $N_1$ and $N_2$ two networks that solve \taskone{} and \tasktwo{}, respectively. Each network $N_k, k\in \{1,2\}$ consists of an encoder $E_k$ and a decoder $D_k$, such that $N_k(x) = D_k(E_k(x))$, $x$ being the input image.

\subsection{Across Tasks and Domains Transfer}
\label{subsec:atdt}

In our \atdt{} framework we aim at learning the relationships between \taskone{} and \tasktwo{} through a neural network.   This is achieved by 3 steps, each  represented as a block in \autoref{fig:atdt_original}:\\

\textbf{Training $N_1$ and $N_2$.}
We train $N_1$ and $N_2$ to solve \taskone{} and \tasktwo{}. Since we assume supervision for \taskone{} on both domains, $N_1$ is trained with images from \domainA{} and \domainB{}. This enables  $N_1$ to learn  a feature space shared across the two domains. $N_2$, instead, is trained only on \domainA{}.
Both networks are trained with a specific supervised task loss \losstask{k} for $\mathcal{T}_k$.\\

\textbf{Training \trnet{}.}
Considering only domain \domainA{}, where we have supervision for both tasks, we then train a transfer network \trnet{} to map the features computed by  $N_1$,  $f_1^{\mathcal{A}}=E_1(x^{\mathcal{A}})$, into those computed by $N_2$,  $f_2^{\mathcal{A}}=E_2(x^{\mathcal{A}})$.  Denoting  the transferred features as $f_{1\rightarrow2}^{\mathcal{A}}={G_{1\rightarrow2}}(f_1^{\mathcal{A}})$, we train the transfer network by minimizing the $L_2$ loss:
\begin{equation}
    \mathcal{L}_{Tr}= ||f_{1\rightarrow2}^{\mathcal{A}} - f_2^{\mathcal{A}}||_2
\end{equation}

\textbf{Inference.}
Once \trnet{} has been trained, we can address \tasktwo{} in \domainB{} by computing the features to solve \taskone{}, $f_1^{\mathcal{B}}=E_1(x^{\mathcal{B}})$, transform them into features amenable to \tasktwo{},   $f_{1\rightarrow2}^{\mathcal{B}}={G_{1\rightarrow2}}(f_1^{\mathcal{B}})$, and finally decode these features into the required dense output by $D_2$:
\begin{equation}
    \hat{y}^B_2 = D_2(f_{1\rightarrow2}^{\mathcal{B}})
\end{equation}\\

After presenting the base \atdt{} framework, in the next sub-sections we will describe two strategies deployed to boost the feature alignment across domains and tasks. \autoref{fig:framework} provides a detailed view of these two strategies which in our final proposed framework replace the initial steps of the training protocol (\ie, Training $N_1$ and $N_2$).

\begin{figure*}[htbp]
    \centering
	\includegraphics[width=15cm]{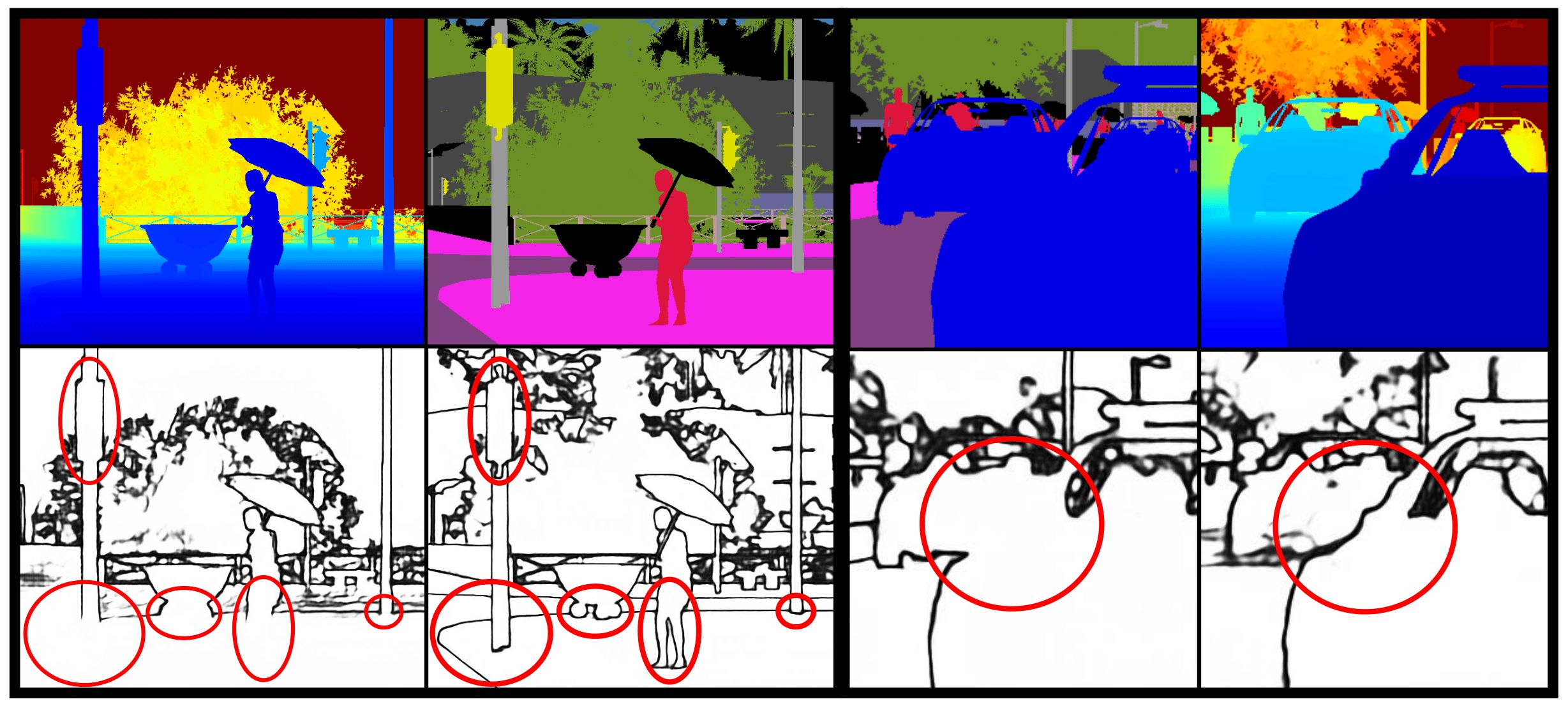}
	\caption{Two task transfer scenarios: depth-to-semantic on the left, the opposite on the right. First row: ground-truth depth and semantic segmentation maps; second row: corresponding edge maps. Red circles highlight information needed in the target task but missing in the source one. 
	}
	\label{fig:teaser}
\end{figure*}

\subsection{Feature Alignment Across Domains}\label{sec:nda}
For the effectiveness of the approach delineated in \autoref{subsec:atdt},  it is crucial that \trnet{} can generalize well to the target  unseen domain \domainB{} even if trained only with data from the source domain \domainA{}.

\begin{figure*}[ht]
    \centering
    \setlength{\tabcolsep}{1pt}
    \begin{tabular}{cccccccc}
        \domainA{} & \domainB{} & \domainA{} & \domainB{} & \domainA{} & \domainB{} & \domainA{} & \domainB{} \\
        \includegraphics[width=0.12\textwidth]{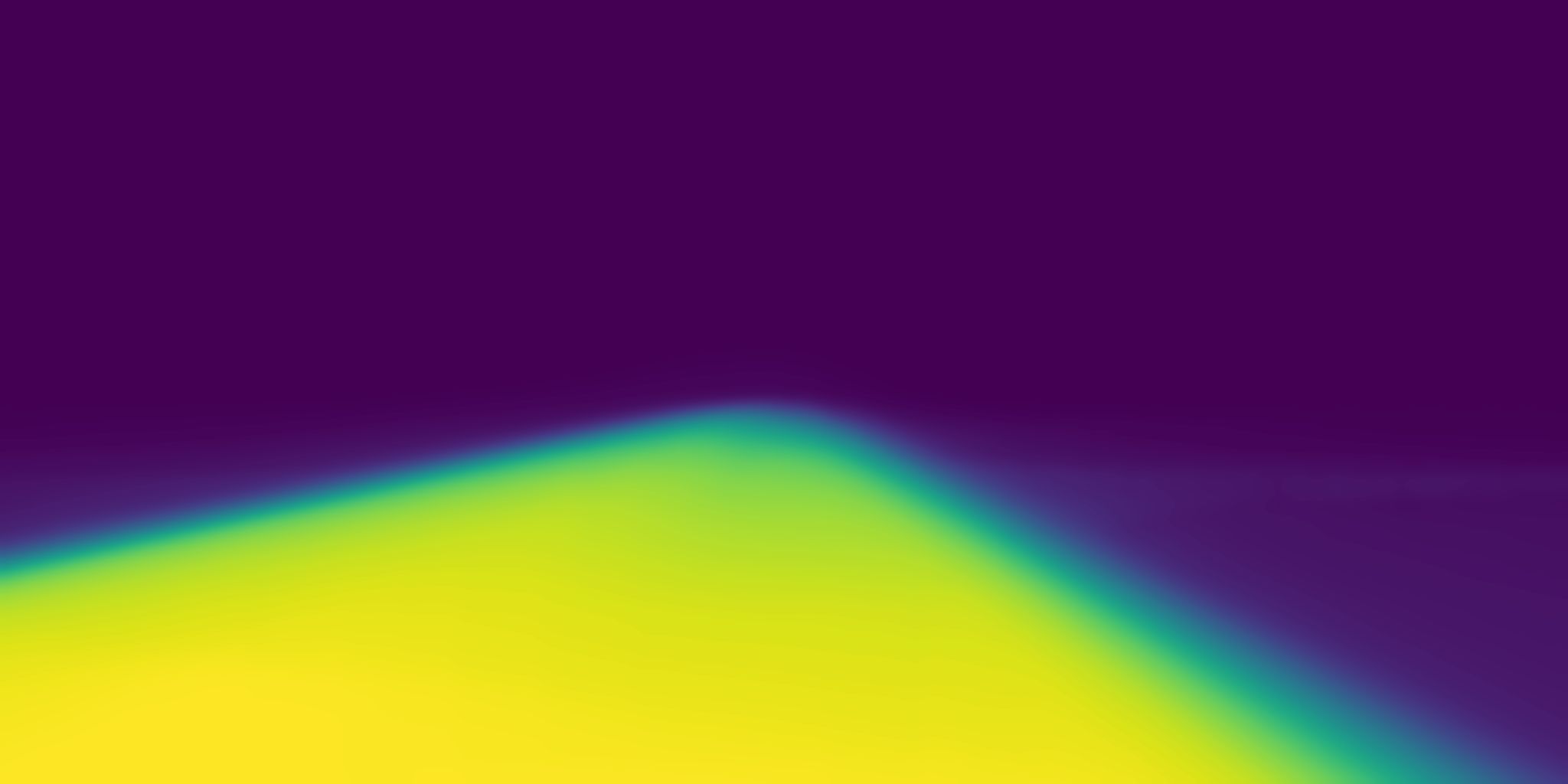} &
        \includegraphics[width=0.12\textwidth]{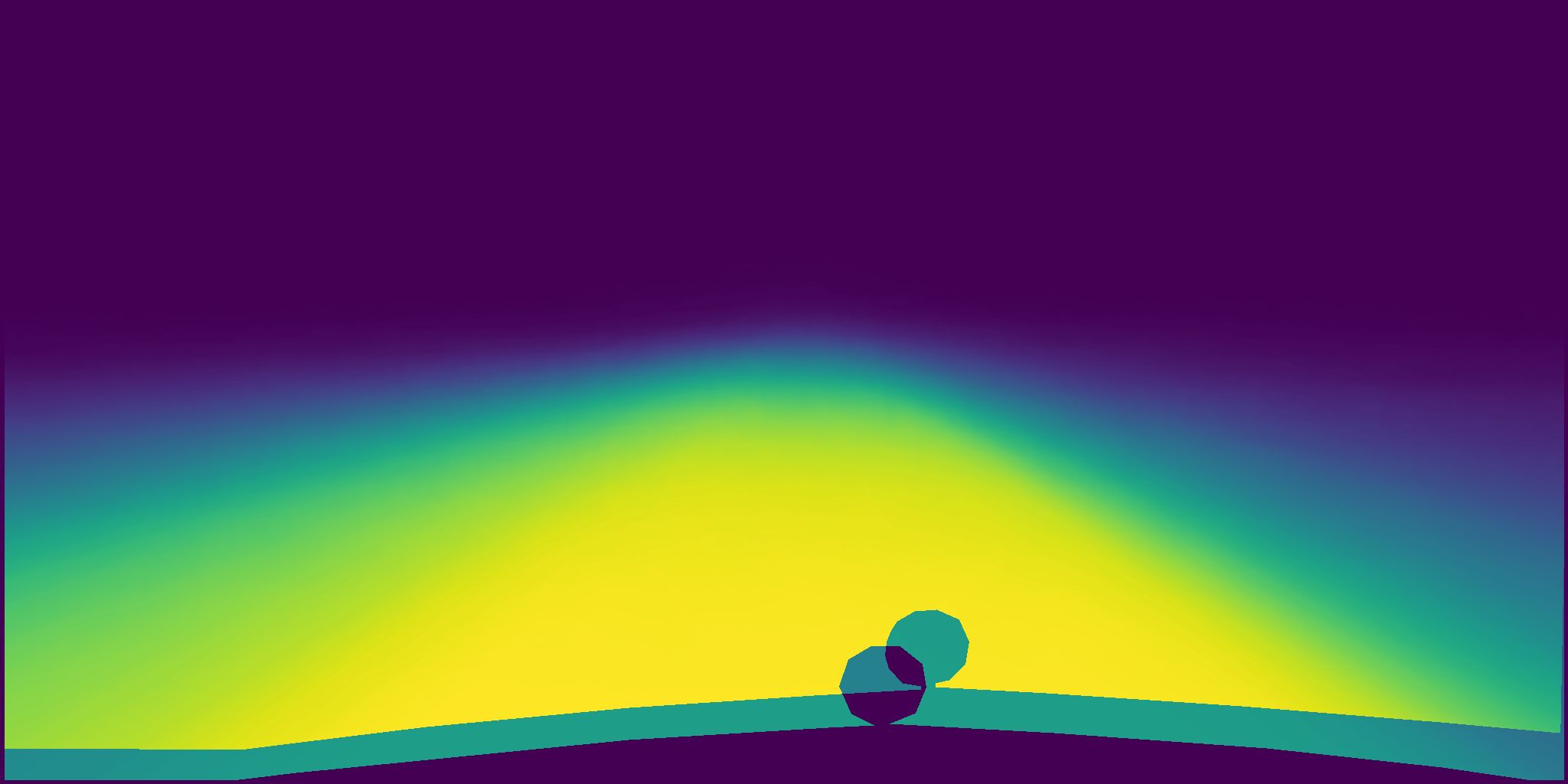} &
        \includegraphics[width=0.12\textwidth]{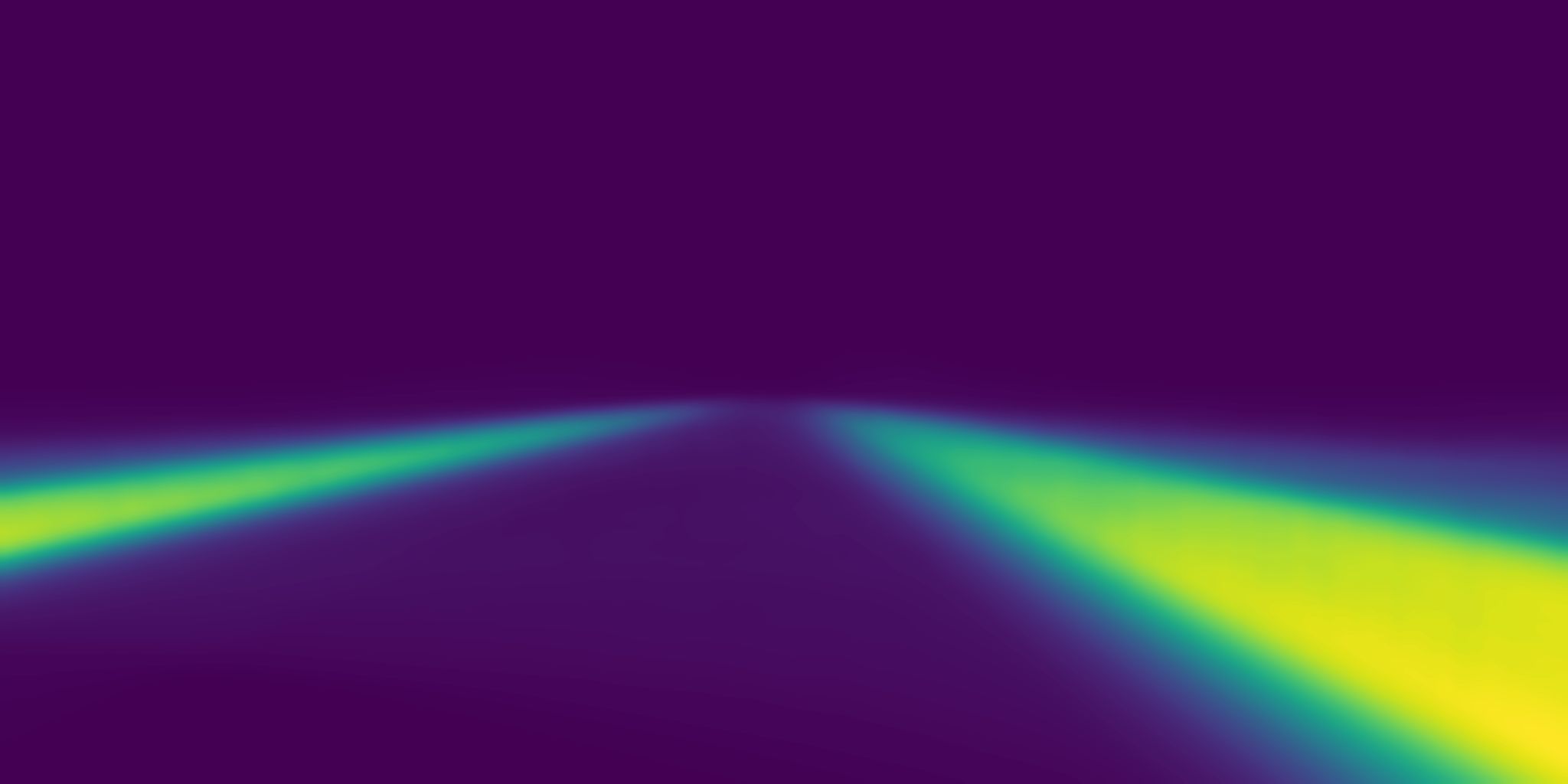} &
        \includegraphics[width=0.12\textwidth]{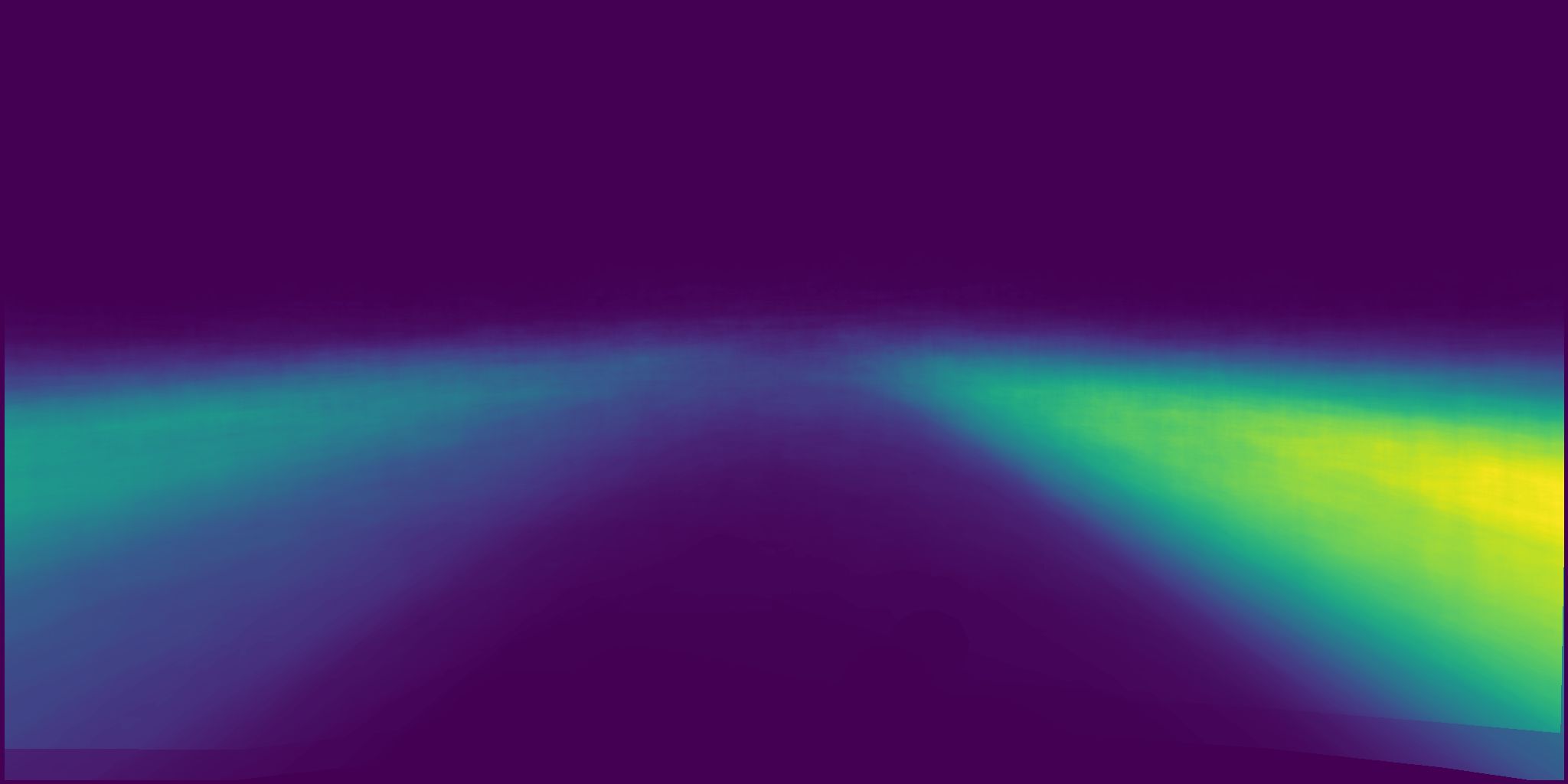} &
        \includegraphics[width=0.12\textwidth]{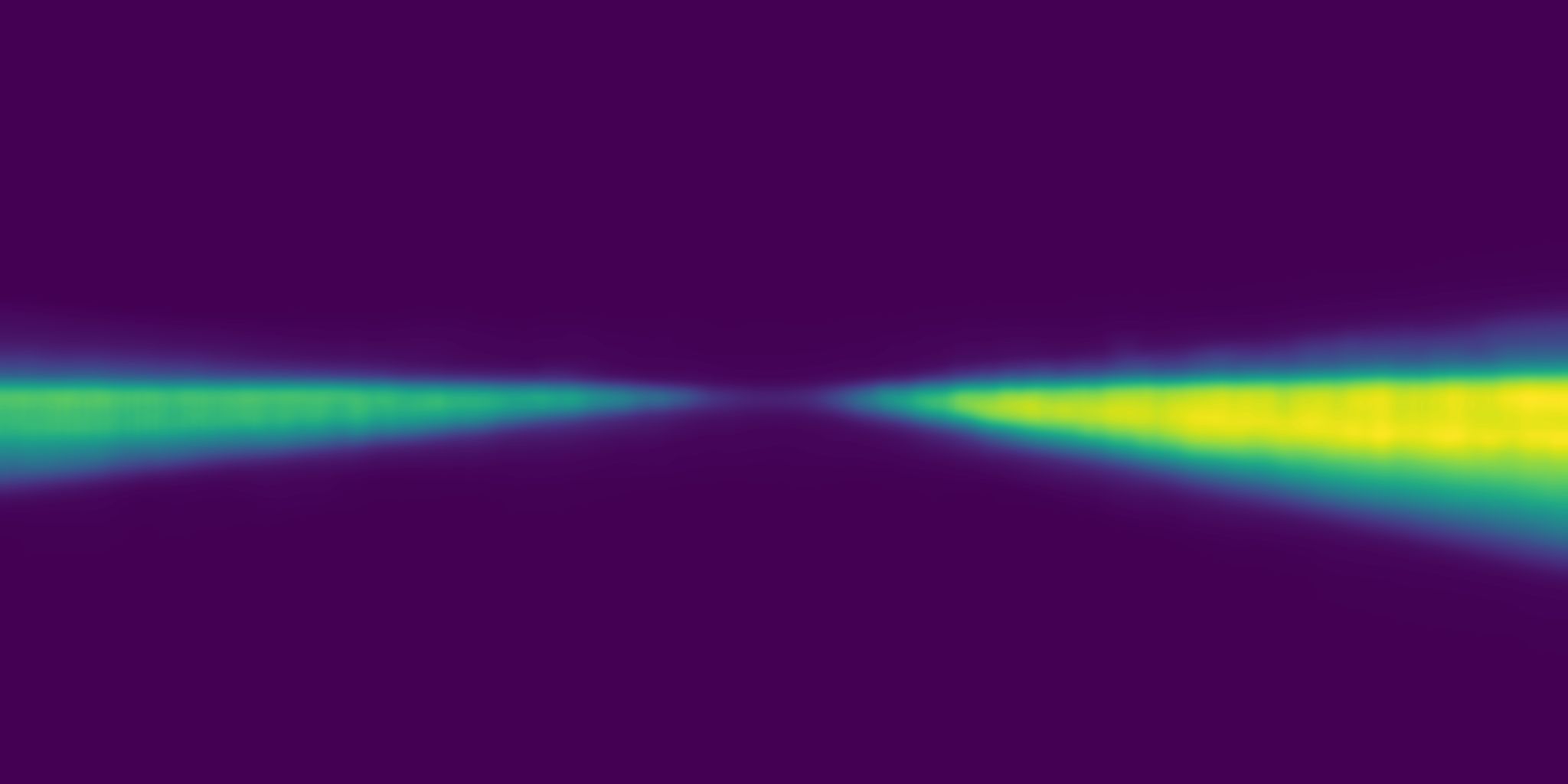} & 
        \includegraphics[width=0.12\textwidth]{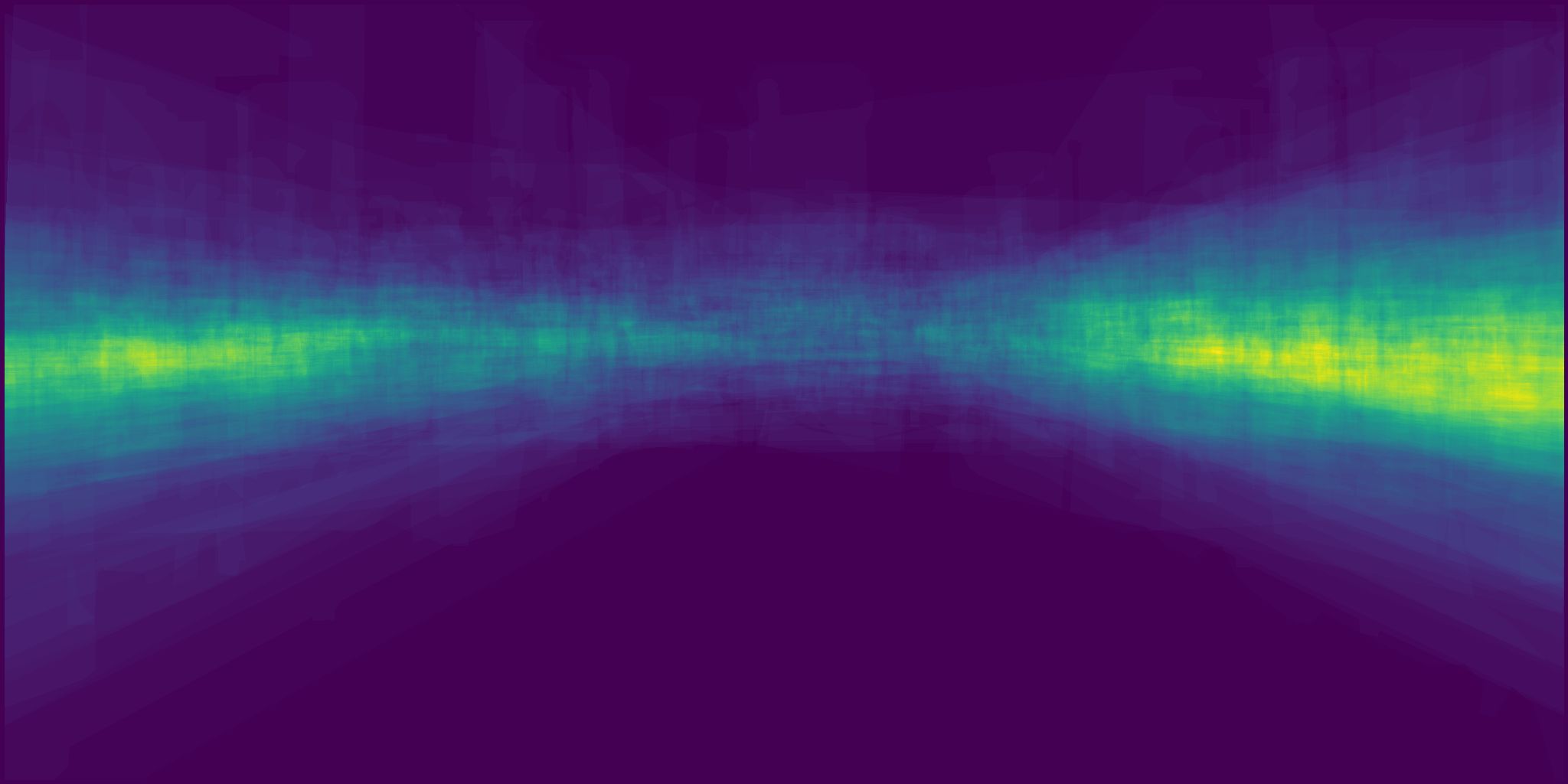} &
        \includegraphics[width=0.12\textwidth]{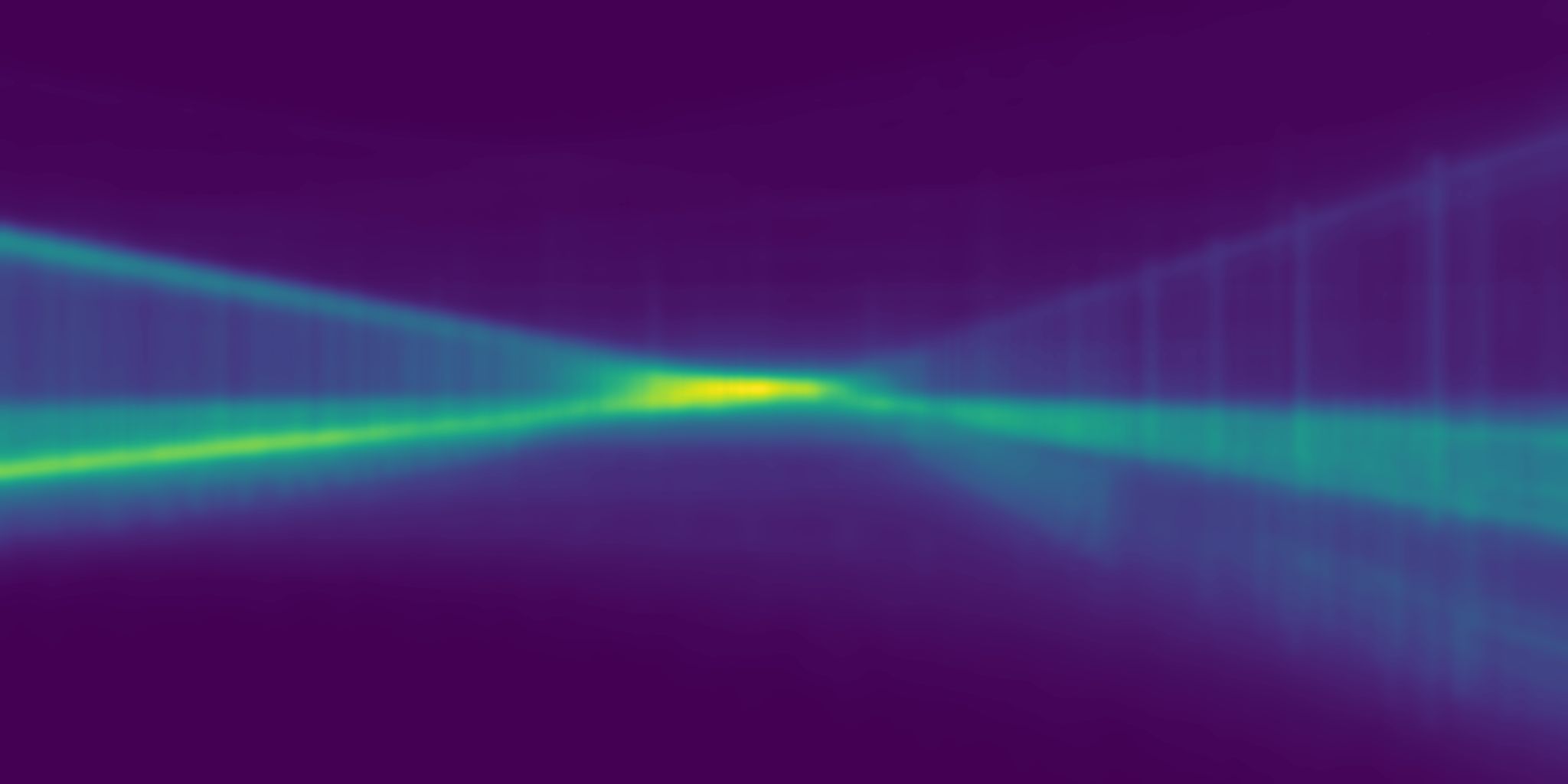} &
        \includegraphics[width=0.12\textwidth]{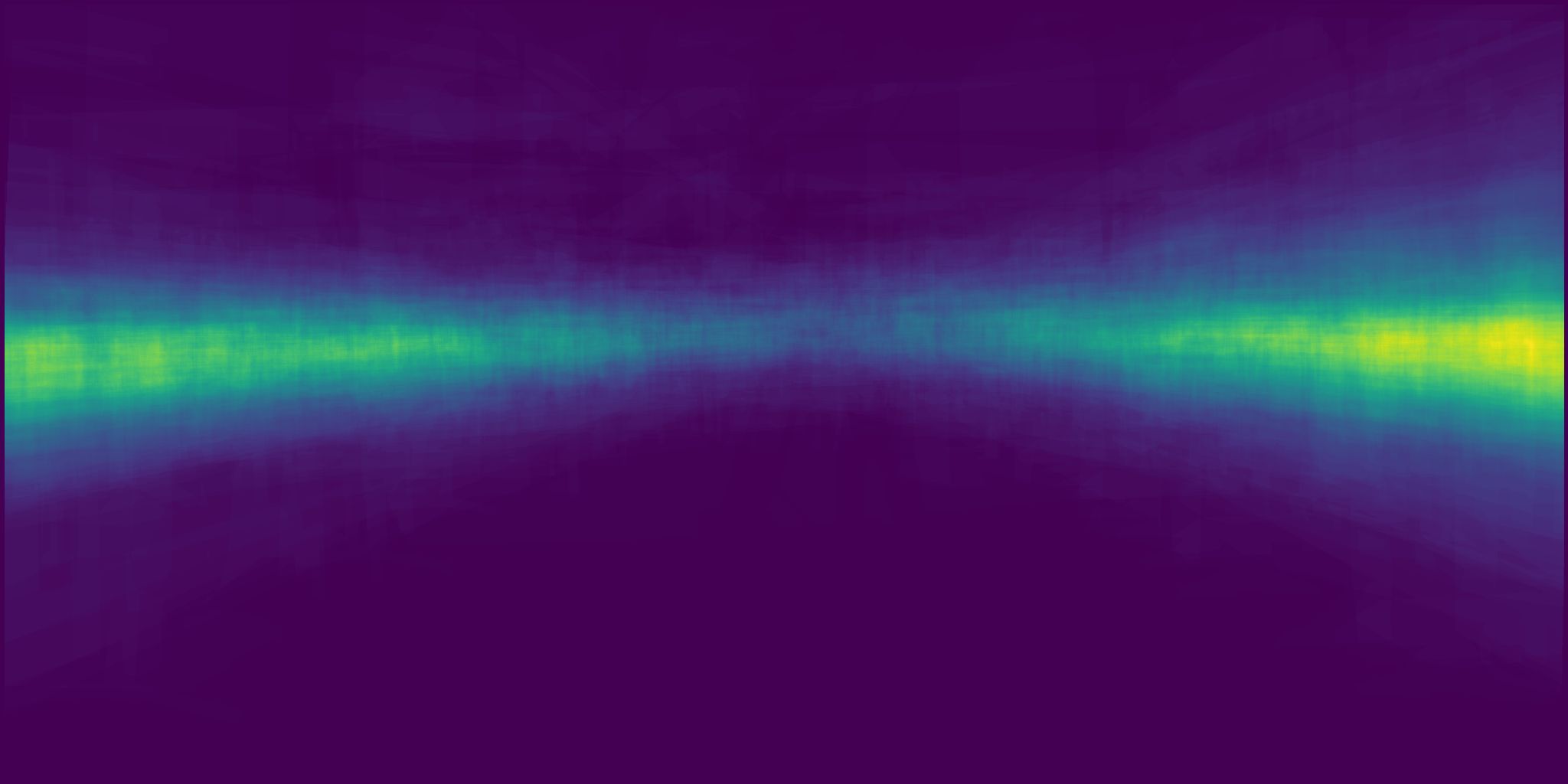} \\
        \multicolumn{2}{c}{\textit{Road}} & \multicolumn{2}{c}{\textit{Sidewalk}} & \multicolumn{2}{c}{\textit{Wall}} & 
        \multicolumn{2}{c}{\textit{Fence}} \\
        \domainA{} & \domainB{} & \domainA{} & \domainB{} & \domainA{} & \domainB{} & \domainA{} & \domainB{} \\
        \includegraphics[width=0.12\textwidth]{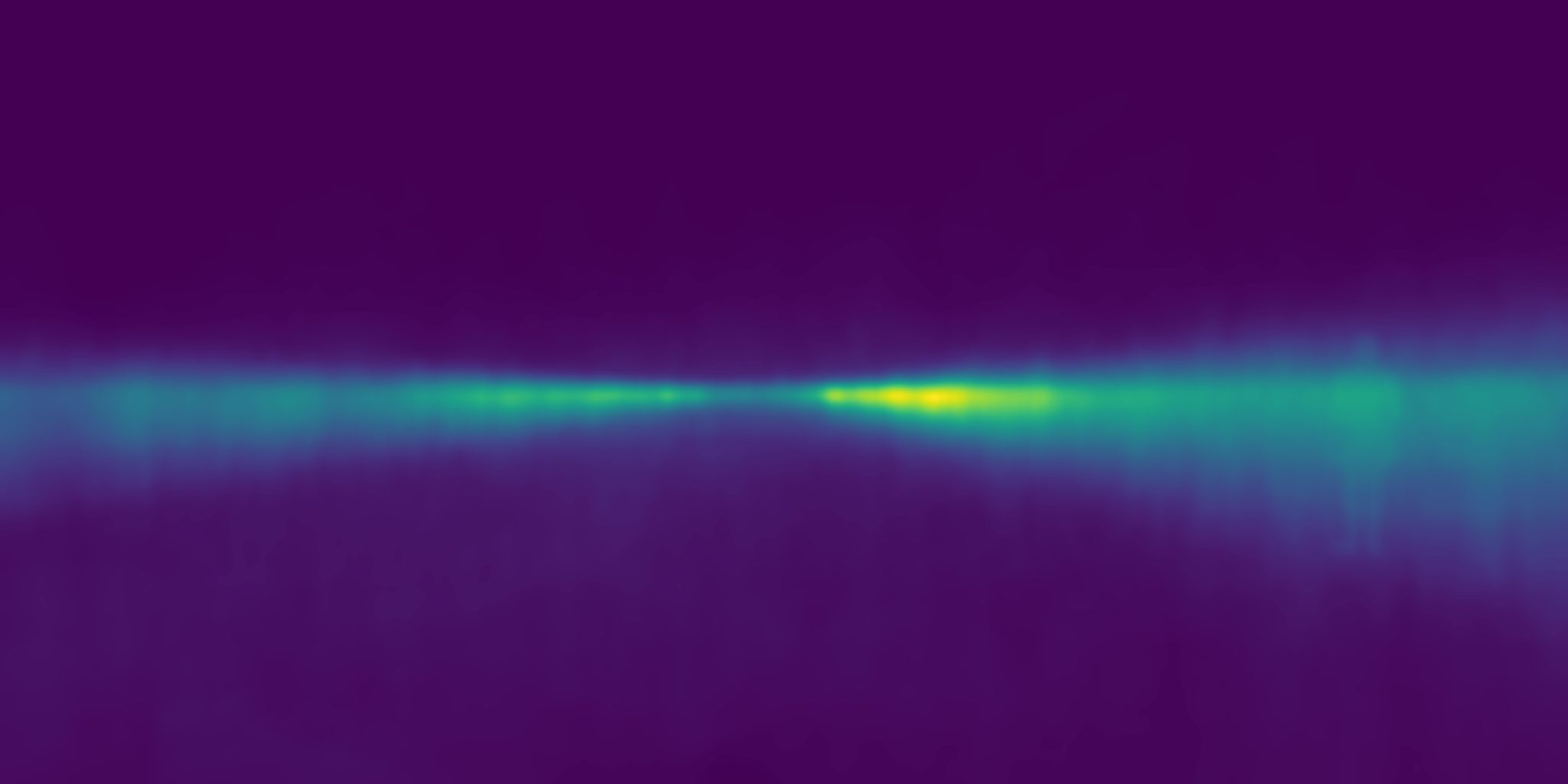} &
        \includegraphics[width=0.12\textwidth]{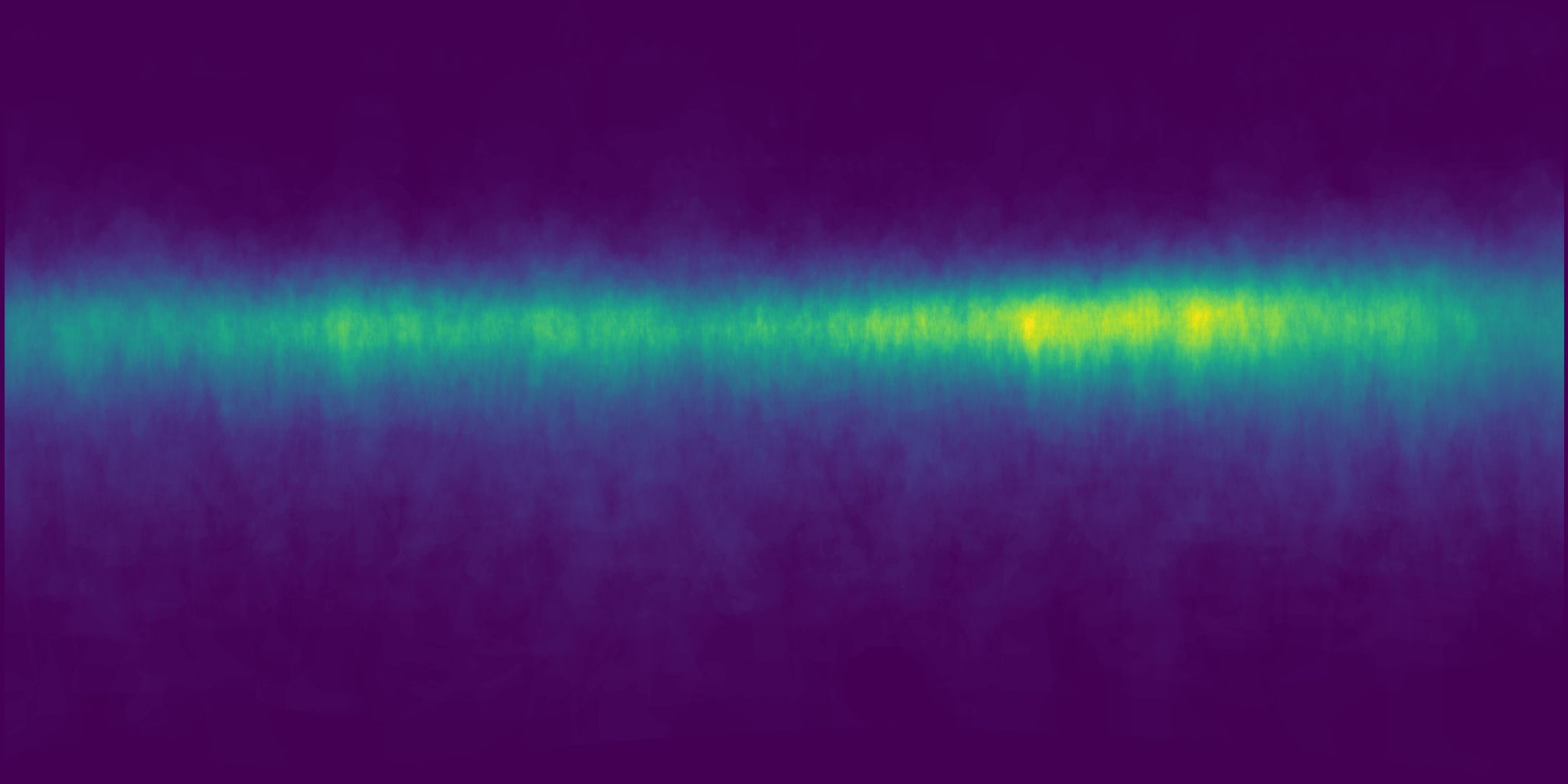} &
        \includegraphics[width=0.12\textwidth]{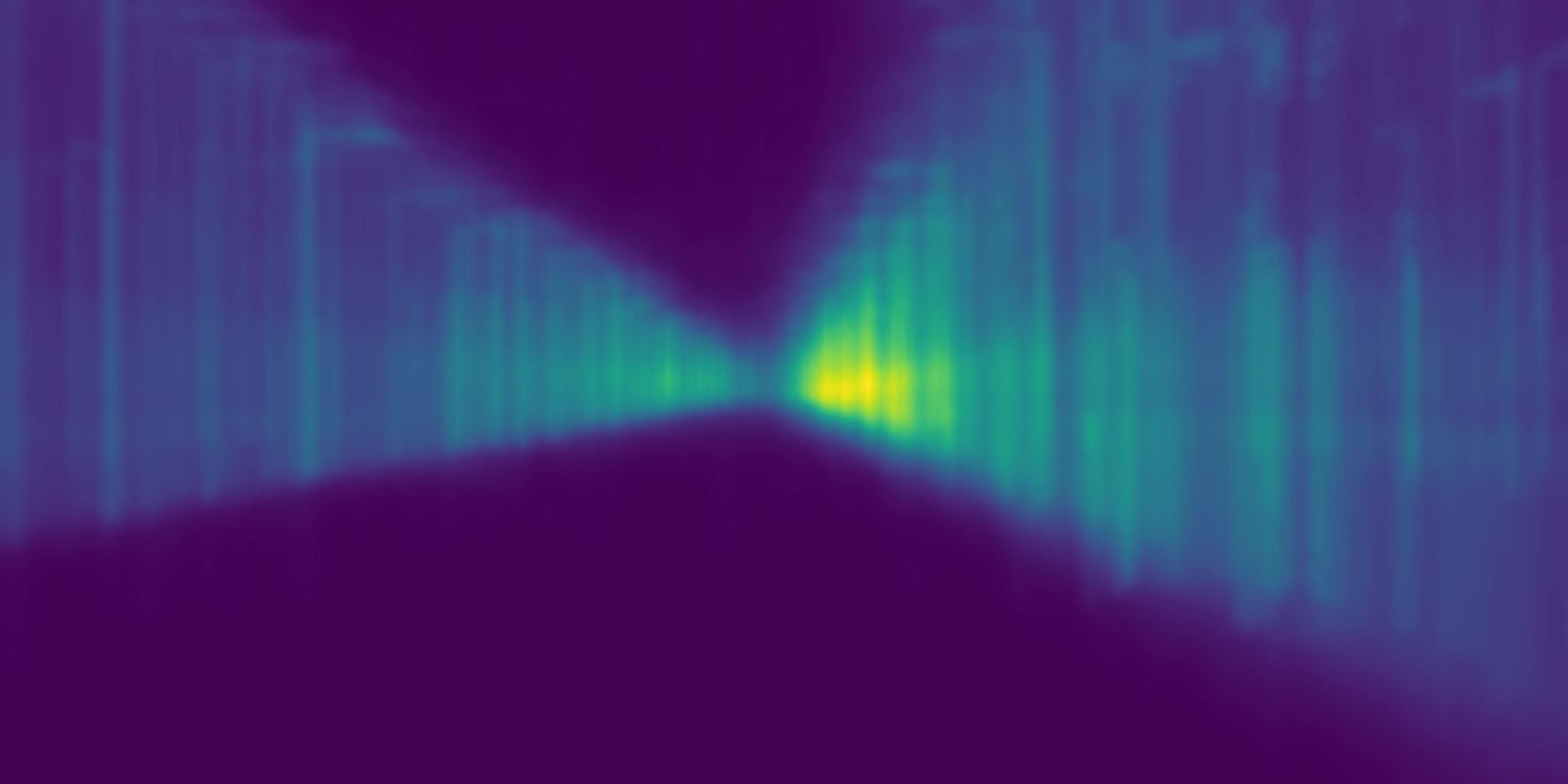} &
        \includegraphics[width=0.12\textwidth]{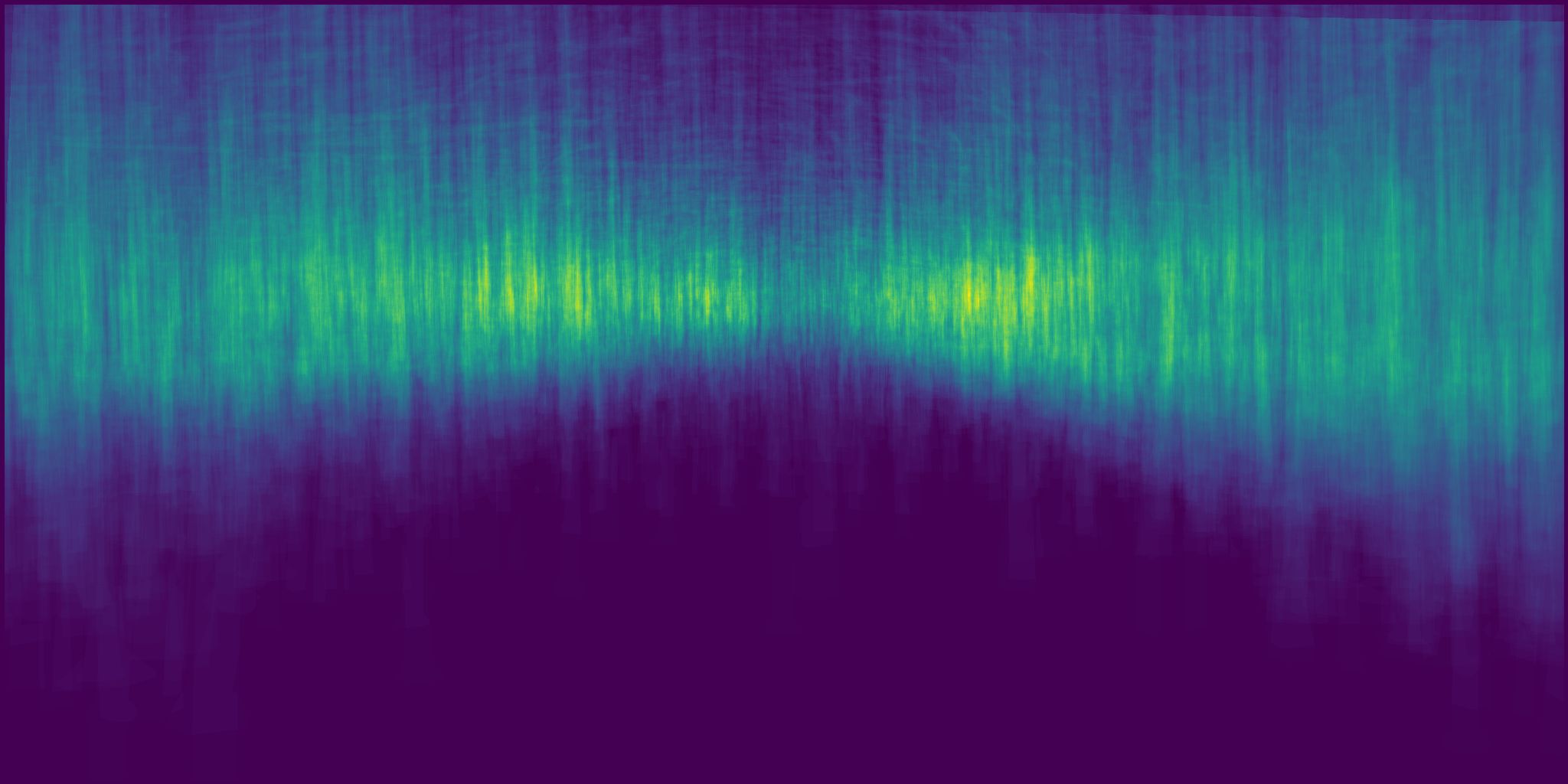} &
        \includegraphics[width=0.12\textwidth]{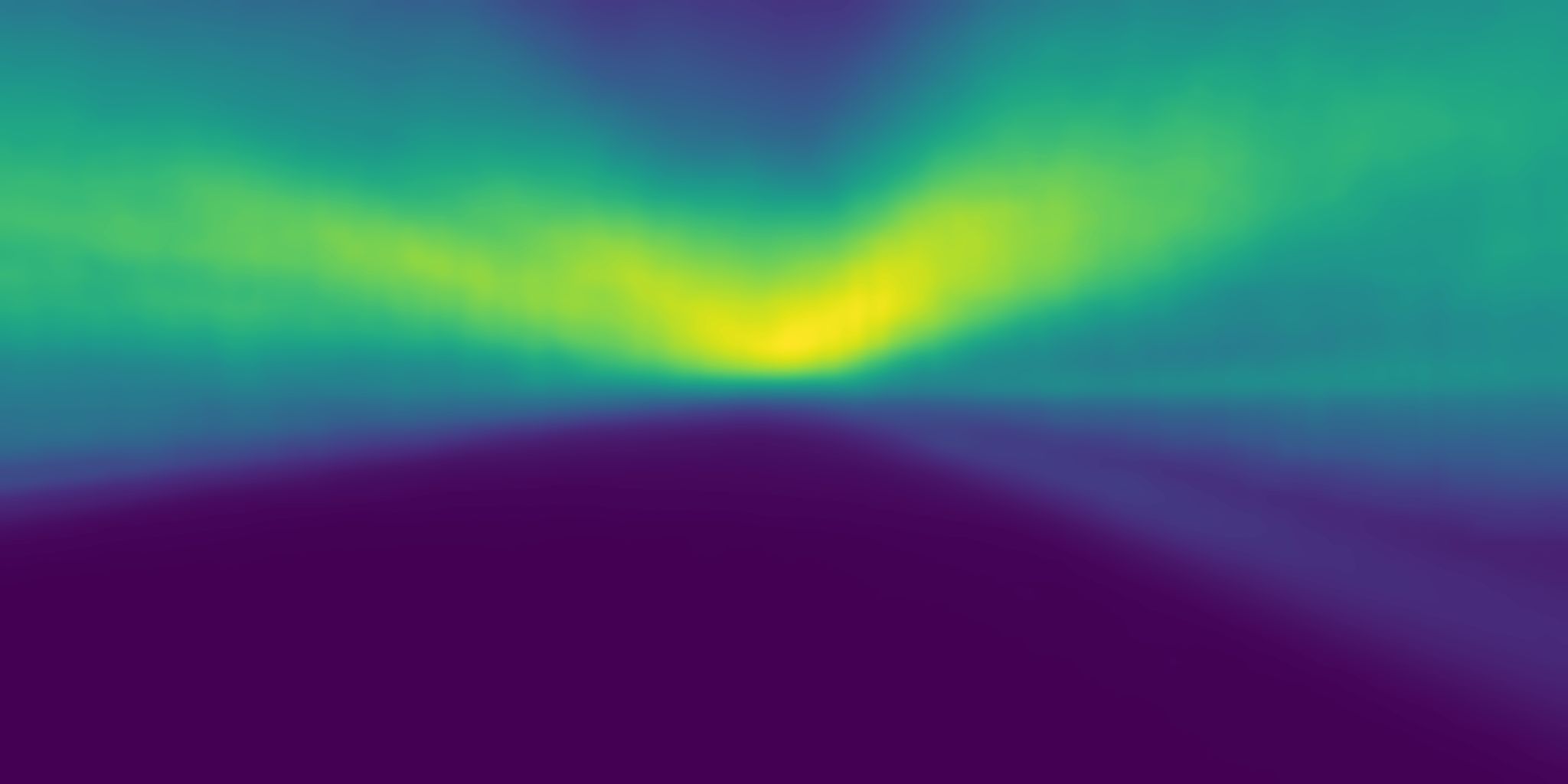} &
        \includegraphics[width=0.12\textwidth]{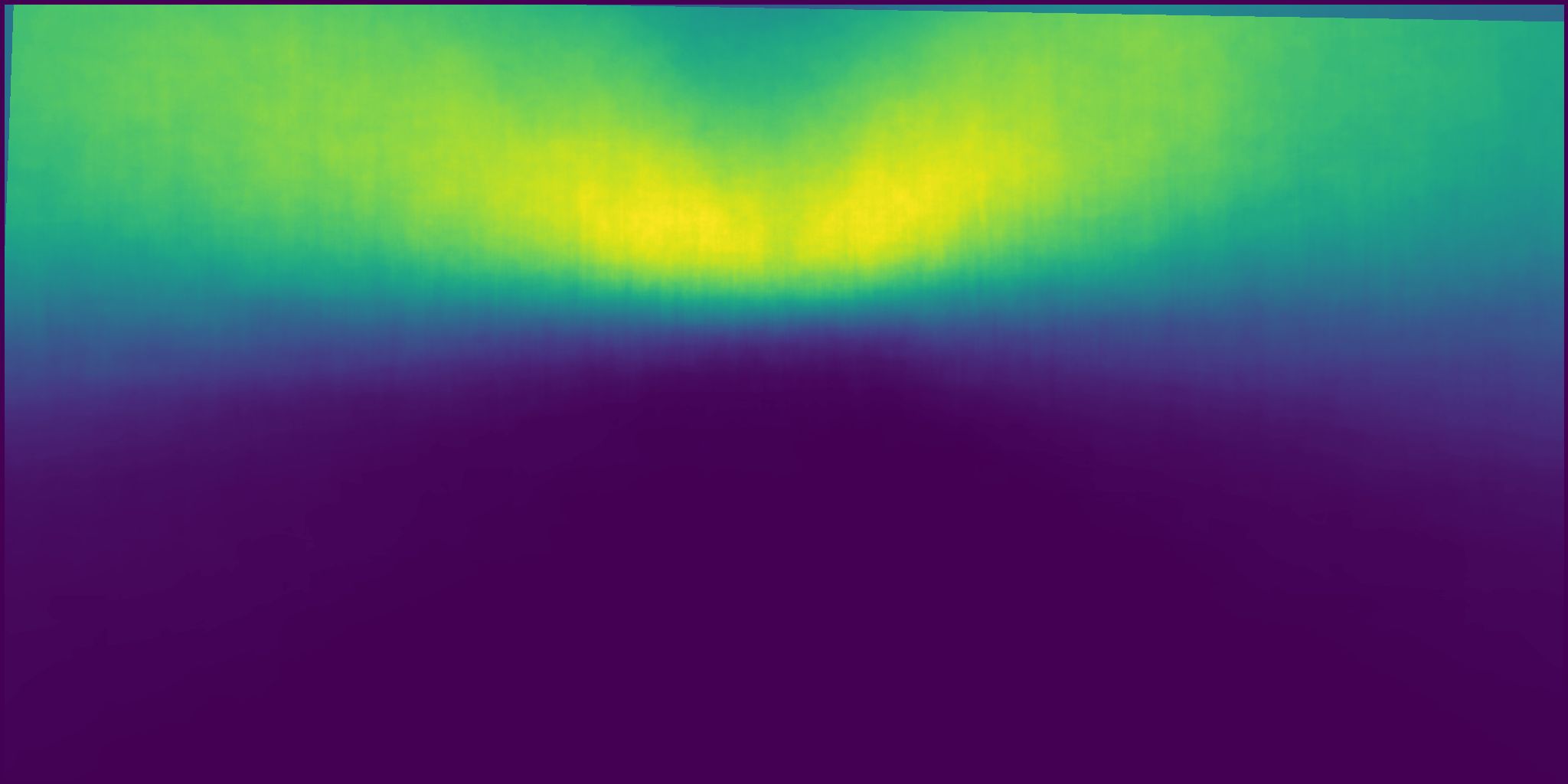} &
        \includegraphics[width=0.12\textwidth]{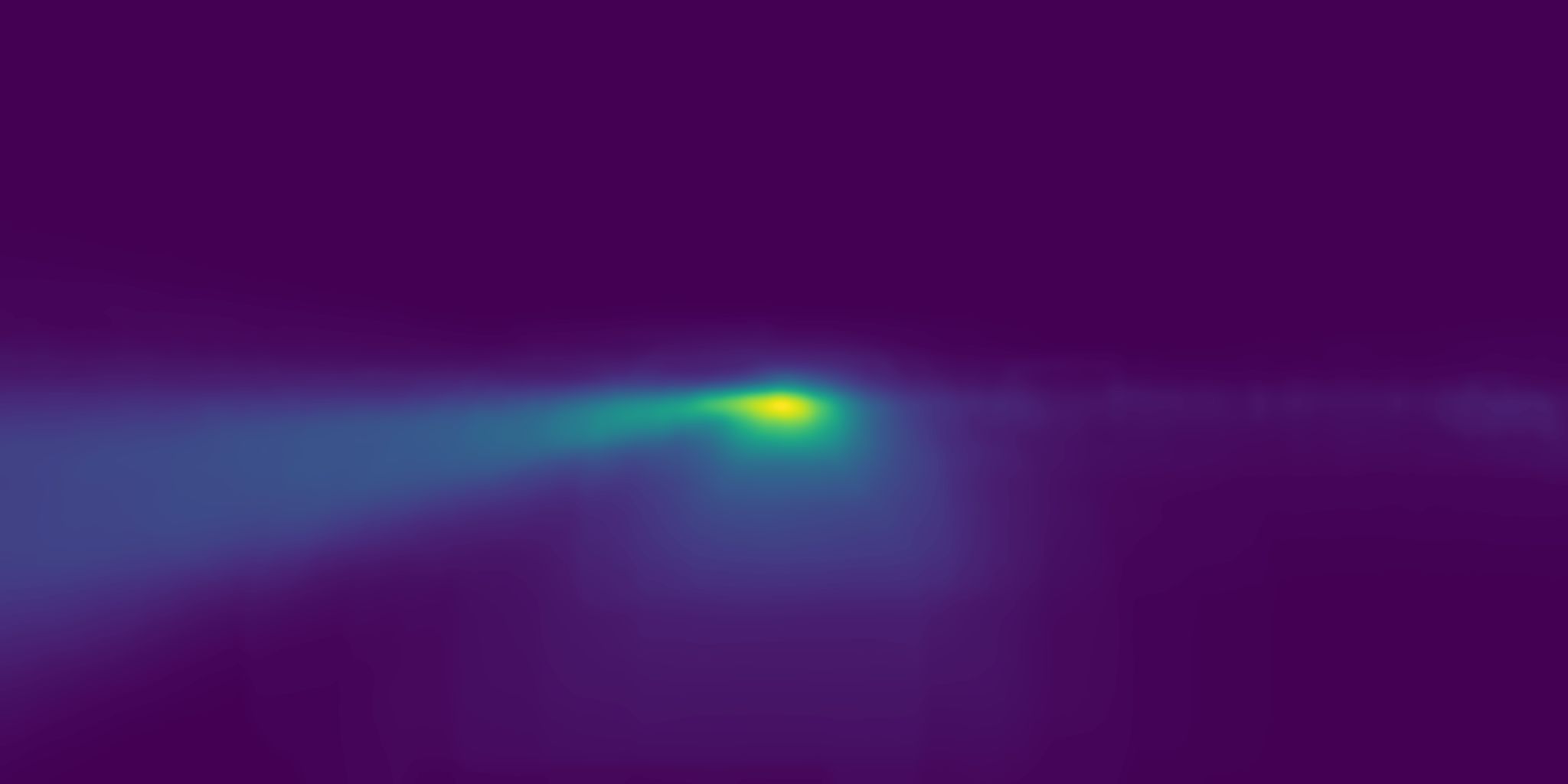} &
        \includegraphics[width=0.12\textwidth]{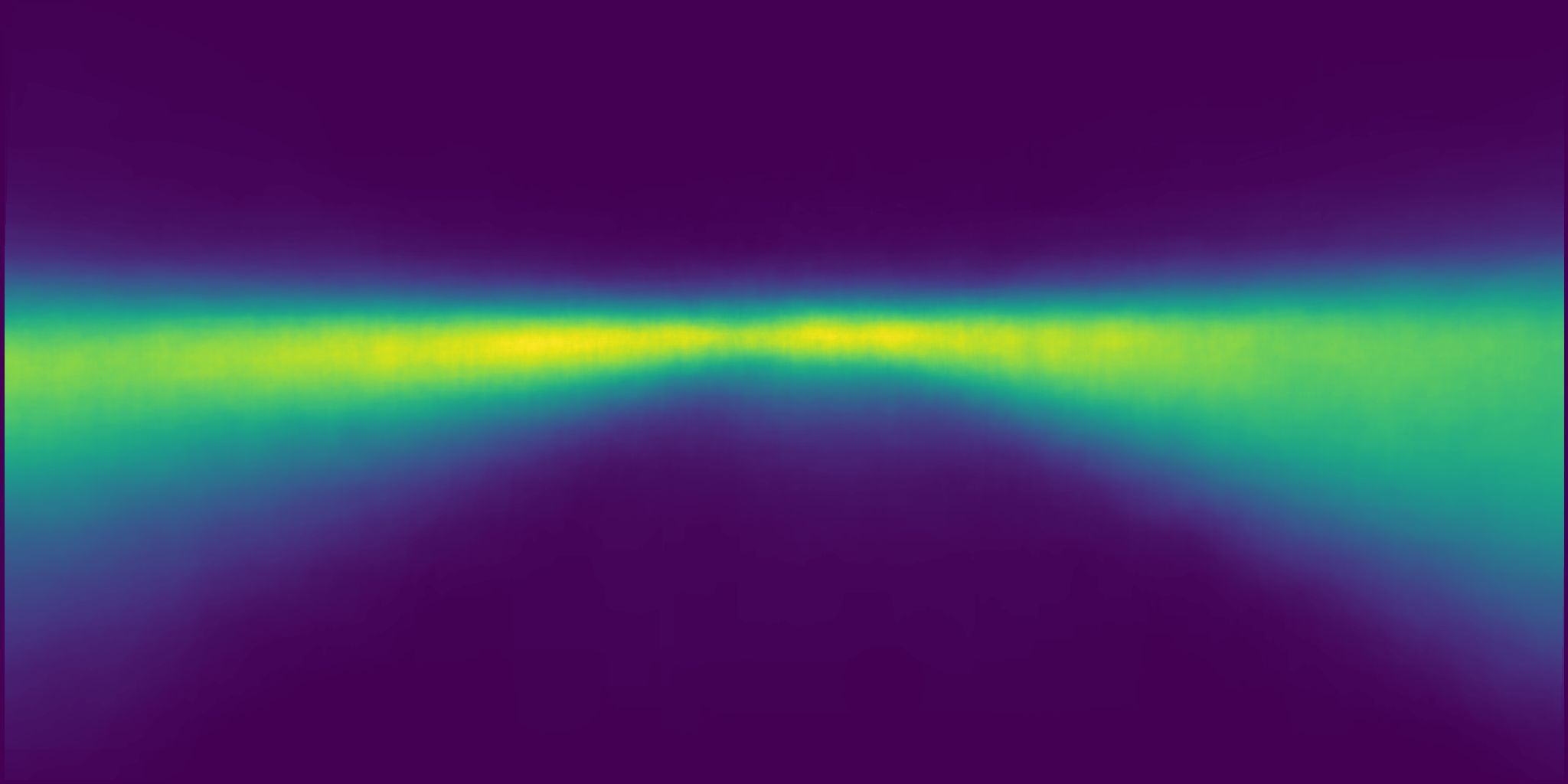}\\
        \multicolumn{2}{c}{\textit{Person}} & 
        \multicolumn{2}{c}{\textit{Pole}} &
        \multicolumn{2}{c}{\textit{Vegetation}} & 
        \multicolumn{2}{c}{\textit{Vehicle}} \\
        & \domainA{} & \domainB{} & \domainA{} & \domainB{} & \domainA{} & \domainB{} & \\
        &
        \includegraphics[width=0.12\textwidth]{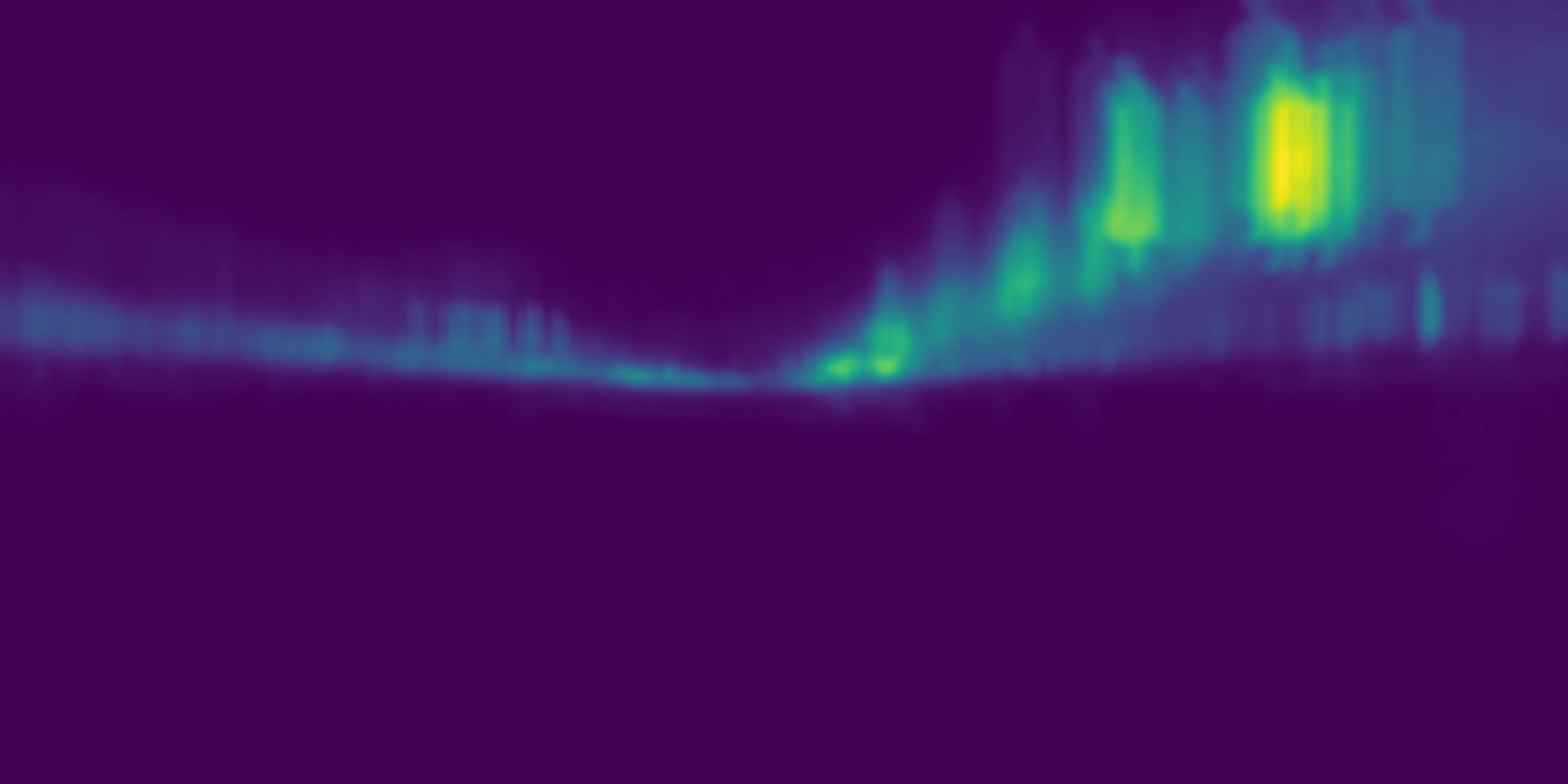} &
        \includegraphics[width=0.12\textwidth]{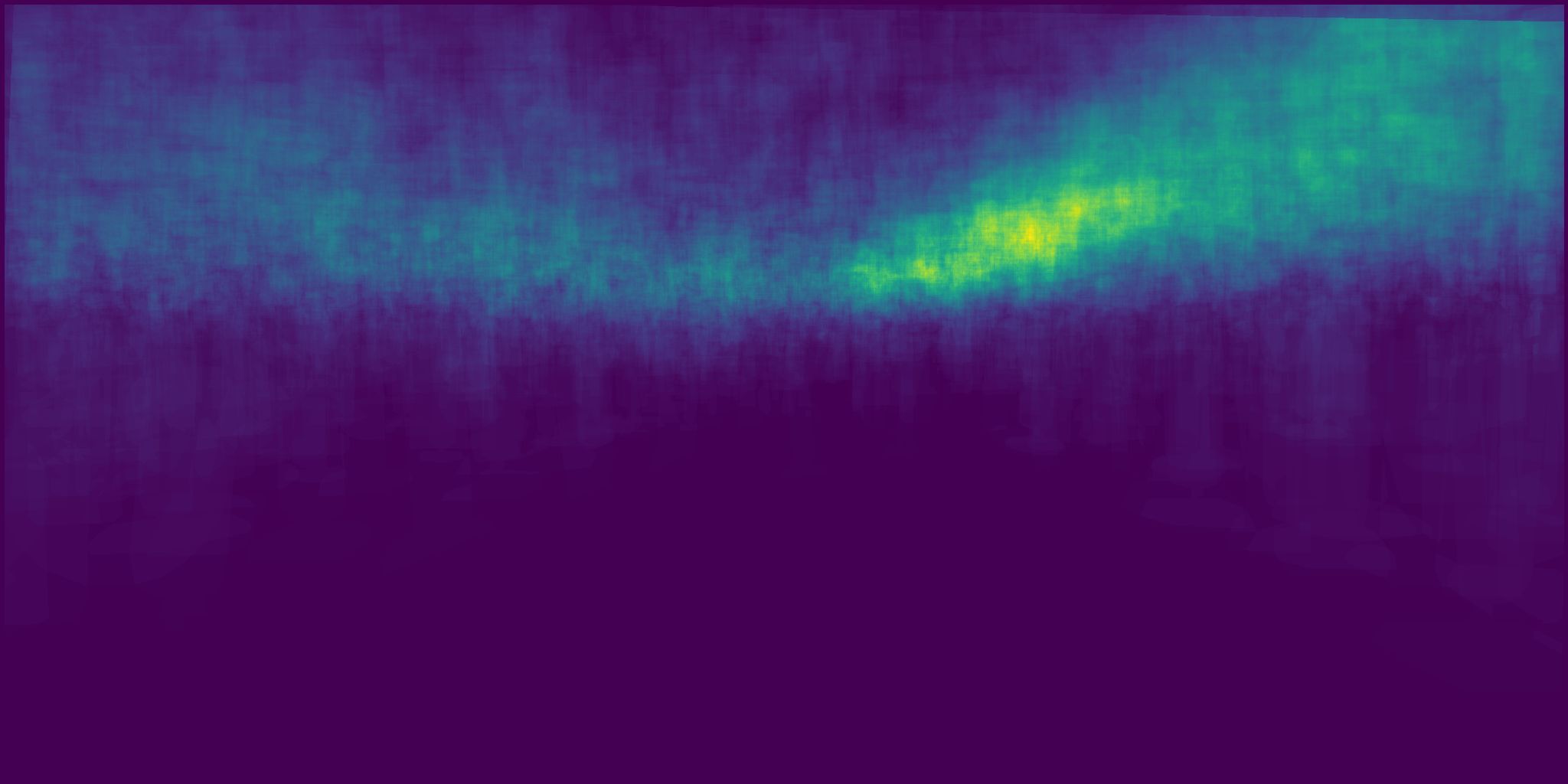} &
        \includegraphics[width=0.12\textwidth]{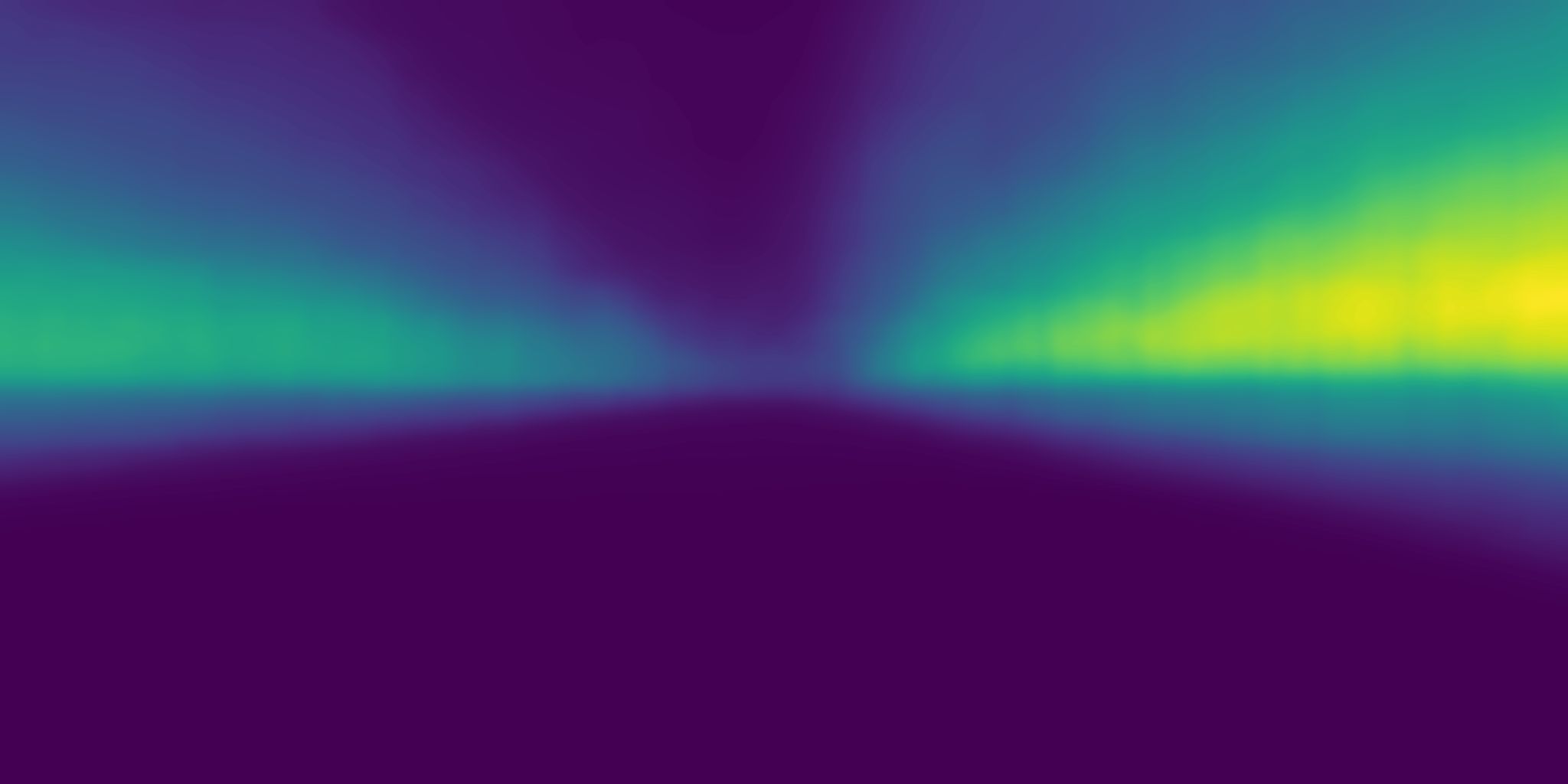} &
        \includegraphics[width=0.12\textwidth]{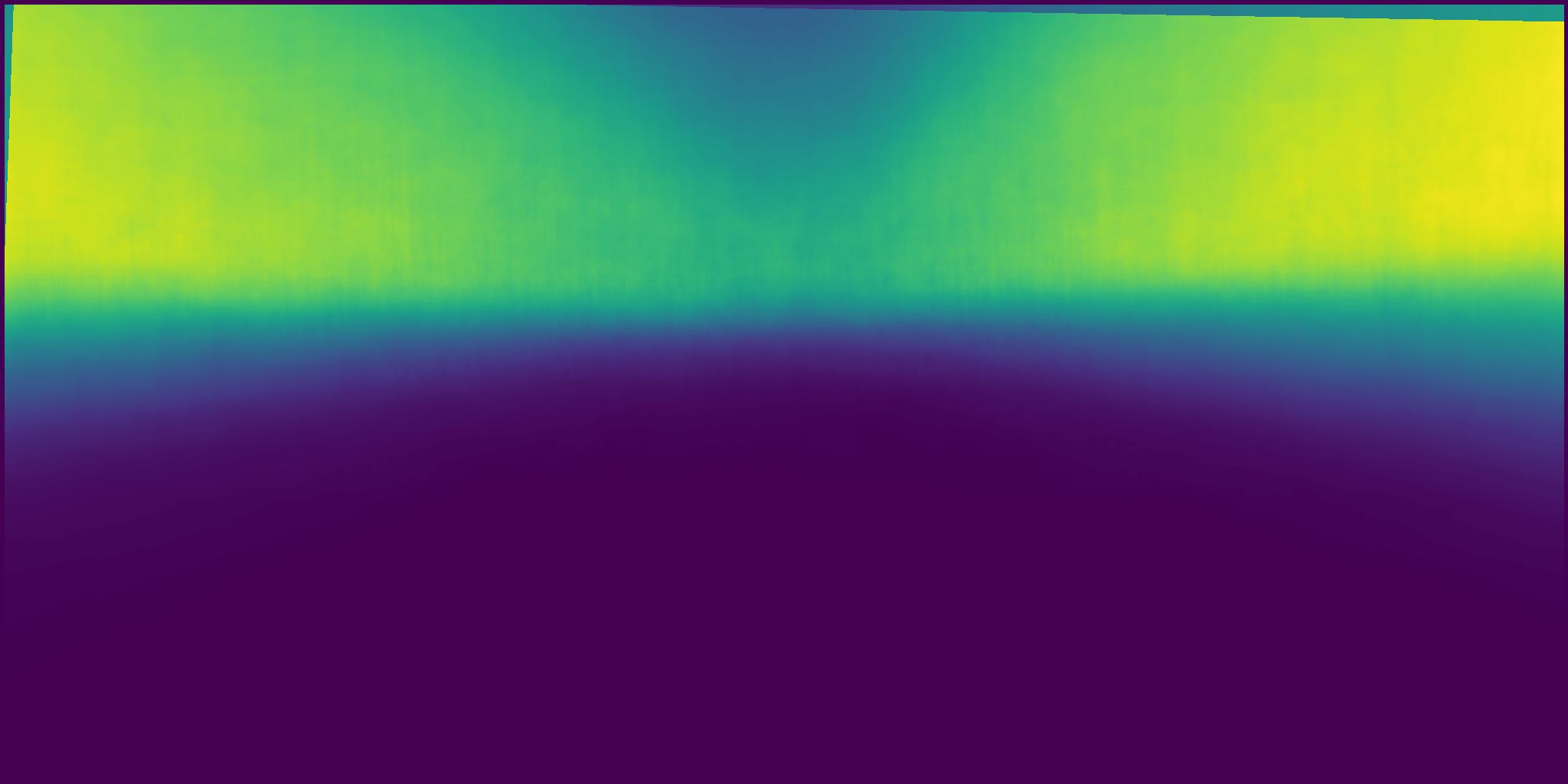} &
        \includegraphics[width=0.12\textwidth]{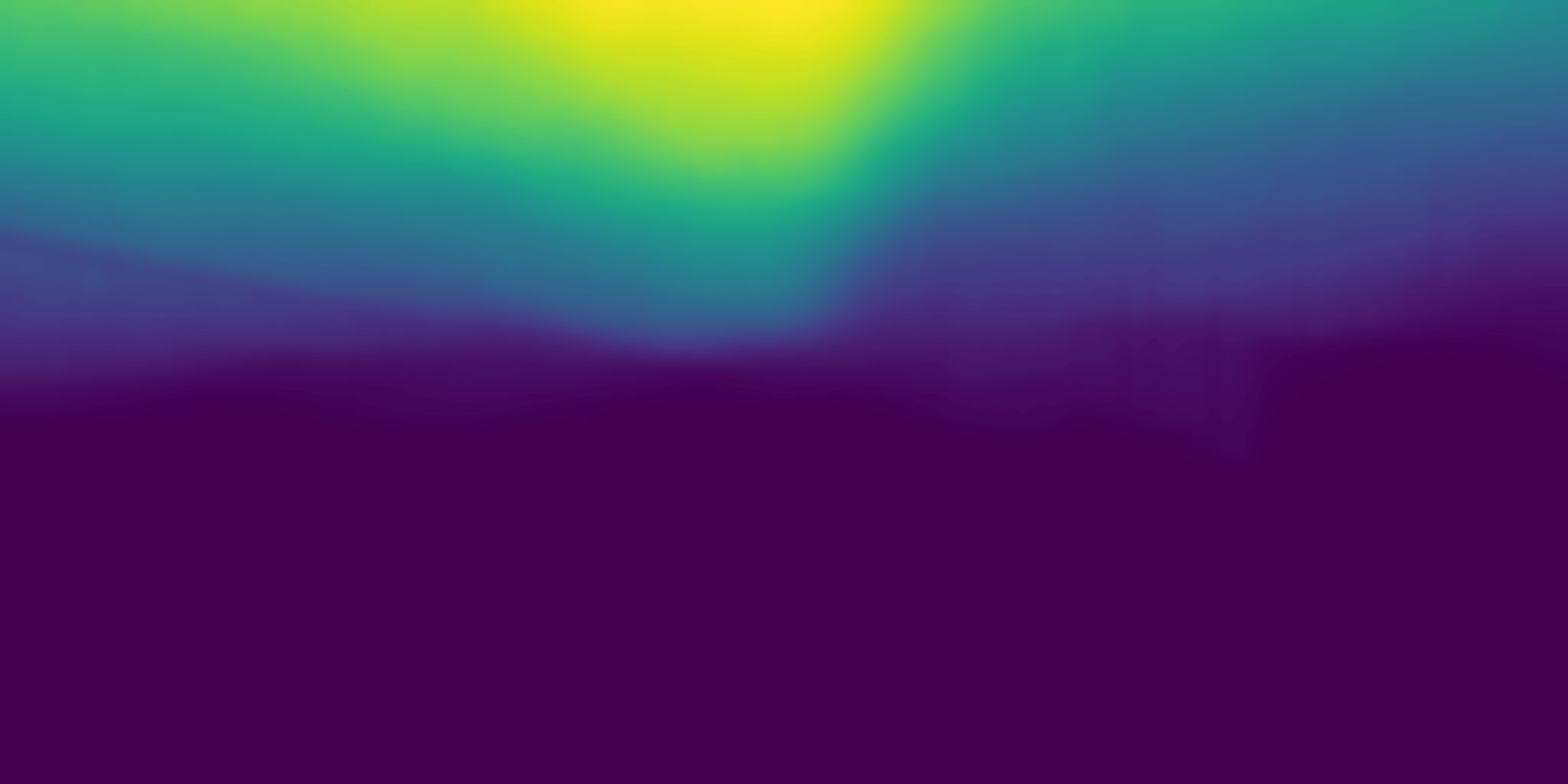} &
        \includegraphics[width=0.12\textwidth]{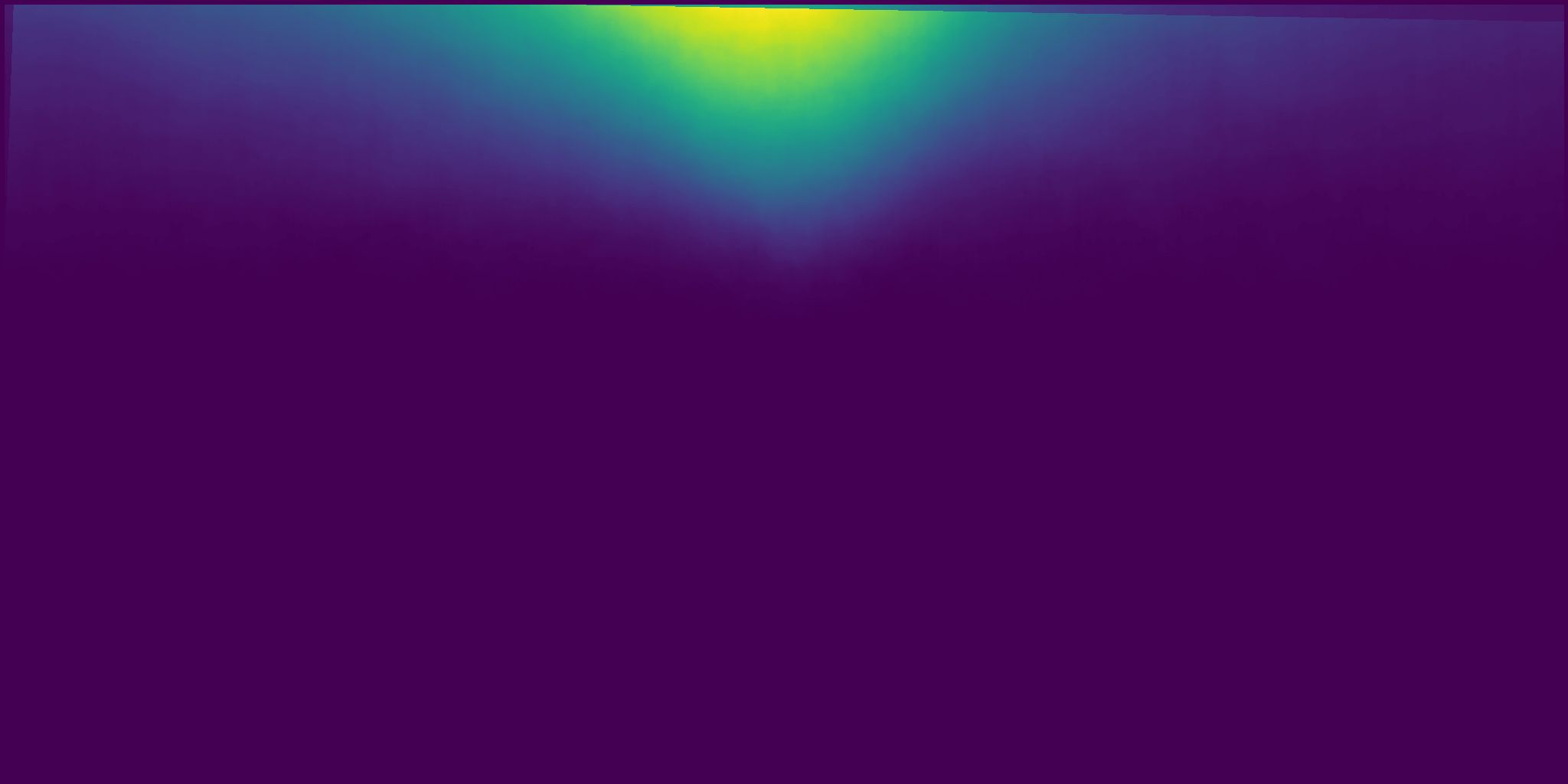} & \\
        &
        \multicolumn{2}{c}{\textit{Traffic Signs}} & \multicolumn{2}{c}{\textit{Building}} & \multicolumn{2}{c}{\textit{Sky}} & \\
    \end{tabular}
    \caption{\rev{Spatial Priors Similarities Across Domains. Considered the semantic segmentation task, we compute the number of occurrences of each class at each pixel location for both domains. Domain \domainA{} is CARLA, \domainB{} is Cityscapes. We visualize the occurrence maps with a \textit{viridis} colormap.}}
    \label{fig:sp_priors}
\end{figure*}

The DA literature presents us with several ways to accomplish this.
One may operate on the input space \cite{Bousmalis_2017}, on the feature space \cite{tzeng2017adversarial} or on the output space of the network \cite{Tsai_2018}.
In our setting, though, both input and output space of \trnet{} are high dimensional latent spaces and, as reported in \cite{Tsai_2018}, unsupervised domain adaptation techniques tend to fail when applied to such spaces while addressing  dense tasks.
Yet, we can address the domain shift issue with a direct approach in the input space of \trnet{}, \ie, the feature space of $N_1$, which is already shared between \domainA{} and \domainB{} due to the network being trained supervisedly with images from both domains.
\rev{
We leverage on the intuition that scene spatial priors are typically domain invariant in many adaptation scenarios. We consider it as a reasonable assumption for several domain adaptation settings, where we select the source domain by considering visual similarities with the target domain.
For instance, in autonomous driving scenarios we typically have cameras placed from a car viewpoint, and scenes are urban scenarios in both synthetic \cite{Dosovitskiy17, Richter_2017} and real \cite{Cordts_2016_CVPR, yu2020bdd100k, Geiger2013IJRR} datasets.
Thus, if we consider the task of semantic segmentation in all datasets (synthetic and real) we typically find \textit{road} pixels in the bottom part of the images and instead sky pixels in the top part of the images.
To visualize this property we select a synthetic domain \domainA{} CARLA \cite{Dosovitskiy17} and a real domain \domainB{} Cityscapes \cite{Cordts_2016_CVPR}. Then, we count for each pixel location the number of occurrences of each class. We show the result of this experiment in \autoref{fig:sp_priors}, using a \textit{viridis} colormap to display these occurrency maps for each class and for both domains \domainA{} and \domainB{}. We can clearly see that the maps have a structure similar across domains, e.g., building are concentrated in the top image regions.

Leveraging this property, we propose to align more closely the features computed by $E_1$ on the images from both domains, \ie, $f_1^A$ and $f_1^B$, by enforcing similarity of the  $L_2$ norms across channels at the same spatial location.}
Starting from features $f_1^A$ and $f_1^B$ of dimensionality $H\times W \times C$, where $H$, $W$ and $C$ are the height, width and number of channels of the feature maps, we calculate the $L_2$ norm along the $C$ axis and minimize the absolute difference at each spatial location $i,j$. Hence, our NDA (Norm Discrepancy Alignment) Loss is defined as follows:

\begin{equation}
    \mathcal{L}_{NDA} = \frac{1}{W\times H} \sum_{i=1}^H\sum_{j=1}^W \left | \lVert f^A_{1_{i,j}} \rVert_2  - \lVert f^B_{1_{i,j}} \rVert_2 \right |
\end{equation}

\subsection{Feature Alignment Across Tasks}\label{sec:transfer}
While the NDA loss presented above aims at improving the generalization across domains of the feature mapping network \trnet{}, its effectiveness can be further improved by aligning features also across tasks.
Accordingly, we conjecture that $f_1$ features should capture as much information as possible on the details of the scene, even though some of this information may not be necessary to solve \taskone{}, because, when transferred by  \trnet{}, such a richer representation could help to solve \tasktwo{} more effectively.
For this reason, while training $N_1$ for \taskone{}, we  train jointly an additional decoder, $D_{aux}$, to solve an auxiliary task, \taskaux{}, aimed at enriching the learnt representation $f_1$ . 
However, though multi-task learning of \taskone{} and \taskaux{} could help to encode more detailed information into $f_1$ features, it does not guarantee that the decoder $D_2$, used at inference time on the features $f_{1 \rightarrow 2}$ transferred from $\mathcal{T}_1$ to $\mathcal{T}_2$, may effectively deploy this additional information if it has been trained only to solve \tasktwo{} in isolation.
This leads us to reckon that $D_{aux}$ should be trained  jointly with $N_2$ too, such that the additional information required to solve \taskaux{} may be incorporated also within the features $f_2$ learnt by $E_2$.

Therefore, given auxiliary task labels $y^A_{aux}$ and $y^B_{aux}$ for \domainA{} and \domainB{}, we train $N_1$ and $N_2$ jointly with a single auxiliary decoder $D_{aux}$ using an auxiliary loss $\mathcal{L}_{aux}$.
Purposely, we obtain auxiliary predictions from both encoders with the shared decoder $D_{aux}$ as  $\hat{y}_{k_{aux}}=D_{aux}(E_k(x))$, $k \in \{1,2\}$. 
Similarly to the simpler formulation of our framework presented in \autoref{subsec:atdt}, to compute the auxiliary loss we feed images of both domains through $E_1$, while we pass only images from \domainA{} through $E_2$.
We do not pass images belonging to \domainB{} through $E_2$ while training $D_{aux}$ since this would be the only kind of supervision for $E_2$ in \domainB{} and it may skew $E_2$ output to be more effective on \taskaux{} than on \tasktwo{}.

\rev{
\subsection{Overall $N_1$ and $N_2$ loss}
When training simultaneously $N_1$ and $N_2$, the overall loss is:

\begin{equation}
\begin{split}
    \mathcal{L} =
\lambda_{T_1}\mathcal{L}_{T_1}(y_1^A, \hat{y}_1^A) + \lambda_{T_1}\mathcal{L}_{T_1}(y_1^B, \hat{y}_1^B)  \\
+ \lambda_{T_2}\mathcal{L}_{T_2}(y_2^A, \hat{y}_2^A) + \lambda_{aux}\mathcal{L}_{aux}(y_{1_{aux}}^A, \hat{y}_{1_{aux}}^A) +  \\ \lambda_{aux}\mathcal{L}_{aux}(y_{1_{aux}}^B, \hat{y}_{1_{aux}}^B) +  \lambda_{aux}\mathcal{L}_{aux}(y_{2_{aux}}^A, \hat{y}_{2_{aux}}^A) +  \\
\lambda_{NDA}\mathcal{L}_{NDA}(f_1^A, f_1^B) 
\end{split}
\end{equation}
}

\section{Experimental Settings}\label{sec:architecture}

\textbf{Tasks.}
We fix \taskone{} and \tasktwo{} to be monocular depth estimation and semantic segmentation, or vice versa. These two visual tasks can be addressed using the same encoder-decoder architecture, with changes needed only in  the final layer. Semantic segmentation is solved by minimizing a pixel-wise cross entropy loss, monocular depth estimation by minimizing an $L_1$ loss. 
We select edge detection as our \taskaux{} since it seems particularly amenable to improve the effectiveness of our framework in capturing and transferring  important structural information that might otherwise be lost. Let us consider the case of \taskone{} being depth estimation and \tasktwo{} semantic segmentation. The features $f_1$ needed to compute depth may ignore the boundaries between semantically distinct regions showing up at the same distance from the camera: in \autoref{fig:teaser} (left) this is the case, \eg, of the boundaries between legs or tyres and ground, as well as  between street signs and poles. Therefore, even if fed to a perfect \trnet{}, $f_1$ may not contain all the information needed to restore the semantic structure of the image. By solving jointly edge detection on the input image, instead, we force our $N_1$  network to extract additional information that would not need to be captured should the learning objective be concerned  with depth estimation only. Similarly, \autoref{fig:teaser} (right) highlights how depth discontinuities do not necessarily correspond to semantic boundaries, such that a network $N_1$ trained in isolation to assign semantic labels to pixels may not need to learn information relevant to estimate the depth structure of the image. 
Besides, it is worth pointing out that edge detection can be solved using again the same decoder architecture as  \taskone{} and \tasktwo{}. Since the edge proxy-labels that we adopt are gray-scale images \cite{soria2020dexined}, in our experiments we implement the $\mathcal{L}_{aux}$ loss introduced in \autoref{sec:transfer} as a standard $L_2$ loss.
\rev{In all our experiments we set $\lambda_{aux}$ to 0.5,  $\lambda_{NDA}$ to 0.001, $\lambda_{T_1}$ and $\lambda_{T_2}$ to 1 to balance loss values.}
\\

\textbf{Datasets.}
We test the effectiveness of our method in an autonomous driving scenario.
We set \domainA{} and \domainB{} to be a synthetic and a real dataset, respectively. The former consists of a collection of images generated with the Carla simulator \cite{Dosovitskiy17}, while the latter is the popular Cityscapes dataset \cite{Cordts_2016_CVPR}. We generated the Carla dataset mimicking the camera settings of the real scenes. We render 3500, 500, and 1000 images for training, validation, and testing, respectively. For each image, we store the associated depth and semantic labels provided by the simulator. The Cityscapes dataset is a collection of 2975 and 500 images to be used for training and validation, respectively. As for our evaluation, we use the 500 Cityscapes validation images since test images are not equipped with labels. Moreover, as in  Cityscapes only the semantic labels are provided, we use depth proxy-labels obtained with the SGM stereo algorithm \cite{hirschmuller2005accurate}, by filtering the erroneous predictions in the generated disparities with a left-right consistency check. This can be considered as an added value because it shows the ability to transfer knowledge when learning from noisy labels.
Finally, we use a pre-trained\footnote{Neither \domainA{} nor \domainB{} belong to the training set of this network.} state-of-the-art neural network\cite{soria2020dexined}  as an off-the-shelf edge detector to extract from the images belonging to \domainA{} and \domainB{} the edges used as proxy-labels to train \taskaux{}.\\

\textbf{Architecture.}
To solve each task, we use two dilated ResNet50 \cite{Yu2017} as encoder and a stack of bilinear upsample plus convolutional layers as decoder. The encoder shrinks both input dimensions with a factor of 1/16, while the decoder upsamples the feature map until a prediction with the same spatial resolution as the input image is obtained. The two networks for \taskone{} and \tasktwo{} are identical, but for the final prediction layer, which is task dependent. The two previously defined encoders are also used to capture good features for edge detection, which is solved using $D_{aux}$, that shares the same architecture as the decoders used in $N_1$ and $N_2$. 
\trnet{} is a simple CNN made out of 6 pairs of convolutional and batch normalization layers with kernel size $3\times3$ which do not perform any downsampling or upsampling operation. \\

\textbf{Training and Evaluation Protocol.}
During the training phase of the transfer network \trnet{}, the model is evaluated on the validation set of Carla. Of course, it is possible that optimality on Carla does not translate into optimal performance on Cityscapes. Yet, we cannot use data from the target domain neither for hyper-parameters tuning nor for early stopping, because in our setting these data would not be available in any real  scenario. Therefore, the Cityscapes validation set is only used at test time to measure the final performances of our framework method.

\begin{figure}[!htbp]
	\centering
	\includegraphics[width=0.45\textwidth]{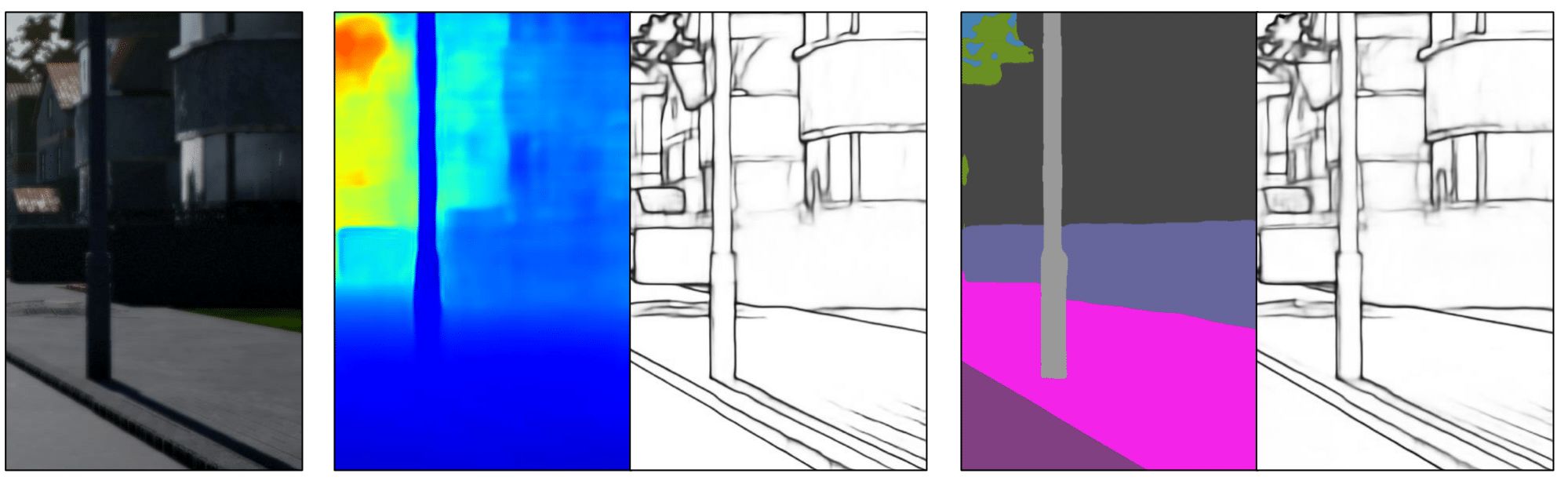}
	\caption{From left to right: RGB input image of domain \domainA{} , depth prediction from $N_1$, edges from $f_1$, semantic segmentation from $N_2$ and edges from $f_2$. Task features $f_1$ and $f_2$ encode richer details than strictly needed to solve either tasks as we can recover all edges from both of them by $D_{aux}$.}
	\label{fig:edge_details} 
\end{figure}

\begin{table*}[!htbp]
    \center
    \caption{Experimental results of \depsem{} scenario. Baseline stands for $N_2$ trained on \domainA{} and tested on \domainB{}, Transfer Oracle represents \trnet{} trained only on \domainB{}, Oracle refers to $N_2$ trained and tested on \domainB{}. Best results highlighted in bold.}
	\label{tab:depth2sem}
	\setlength{\tabcolsep}{2.5pt}
	\scalebox{1}{
	\begin{tabular}{cc|c|ccccccccccc|cc}
		\toprule
		\domainA{} & \domainB{} & Method & \rotatebox{90}{Road} & \rotatebox{90}{Sidewalk} & \rotatebox{90}{Walls} & \rotatebox{90}{Fence} & \rotatebox{90}{Person} & \rotatebox{90}{Poles} & \rotatebox{90}{Vegetation} & \rotatebox{90}{Vehicles} & \rotatebox{90}{Tr. Signs} & \rotatebox{90}{Building}  & \rotatebox{90}{Sky} & \textbf{mIoU} & \textbf{Acc} \\
		\midrule 
		\carla{} & CS & Baseline     & 78.99 & 38.81 & 1.34 & 5.80  & 24.02 & 24.47 & 71.98 & 52.23 & 5.57  & 65.17 & 59.10 & 38.86 & 78.58 \\
		
	    \carla{} & CS & ZDDA \cite{peng2018zero} & 
	    85.93 & 41.28 & 4.62 & 8.63 & 38.80 & 25.94 & 72.78 & 58.37 & 18.44 & 73.74 & \textbf{78.16} & 46.06 & 82.82 \\
		\carla{} & CS & \textbf{\atdt{}}    & \textbf{90.57}          & \textbf{48.46}          & \textbf{7.37} & \textbf{12.27}          & \textbf{41.16} & \textbf{31.90}          & \textbf{81.96}          & \textbf{72.77}          & \textbf{23.44}          & \textbf{77.85}          & 76.33       & \textbf{51.28} & \textbf{87.57} \\ 
		\midrule
		CS & CS & Transfer Oracle     & 89.69 & 48.05 & 11.46& 29.58 & 59.68 & 35.84 & 85.83 & 85.57 & 34.03 & 78.17 & 85.54 & 58.50 & 88.84\\
		- & CS & Oracle                          & 96.74 & 78.28 & 29.26& 40.78 & 72.39 & 51.28 & 90.69 & 91.94 & 58.92 & 86.33 & 89.23 & 71.44 & 93.90\\
		\bottomrule
	\end{tabular}
	}
\end{table*}

\begin{figure*}[!htbp]
	\centering
	\includegraphics[width=0.9\textwidth]{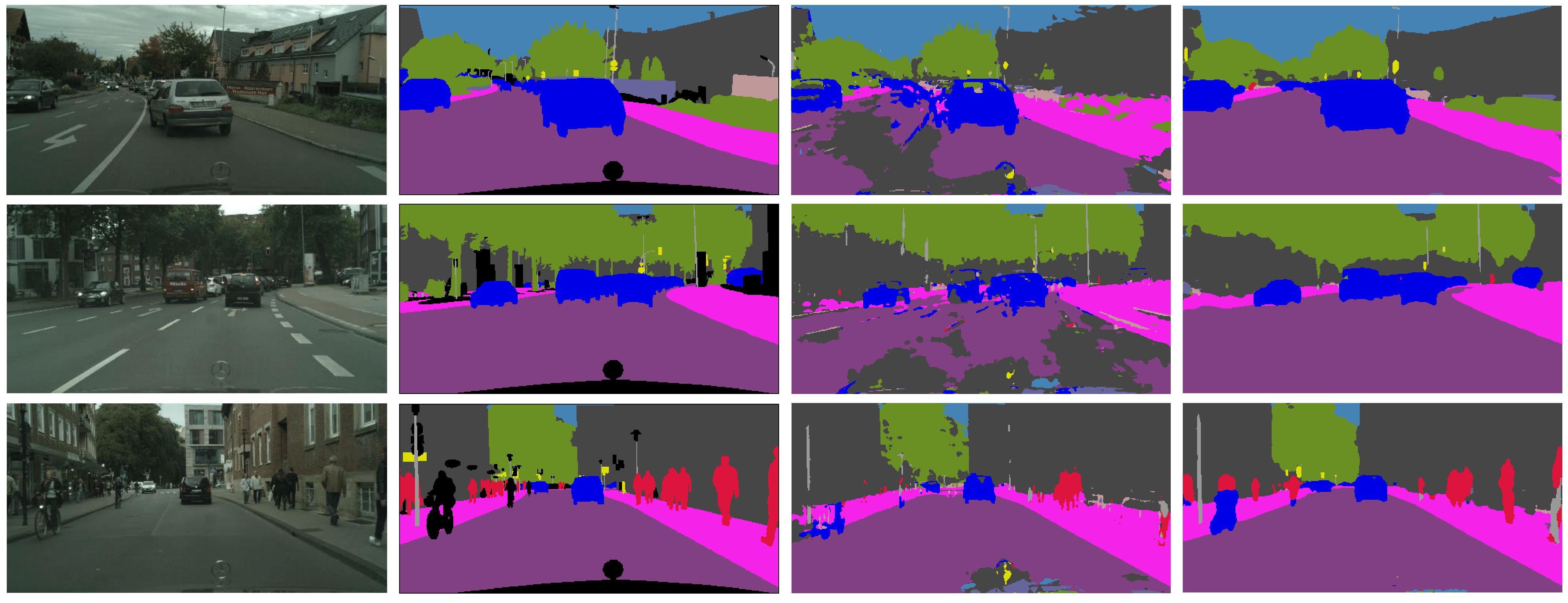}
	\caption{Qualitative results of the \depsem{} scenario. From left to right: RGB image, ground-truth, baseline trained only on domain \domainA{}, ours.}
	\label{fig:qualitatives_depsem}
\end{figure*}

\textbf{Metrics.}
To evaluate the performance on the semantic segmentation task two metrics are used: pixel accuracy, shortened \emph{Acc.} (i.e the percentage of pixels with a correct predicted label) and Mean Intersection Over Union, shortened \emph{mIoU}, as defined in  \cite{Cordts_2016_CVPR}. To render these metrics comparable among the used datasets, we solve semantic segmentation on the 10 shared classes (Road, Sidewalk, Walls, Fence, Person, Poles, Vegetation, Vehicles, Traffic Signs, Building) plus the Sky category, which is defined as the set of points with infinite depth. Some of the Cityscapes classes are collapsed into one class: car and bicycle into vehicle, traffic signs and traffic light into traffic sign. The remaining categories of Cityscapes are instead ignored.\\
When testing the depth estimation task, we report the standard metrics described in \cite{eigen2014depth}: Absolute Relative Error (Abs Rel), Square Relative Error (Sq Rel), Root Mean Square Error
(RMSE), logarithmic RMSE and $\delta_1$, $\delta_2$ and $\delta_3$ accuracy scores. Each $\delta_\alpha$ is obtained by computing, for each pixel of the input image, the maximum among ratio and inverse ratio between the predicted value and the ground-truth. $\delta_\alpha$ represents the percentage of pixels whose such ratio is lower than $1.25^\alpha$.

\section{Experimental Results}
We provide results for two different settings: transferring features from depth estimation to semantic segmentation (\autoref{sec:depsem}) as well as from semantic segmentation to depth estimation (\autoref{sec:semdep}).

In both scenarios, as already mentioned, we used edge detection as auxiliary task, motivated by the idea that either semantic segmentation and depth estimation can benefit from edge information. \autoref{fig:edge_details} shows that with our multi-task learning protocol we are able to restore all the details of the scene from both $f_1$ and $f_2$, proving that $N_1$ and $N_2$ have indeed learned to encode into their features richer information than that strictly needed to solve \taskone{} and \tasktwo{}.

\subsection{Depth to Semantics}\label{sec:depsem}
In this setup, denoted as \depsem{}, the goal of our framework is to transform depth features into semantic segmentation features. This mapping is learned using \carla{} as domain \domainA{} and \cityscapes{} as domain \domainB{}.
We report results in \autoref{tab:depth2sem}: the first row shows results obtained with no adaptation (\ie, training $N_2$ on \carla{} and testing it directly on \cityscapes{}), while from the second row we can see that our final framework yields 51.28\% mIoU and 87.57\% Acc with an improvement of +12.48\% and +8.99\%  wrt to the baseline.

\rev{Even though \atdt{} is the first work to address the across tasks and domains scenario, we compare it against a related work, ZDDA \cite{peng2018zero}, which also leverage auxiliary data from a different tasks to perform domain adaptation. We apply it in our setup using as the "Source" and "Target" domains \carla{} and \cityscapes{} respectively. We address the \depsem{} scenario using depth maps as "task-irrelevant" data. We skip the last sensor fusion step (Step 3) because it was not applicable in our scenario since we do not have task-irrelevant data at test time, and thus we stop training after the adaptation step (Step 2).
We report results of this alternative approach in the second row of \autoref{tab:depth2sem}.
As we can notice, ZDDA is effective in our scenario and achieves better performance compared to the baseline. However, \atdt{} obtains much better results, surpassing ZDDA in all metrics. This is not surprising since ZDDA focus on extracting features only from task-irrelevant data, which can be sub-optimal for the relevant task as these data do not provide the same amount of information as the task-relevant data, e.g., features extracted only from depth images would not contain several useful information for semantic segmentation such as colors or textures.
}

Furthermore, as we are transferring features from another task, it is worth trying to investigate on the upper bound in performance due to the inherent transferability of the features between the two tasks. Purposely,  we train \trnet{} using only Cityscapes to learn a mapping function in a supervised fashion as explained in \autoref{sec:transfer} on \domainB{} and test on the validation set of \domainB{}. These results are shown in the third row of the table (denoted as Transfer Oracle): given a transfer architecture, there seems to be an upper bound in performance due to the nature of the two tasks, which in the considered setting amounts to a 58.5\% mIoU. Thus, our proposal exhibit a gap wrt the Transfer Oracle that is only about -7.2\% mIoU. We also report the performance of $N_2$ trained on \domainB{} and tested on \domainB{}, \ie, the absolute upper bound in performance (last row of the table, denoted as Oracle).

Some qualitative results dealing with the \depsem{} scenario are depicted in \autoref{fig:qualitatives_depsem}. It is possible to appreciate the overall improvement of our method wrt the baseline, either in flat areas (\eg, roads, sidelwalks and walls), objects shapes (\eg, cars and persons) and fine-grained details (\eg, poles and traffic signs).

\subsection{Semantics to Depth}\label{sec:semdep}

\begin{figure*}[!htbp]
	\centering
	\includegraphics[width=0.9\textwidth]{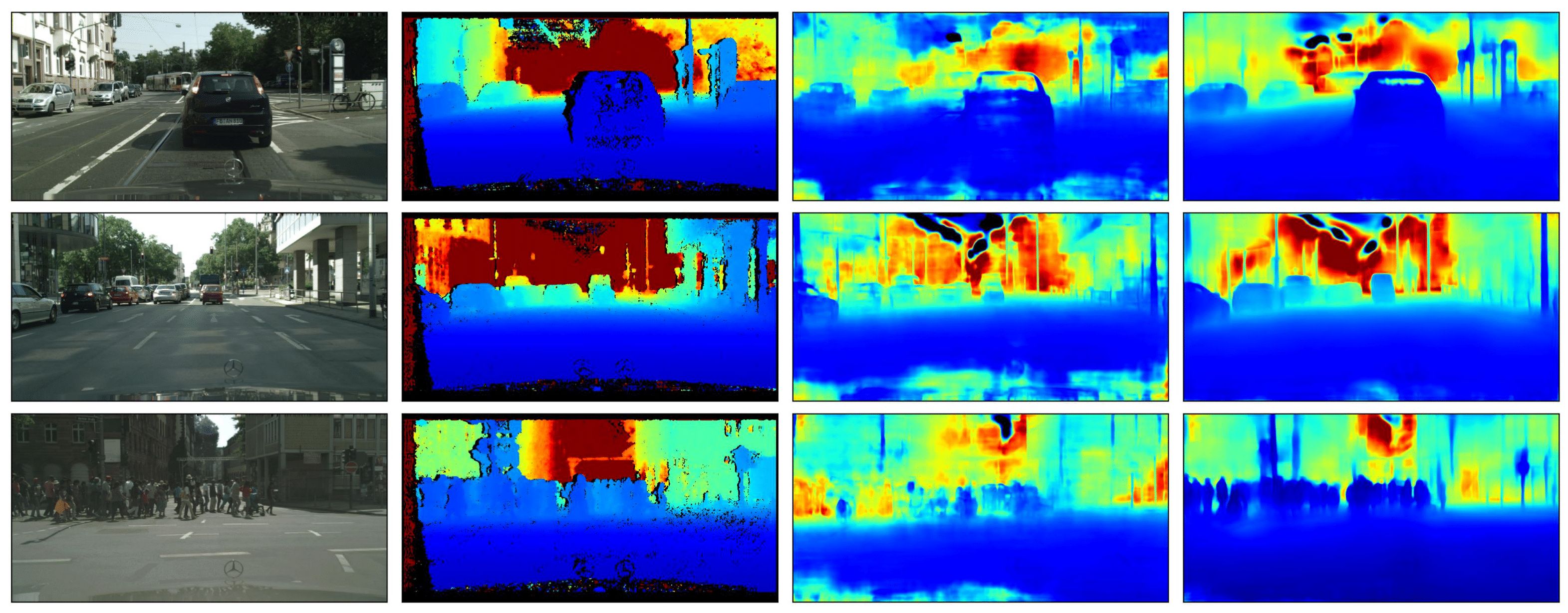}
	\caption{Qualitative result of the \semdep{} scenario. From left to right: RGB image, ground-truth, baseline network trained only on domain \domainA{}, ours.}
	\label{fig:qualitatives_semdep}
\end{figure*}

In this setup, which we define as \semdep{}, the goal of our framework is to transform semantic features into depth features. This mapping is learned using \carla{} as domain \domainA{} and \cityscapes{} as domain \domainB{}, as done for the \depsem{} scenario.
Results are reported in \autoref{tab:sem2depth}. Similarly to the \depsem{} scenario, in the first row we show results with no adaptation (\ie, our baseline), while the third row presents the ones obtained with our framework. 
\rev{Also for this setup we report performances of ZDDA \cite{peng2018zero} (second row), in which we use semantic maps as task-irrelevant data. We can see that ZDDA achieves slight better performance of the baseline in 5 metrics out of 7, but still inferior to our approach.
}
Moreover, we report  results from the Transfer Oracle and the Oracle, implemented as described for the \depsem{} scenario. It is possible to appreciate that our framework outperforms the baseline on 6 out of 7 metrics, closing remarkably the gap with the practical upper bound of the Transfer Oracle.
In \autoref{fig:qualitatives_semdep}, we show some qualitative results of the \semdep{} scenario. While predictions look quite noisy in the background, we can see a good improvement in the foreground area thanks to our method. Shapes are recovered almost perfectly, both for big and small objects, even with difficult subjects like the crowd in the bottom row. It is also worth pointing out that the depth predictions yielded by our method turn out much smoother than the ones produced by the baseline and generally less noisy than the ground-truth that, as explained in \autoref{sec:architecture}, consists of proxy-labels computed with SGM \cite{hirschmuller2005accurate}. 

\begin{table*}[!htbp]
	\center
	\caption{Experimental results of \semdep{} scenario. Baseline stands for $N_2$ trained on \domainA{} and tested on \domainB{}, Transfer Oracle represents \trnet{} trained only on \domainB{}, Oracle refers to $N_2$ trained and tested on \domainB{}. Best results highlighted in bold.}
	\setlength{\tabcolsep}{2.5pt}
	\scalebox{1}{
	\begin{tabular}{cc|c|cccc|ccc}
		\toprule
		\multicolumn{3}{c|}{} & \multicolumn{4}{c|}{\cellcolor{blue!25}Lower is better}& \multicolumn{3}{c}{\cellcolor{LightCyan}Higher is better}\\
		\domainA{} & \domainB{} & Method & \cellcolor{blue!25}Abs Rel & \cellcolor{blue!25}Sq Rel & \cellcolor{blue!25}RMSE & \cellcolor{blue!25}RMSE log & \cellcolor{LightCyan}$\delta_1$ & \cellcolor{LightCyan}$\delta_2$ & \cellcolor{LightCyan}$\delta_3$\\
		\midrule
		\carla{} & CS & Baseline & 0.7398 & 15.169 & 14.774 & 0.641 & \textbf{0.406} & 0.650 & 0.781 \\
		
		\carla{} & CS & ZDDA \cite{peng2018zero} & 0.5206 & 7.5491 & 13.347 & 0.633 & 0.345 & 0.638 & 0.858 \\
		\carla{} & CS & \textbf{\atdt{}} & \textbf{0.3928} & \textbf{4.9094} & \textbf{12.363} & \textbf{0.444} & 0.372 & \textbf{0.757} & \textbf{0.923} \\
		
		\midrule
		CS & CS & Transfer Oracle & 0.2210 & 2.2962 & 9.032 & 0.275 & 0.669 & 0.914 & 0.972 \\
		- & CS & Oracle & 0.1372 & 1.6214 & 8.566 & 0.244 & 0.816 & 0.938 & 0.976 \\
		\bottomrule
	\end{tabular}
	}
	\label{tab:sem2depth}
\end{table*}

\section{Ablation Studies}\label{sec:ablations}
In the following sections, we study the effectiveness of the key design choices behind our proposal.
 
\begin{table*}[!htbp]
	\center
	\small
	\caption{Ablation study in the \depsem{} scenario. Best results highlighted in bold. Aux refers to the framework trained with the auxiliary task. NDA refers to the framework trained with our NDA loss.}
	\label{tab:depth2sem_ablation}
	\setlength{\tabcolsep}{2.5pt}
	\scalebox{1}{
	\begin{tabular}{cc|cc|ccccccccccc|cc}
		\toprule
		\domainA{} & \domainB{} & \rotatebox{90}{Aux} & \rotatebox{90}{NDA} & \rotatebox{90}{Road} & \rotatebox{90}{Sidewalk} & \rotatebox{90}{Walls} & \rotatebox{90}{Fence} & \rotatebox{90}{Person} & \rotatebox{90}{Poles} & \rotatebox{90}{Vegetation} & \rotatebox{90}{Vehicles} & \rotatebox{90}{Tr. Signs} & \rotatebox{90}{Building}  & \rotatebox{90}{Sky} & \textbf{mIoU} & \textbf{Acc} \\
		\midrule 
		\carla{} & CS & &                 & 89.95           & 46.77         & 5.16          & 10.21          & 28.93          & 28.92          & 77.50          & 71.37          & 19.24          & 75.29          & 75.12          & 48.04 & 85.90 \\
		\carla{} & CS & \checkmark &                 & 90.12           & 48.90         & 4.18          & 11.63          & 37.40          & 31.98          & \textbf{82.34} & 71.50          & 15.11          & \textbf{78.04} & \textbf{80.61} & 50.16 & 87.21 \\
		\carla{} & CS & & \checkmark & \textbf{91.21} & \textbf{50.16} & 5.14          & \textbf{13.78} & 36.99          & \textbf{32.10} & 77.72          & \textbf{73.38} & \textbf{23.47} & 76.67          & 72.67          & 50.30 & 86.77 \\
		\carla{} & CS & \checkmark & \checkmark    & 90.57          & 48.46          & \textbf{7.37} & 12.27          & \textbf{41.16} & 31.90          & 81.96          & 72.77          & 23.44          & 77.85          & 76.33          & \textbf{51.28} & \textbf{87.57} \\ 
		\bottomrule
	\end{tabular}
	}
\end{table*}

\subsection{Contribution of \taskaux{} and NDA Loss}
We start by studying the effect of introducing in our framework the auxiliary task and the NDA loss, analyzing their contribution when used separately as well as when combined together. The second and  third row of \autoref{tab:depth2sem_ablation} report the results obtained in the \depsem{} setting by integrating in our method either the auxiliary task (\ie, edge detection) or the NDA loss, respectively. We can see that both design choices bring in an improvement of about +2\% in terms of mIoU with respect to the base \atdt{} framework (first row). Moreover, the last row of the table shows that the auxiliary edge detection task and the NDA loss turn out complementary because, when combined together, they can provide an overall improvement of +3.34\% mIoU.

\begin{figure*}[!htbp]
	\centering
	\includegraphics[width=0.9\textwidth]{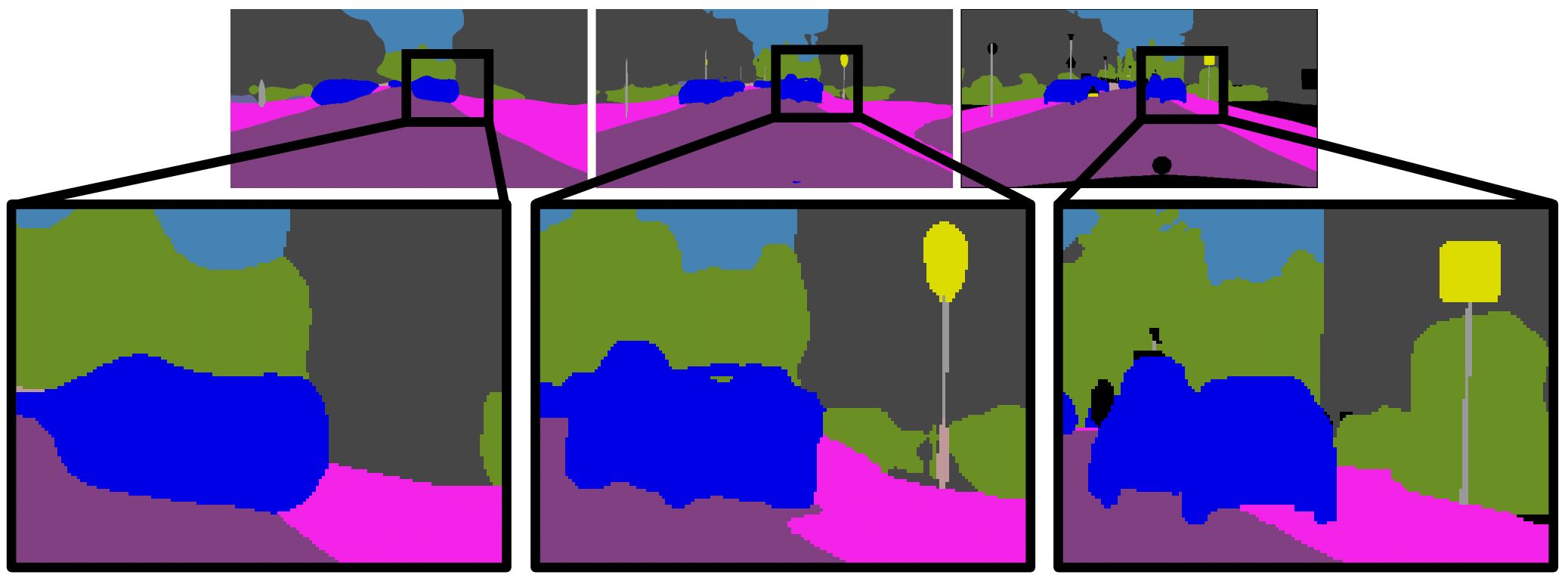}
	\caption{Zoomed results in a \depsem{} scenario. From left to right: base \atdt{} without edge and NDA, our proposed method, ground-truth. We notice how, unlike base \atdt{}, our method is able to recover the fine-grained details of the scene.
	}
	\label{fig:structure_details}
\end{figure*}

\autoref{fig:structure_details} presents some zoomed-in qualitative results: we can see that, even if the base version of \atdt{} already produces satisfactory results at a coarse level, the complete version of our framework can produce much more accurate predictions, especially regarding small details, such as poles, traffic signs and car outlines.

\subsection{Effectiveness of edge detection as auxiliary task}
\begin{table*}[t]
    \caption{Comparison between autoencoder and edge detection as auxiliary tasks in the \depsem{} scenario. Best results highlighted in bold.}
	\label{tab:auxtask}
    \center
	\setlength{\tabcolsep}{2.5pt}
	\scalebox{1}{
	\begin{tabular}{l|ccccccccccc|cc}
		\toprule
		\taskaux{} &\rotatebox{90}{Road} & \rotatebox{90}{Sidewalk} & \rotatebox{90}{Walls} & \rotatebox{90}{Fence} & \rotatebox{90}{Person} & \rotatebox{90}{Poles} & \rotatebox{90}{Vegetation} & \rotatebox{90}{Vehicles} & \rotatebox{90}{Tr. Signs} & \rotatebox{90}{Building}  & \rotatebox{90}{Sky} & \textbf{mIoU} & \textbf{Acc} \\
		\midrule 
		None & 89.95 & 46.77         & 5.16          & 10.21          & 28.93          & 28.92          & 77.50          & 71.37          & \textbf{19.24}          & 75.29          & 75.12          & 48.04 & 85.90 \\
		Autoencoder & \textbf{90.68} & \textbf{50.12} & \textbf{7.45} & 9.08 & 31.40 & 29.43 & 78.72 & 68.51 & 12.95 & 74.67 & 75.68 & 48.07 & 86.31 \\
		Edge detection & 90.12 & 48.90 & 4.18 & \textbf{11.63} & \textbf{37.40} & \textbf{31.98} & \textbf{82.34} & \textbf{71.50} & 15.11 & \textbf{78.04} & \textbf{80.61} & \textbf{50.16} & \textbf{87.21} \\
		\bottomrule
	\end{tabular}
	}
\end{table*}

In this section, we show empirically that in our framework the choice of the proper auxiliary task is key to performance. 

In both the \depsem{} and the \semdep{} scenarios, we propose to use edge detection as auxiliary task because it captures information about the shapes of the objects in the input images and allows for straightforward computation of proxy-labels. To validate this design choice, we tested our framework in the \depsem{} setting, using  $D_{aux}$ to reconstruct  the input images both from $f_1$ and $f_2$, \ie,  the classical autoencoder setting (results in \autoref{tab:auxtask}).
Interestingly, using image reconstruction as auxiliary task results in an mIoU score almost identical to the base \atdt{}.
We consider that the autoencoder task is guided by a reconstruction loss which makes no distinction between the pixels of the input image: such supervision cannot guide effectively $f_1$ and $f_2$ to encapsulate the high-frequency components of the image that are needed to predict the fine-grained details of the scene, which is instead obtained by adopting edge detection as auxiliary task.

\begin{table*}[t]
    \caption{\rev{Auxiliary tasks as source tasks in the \depsem{} scenario. Best results highlighted in bold.}}
	\label{tab:aux_as_source}
    \center
	\setlength{\tabcolsep}{2.5pt}
	\scalebox{1}{
	\begin{tabular}{l|ccccccccccc|cc}
		\toprule
		
		\taskaux{} as \taskone{} &\rotatebox{90}{Road} & \rotatebox{90}{Sidewalk} & \rotatebox{90}{Walls} & \rotatebox{90}{Fence} & \rotatebox{90}{Person} & \rotatebox{90}{Poles} & \rotatebox{90}{Vegetation} & \rotatebox{90}{Vehicles} & \rotatebox{90}{Tr. Signs} & \rotatebox{90}{Building}  & \rotatebox{90}{Sky} & \textbf{mIoU} & \textbf{Acc} \\
		\midrule
		
		Autoencoder & 60.24 & 19.33 & 1.67 & 1.67 & 4.12 & 8.00	& 33.15 & 10.49 & 0.69 & 17.89  & 62.66 & 19.99 & 52.91 \\
		
		Edge Detection & 63.82 & 16.60 & 0.67 & 1.37 & 6.55 & 10.26 & 47.62 & 4.42	& 0.11 & 33.90 & 38.87 & 20.38 & 58.33 \\
		
		\midrule
		
		Depth & \textbf{89.95} & \textbf{46.77} & \textbf{5.16} & \textbf{10.21} & \textbf{28.93} & \textbf{28.92} & \textbf{77.50} & \textbf{71.37} & \textbf{19.24} & \textbf{75.29} & \textbf{75.12} & \textbf{48.04} & \textbf{85.90} \\
		\bottomrule
	\end{tabular}
	}
\end{table*}

\rev{
\subsection{Auxiliary tasks as source tasks}
The main difference between a source and an auxiliary task is that the auxiliary task alone cannot provide enough information to solve \tasktwo{}, but it is useful to enrich \taskone{} features and align feature content across tasks.
To better support our claims, we investigated \atdt{} behaviour when using auxiliary tasks \taskaux{} as source tasks \taskone{} and semantic segmentation as target task \tasktwo{}. The results of these experiments are reported in \autoref{tab:aux_as_source}.
All rows of the table show results of the \emph{base} \atdt{} i.e., trained without $L_{aux}$ and $L_{NDA}$ losses.
As we can notice, using as source task \taskone{} a standard image-reconstruction (row 1, autoencoder) or an edge detection (row 2) lead to much worse results than using depth estimation (row 3). 
We argue that features extracted by $N_1$ for these tasks do not contain enough information to perform semantic segmentation, which are yet contained in features for depth estimation. 
Similar finding were also made by Taskonomy \cite{zamir2018taskonomy}, in which they show that edge detection and image reconstruction (aka autoencoder) are less correlated to semantic segmentation than depth estimation.
On the contrary, we have shown that Edge Detection can be a good auxiliary task in the \depsem{} scenario since it can enrich depth features with missing edges useful for semantic segmentation and it can increase transferability aligning depth and semantic features.
}

\subsection{Importance of simultaneous training of $N_1$, $N_2$ and $D_{aux}$}
In our experiments we use edge detection as auxiliary task and train a shared decoder $D_{aux}$ to reconstruct the edges of the input image from the features extracted by both $E_1$ and $E_2$. In fact, we argue that this procedure should force $E_1$ to encode into the extracted features also edge information that may be not necessary to solve \taskone{} but that may be relevant for \tasktwo{}. 
Besides, we believe that simultaneous training of $N_1$, $N_2$ and $D_{aux}$ is crucial to encourage features coming from $E_1$ and $E_2$ to represent edge information in a similar manner, making it easier to learn \trnet{}.

In \autoref{tab:edgeablation} we report the results concerning the ablation study conducted to validate these intuitions. We consider the \depsem{} scenario using the Carla dataset as domain \domainA{} and Cityscapes as domain \domainB{}. The four rows of the table deal with the following training schemes:

\begin{enumerate}
    \item The \emph{base} AT/DT (\ie, without \taskaux{} and NDA loss) as  baseline.
    \item We first train $N_1$ and $D_{aux}$ on both \domainA{} and \domainB{}. Then, we train $N_2$ on \domainA{}. Finally, we train \trnet{} on features extracted by $E_1$ and $E_2$ on domain \domainA{}.
    \item We train $N_1$ and a first $D_{aux}^1$ on both \domainA{} and \domainB{}. Then, we train $N_2$ and a second $D_{aux}^2$ on \domainA{}. Finally, we train \trnet{} on features extracted by $E_1$ and $E_2$ on domain \domainA{}
    \item Our proposed method, which trains  $N_1$, $N_2$ and a shared $D_{aux}$  simultaneously.
\end{enumerate}

The introduction of edge detection as auxiliary task helps in every scenario. In fact, if we use $D_{aux}$ only while training $N_1$ (second row), we already see an increase of $0.6\%$ in the overall mIoU. We believe that this is explained by the presence of edge details (not strictly necessary to solve \taskone{} but relevant for \tasktwo{}) in the features extracted by $E_1$. However, \trnet{} may experience difficulties in adapting $f_1$ into $f_2$ if edge information is not explicitly present in $f_2$. This is confirmed by the results in the third row of the table, where an additional increase of $1.3\%$ in the overall mIoU is attained by using two different $D_{aux}$ (one during training of $N_1$ and one during training of $N_2$).
Finally, the best results in terms of mIoU and Acc are achieved by our method, \ie, when training $N_1$, $N_2$ and a shared $D_{aux}$ simultaneously. This vouches for the benefit of encoding in a  similar manner the edge information in $f_1$ and $f_2$ in order to enforce feature alignment across tasks.

\begin{table*}[t]
	\center
	\caption{Ablation study on the importance of simultaneous training of the \taskone{}, \tasktwo{}, and the auxiliary task. Best results highlighted in bold. See text for a detailed explanation of the training protocol used in each row.}
	\label{tab:edgeablation}
	\setlength{\tabcolsep}{2.5pt}
	\scalebox{1}{
	\begin{tabular}{l|ccccccccccc|cc}
		\toprule
		method &\rotatebox{90}{Road} & \rotatebox{90}{Sidewalk} & \rotatebox{90}{Walls} & \rotatebox{90}{Fence} & \rotatebox{90}{Person} & \rotatebox{90}{Poles} & \rotatebox{90}{Vegetation} & \rotatebox{90}{Vehicles} & \rotatebox{90}{Tr. Signs} & \rotatebox{90}{Building}  & \rotatebox{90}{Sky} & \textbf{mIoU} & \textbf{Acc} \\
		\midrule
		\emph{base} AT/DT & 89.95 & 46.77 & 5.16 & 10.21 & 28.93 & 28.92 & 77.50 & 71.37 & \textbf{19.24} & 75.29 & 75.12  & 48.04 & 85.90 \\
		Separate ($N_1$ + edge), $N_2$ & 87.24 & 43.30 & 3.08 & 10.17 & 41.77 & 29.04 & 81.81 & 72.35 & 16.58 & 77.10 & 73.10 & 48.69 & 85.89 \\
		Separate ($N_1$ + edge), ($N_2$ + edge) & 88.83 & 47.31 & \textbf{7.10} & 8.59 & \textbf{44.53} & 30.99 & \textbf{83.24} & \textbf{73.54} & 18.05 & \textbf{78.10} & 69.66 & 49.99 & 86.72 \\
		Simultaneous ($N_1$ + $N_2$ + edge) & \textbf{90.12} & \textbf{48.90} & 4.18 & \textbf{11.63} & 37.40 & \textbf{31.98} & 82.34 & 71.50 & 15.11 & 78.04 & \textbf{80.61} & \textbf{50.16} & \textbf{87.21} \\
		\bottomrule
	\end{tabular}
	}
\end{table*}

\begin{table*}[t]
	\center
	\caption{Comparison between NDA loss and other strategies to align $E_1$ features. Best results highlighted in bold.}
	\label{tab:atdtn1align}
	\setlength{\tabcolsep}{2.5pt}
	\scalebox{1}{
	\begin{tabular}{l|ccccccccccc|cc}
		\toprule
		$E_1$ Align. &\rotatebox{90}{Road} & \rotatebox{90}{Sidewalk} & \rotatebox{90}{Walls} & \rotatebox{90}{Fence} & \rotatebox{90}{Person} & \rotatebox{90}{Poles} & \rotatebox{90}{Vegetation} & \rotatebox{90}{Vehicles} & \rotatebox{90}{Tr. Signs} & \rotatebox{90}{Building}  & \rotatebox{90}{Sky} & \textbf{mIoU} & \textbf{Acc} \\
		\midrule 
	    None & 89.95 & 46.77 & 5.16 & 10.21 & 28.93 & 28.92 & 77.50 & 71.37 & 19.24 & 75.29 & 75.12  & 48.04 & 85.90 \\
	    Adv. & 89.89 & 46.01 & 4.22 & 11.89 & \textbf{38.20} & 30.65 & 77.00 & 63.68 & 12.99 & 74.35 & \textbf{81.16} & 48.19 & 85.42 \\
	    
	    LargerNorm \cite{xu2019larger} (1) & 38.37 & 24.17 & 0.56 & 3.66 & 10.50 & 23.04 & 52.61 & 9.41 & 3.42 & 52.64 & 10.54 & 20.81 & 51.49 \\
	    
        LargerNorm \cite{xu2019larger} (25) & 86.82 & 42.23 & 1.94 & 9.00 & 34.92 & 29.02 & 76.39 & 70.97 & 23.38 & 74.97 & 80.00 & 48.15 & 84.62 \\
         LargerNorm \cite{xu2019larger} (500) & 78.94 & 31.25 & 2.53 & 6.00 & 22.08 & 20.55 & 68.18 & 26.21 & 4.35 & 62.28 & 63.53 & 35.08 & 76.53 \\
        
        Asymmetric Adv. \cite{yang2020mind} & 86.69 & 38.57 & \textbf{5.92} & 5.72 & 27.43 &  22.91 & 70.81 & 70.71 & 7.86 & 72.15 & 75.18 & 44.00 & 83.38 \\
		NDA  & \textbf{91.21} & \textbf{50.16} & 5.14 & \textbf{13.78} & 36.99 & \textbf{32.10} & \textbf{77.72} & \textbf{73.38} & \textbf{23.47} & \textbf{76.67} & 72.67 & \textbf{50.30} & \textbf{86.77} \\
		\bottomrule
	\end{tabular}
	}
\end{table*}

\subsection{Alignment strategies for $N_1$}

An alternative approach to align $N_1$ features between domains to ease the transfer process and favor the generalization of \trnet{} consists in  leveraging on the widely adopted adversarial training in feature space. In our setting, this can be obtained by adding a critic that must discriminate whether the features produced by $E_1$ come from \domainA{} or \domainB{}. Thus, the encoder $E_1$ not only has to learn a good feature space for its task, but it is also asked to fool the critic. Afterwards, we can proceed to learn a mapping function \trnet{} among tasks as usual. 
In \autoref{tab:atdtn1align} we compare this standard DA methodology to our NDA loss. Adversarial training (second row) does not introduce significant improvements with respect to not performing DA for \taskone{} (\ie, base \atdt{}, first row), while constraining the features extracted by $E_1$ in a norm aligned space (third row) significantly increases both performance metrics with respect to the baseline.
Our intuition is that, although adversarial training can be useful for domain alignment, it alters the learned feature space with the goal of fooling the critic and this training objective can lead to worse performances on the current task.
Our NDA loss, on the other hand, acts as a regularizer that favors the learning of an homogeneous latent space across the domains involved in our experiments, improving the generalization capability of the transfer network without degrading the performances in the single tasks.
\rev{Then, from the third to the fifth row, we compare our NDA loss with another strategy,  LargerNorm \cite{xu2019larger}, that also align features across domains operating on the feature norms. 
They show that features are more transferable across domains if we constrain feature norms to be equal to an arbitrary large number. We notice that the method is very sensible to the norm value, and it could be hard to select without using target labels. When using an appropriate norm value (25, fourth row), the method achieves a slight improvement over the baseline without alignment. However, since it just force all features globally to be a large number, it is not well-suited for tasks in which we have a spatial dimensions such as semantic segmentation.
Moreover, in the sixth row, we experiment also with a more recent adversarial loss formulation, Asymmetric Adv. \cite{yang2020mind}, which preserve discriminability while performing domain alignment by changing only target features instead of both source and target ones. However, we notice that this method is achieving the worst results among feature alignment strategies, even worse than the baseline. Our motivation is that aligning feature distribution in such a high dimensional feature space with a spatial structure might be too difficult to achieve by only changing target features.
Finally, we notice that NDA achieves the best performance probably because it only align features norm rather than the whole marginal distribution, which is an easier goal that can be achieved also in high-dimensional space. Moreover, NDA operates at each spatial location independently rather than globally, exploiting the spatial priors similarity across domains, reaching better performances.
}

\subsection{Aligning $N_2$ features}
\label{sec:n2_align}

\begin{table*}[ht]
    \center
    \caption{Results of aligning output space of $E_2$ in a \depsem{} scenario. Best results highlighted in bold.}
	\label{tab:atdte2align}
	\setlength{\tabcolsep}{2.5pt}
	\scalebox{1}{
	\begin{tabular}{l|ccccccccccc|cc}
		\toprule
		$E_2$ Align. &\rotatebox{90}{Road} & \rotatebox{90}{Sidewalk} & \rotatebox{90}{Walls} & \rotatebox{90}{Fence} & \rotatebox{90}{Person} & \rotatebox{90}{Poles} & \rotatebox{90}{Vegetation} & \rotatebox{90}{Vehicles} & \rotatebox{90}{Tr. Signs} & \rotatebox{90}{Building}  & \rotatebox{90}{Sky} & \textbf{mIoU} & \textbf{Acc} \\
		\midrule 
		None & \textbf{89.95} & \textbf{46.77} & 5.16 & \textbf{10.21} & 28.93 & 28.92 & \textbf{77.50} & 71.37 & 19.24 & \textbf{75.29} & 75.12 & \textbf{48.04} & \textbf{85.90} \\
		Adv. & 89.36 & 46.03 & 5.59 & 8.22 & \textbf{36.45} & 25.44 & 75.15 & \textbf{72.29} & 12.69 & 74.12 & \textbf{75.79} & 47.38 & 85.31 \\
		
		Asymmetric Adversarial \cite{yang2020mind} & 87.90 & 42.81 & \textbf{7.64} & 8.44 & 26.02 & \textbf{29.11} & 72.54 & 69.01 & \textbf{24.01} & 71.71 & 70.42 & 46.33 & 83.61 \\
		NDA & 44.94 & 23.82 & 3.81 & 2.09 & 30.74 & 24.21 & 42.08 & 68.84 & 11.69 & 35.67 & 11.10 & 27.18 & 56.17 \\
		\bottomrule
	\end{tabular}
	}
\end{table*}

\begin{table*}[ht]
    \center
    \caption{\rev{Results of aligning output space of $D_2$ in a \depsem{} scenario. Best results highlighted in bold.}}
	\label{tab:atdtn2align}
	\setlength{\tabcolsep}{2.5pt}
	\scalebox{1}{
	\begin{tabular}{l|ccccccccccc|cc}
		\toprule
		
		$D_2$ Align. &\rotatebox{90}{Road} & \rotatebox{90}{Sidewalk} & \rotatebox{90}{Walls} & \rotatebox{90}{Fence} & \rotatebox{90}{Person} & \rotatebox{90}{Poles} & \rotatebox{90}{Vegetation} & \rotatebox{90}{Vehicles} & \rotatebox{90}{Tr. Signs} & \rotatebox{90}{Building}  & \rotatebox{90}{Sky} & \textbf{mIoU} & \textbf{Acc} \\
		\midrule 
		
		None & \textbf{89.95} & \textbf{46.77} & \textbf{5.16} & \textbf{10.21} & \textbf{28.93} & \textbf{28.92} & \textbf{77.50} & \textbf{71.37} & \textbf{19.24} & \textbf{75.29} & \textbf{75.12} & \textbf{48.04} & \textbf{85.90} \\
		
		Adv. & 87.48 & 45.73 & 0.63 & 2.12 & 26.22 & 26.39 & 61.40 & 66.92 & 12.97 & 66.39 & 74.77 & 42.82 & 81.87 \\
		\bottomrule
	\end{tabular}
	}
\end{table*}

\begin{table*}[!ht]
	\center
	\caption{Results of aligning input and/or output space of \trnet{} in a \depsem{} scenario. Best results highlighted in bold.}
	\label{tab:atdtadversarial}
	\setlength{\tabcolsep}{2.5pt}
	\scalebox{1}{
	\begin{tabular}{ll|ccccccccccc|cc}
		\toprule
		Input Align. & Output Align. &\rotatebox{90}{Road} & \rotatebox{90}{Sidewalk} & \rotatebox{90}{Walls} & \rotatebox{90}{Fence} & \rotatebox{90}{Person} & \rotatebox{90}{Poles} & \rotatebox{90}{Vegetation} & \rotatebox{90}{Vehicles} & \rotatebox{90}{Tr. Signs} & \rotatebox{90}{Building}  & \rotatebox{90}{Sky} & \textbf{mIoU} & \textbf{Acc} \\
		\midrule 
		- & NDA & 42.97 & 19.60 & 2.31 & 1.36 & 4.21 & 15.74 & 18.42 & 11.77 & 7.19 & 36.72 & 38.99 & 18.12 & 43.63\\
		- & Adv & 90.80 & 48.91 & \textbf{6.16} & 11.84 & 35.32 & 30.29 & \textbf{78.78} & 71.17 & 18.51 & 75.66 & 75.03  & 49.32 & 86.43 \\
	    
	    & Asymmetric Adv. \cite{yang2020mind}  & 85.49 & 40.70 & 4.94 & 10.49 &  34.02 & 30.26 & 76.31 & 70.30 & 17.07 & 74.30 & 72.94 & 46.99 & 83.86 \\
		- & NDA + Adv & 91.03 & 48.93 & 6.14 & 12.24 & 35.91 & 31.05 & 77.93 & 70.28 & 16.65 & 75.50 & 74.47  & 49.10  & 86.28 \\
		
		- & Adv D2 & 90.20 & 47.54 & 5.92 & 11.76 & \textbf{37.03} & 29.52 & 77.98 & 72.42 & 19.28 & 75.82 & \textbf{77.03} & 49.50 & 86.28 \\
		NDA & Adv & 90.67 & 49.49 & 5.54 & 12.29  & 36.73 & 28.49 & 78.28 & 70.19 & 22.05 & 76.47 & 76.35 & 49.69  & 86.73 \\
		NDA & - & \textbf{91.21} & \textbf{50.16} & 5.14 & \textbf{13.78} & 36.99 & \textbf{32.10} & 77.72 & \textbf{73.38} & \textbf{23.47} & \textbf{76.67} & 72.67 & \textbf{50.30} & \textbf{86.77} \\
		
		\bottomrule
	\end{tabular}
	}
\end{table*}

We tried to perform feature alignment across domains also on the features $f_2$ extracted by $E_2$, either by deploying adversarial training or imposing our NDA loss. The idea is to favor the generalization of \trnet{} by making more homogeneous not only its input space (\ie, the features produced by $E_1$, aligned with our NDA loss), but also its output space, \ie, the features produced by $E_2$. However, the setting is not completely symmetric: when learning $E_2$, we do not have supervision available for \domainB{}, and the only loss shaping the feature space for its images would be the alignment loss. We report results of this ablation study in \autoref{tab:atdte2align} and discuss them below.

\rev{
In the first row, we report the results provided by the \emph{base} \atdt{} (without $L_{NDA}$ and $L_{aux}$). In the following two rows, we show results obtained by an adversarial (row 2) and an asymmetric adversarial \cite{yang2020mind} (row 3) training on the features $f_2$, using the same procedures described in the previous sub-section for $f_1$. We can observe that, not only both adversarial trainings does not improve  (like  adversarial training applied to $E_1$), but they even decrease the overall mIoU compared to the baseline.
Finally, in the fourth row, we report the results obtained by our NDA loss on $f_2$: the NDA loss destroys the feature space of \tasktwo{} when applied in this context, as vouched by the drop of $20\%$ in the overall mIoU wrt to base \atdt{}.

During \atdt{} inference, we use also $D_2$  to yield the final task predictions. Nevertheless, $D_2$ has been trained only on \domainA{}, thus its performance may be harmed when using \domainB{} images.
Thus, we ran an additional test reported in \autoref{tab:atdtn2align}. Following \cite{Tsai_2018} we train $N_2$ (i.e., $E_2$ and $D_2$) using an adversarial loss on the $D_2$ output space, thus making $D_2$ aware of \domainB{}. 
Then, we train \trnet{} to map features of $E_1$ into features of $E_2$, and during inference we employ the previously trained decoder $D_2$ to produce the final outputs reporting the results in row \textit{Adv.}. 
We notice a clear drop in performance w.r.t. \textit{base} \atdt{} (row \textit{None}), i.e. \atdt{} trained without $L_{NDA}$ and $L_{aux}$. 
}

We formulate the following hypothesis to explain the above results: all adversarial trainings and NDA loss try to align $f_2^\mathcal{A}$ and $f_2^\mathcal{B}$. While $f_2^\mathcal{A}$ are shaped also by the supervision of \tasktwo, $f_2^\mathcal{B}$ evolve only according to the additional loss we impose, as we do not have supervision for \tasktwo{} on \domainB{}. However, $E_2$ is shared across domains, and therefore may be pushed to produce worse representations for both domains while it tries to accomplish the adversarial objectives or the NDA loss minimization for \domainB{}. If this happens, mappings learned by \trnet{} from $f_1^\mathcal{A}$ to $f_2^\mathcal{A}$ will hallucinate worse features for \tasktwo{} on \domainB{}.
To understand why adversarial trainings leads to small decreases in performances compared to the use of NDA loss, we ought to consider that adversarial training implies a discriminator that cannot be easily fooled by totally degenerated features, while, without any additional constrain from task supervision, the NDA loss may yield  totally collapsed representation.

\subsection{Aligning \trnet{} features}
Although feature alignment does not turn out beneficial when training $N_2$, one may still expect to obtain better hallucinated features if the representations obtained when transferring $f_1^\mathcal{A}$ and $f_1^\mathcal{B}$ are aligned.
We empirically found out that even though output space aligning strategies deployed when training \trnet{} can lead to improvements in performance, input space alignment using our NDA loss deployed when training $N_1$  is more effective.
Moreover, combining input and output space alignment techniques does not lead to further improvements.
We performed this ablation study in the \depsem{} scenario using Carla as \domainA{} and Cityscapes as \domainB{}. The results of these experiments are reported in \autoref{tab:atdtadversarial}.

First, we applied our NDA loss to the output-space of \trnet{}. Similarly to what discussed in the previous section, we notice that, without supervision on \domainB{}, the representations transformed from \trnet{} while minimizing the NDA loss yield a drastic drop in the framework performance (row 1).
We also tried to align the output space features by training \trnet{} alongside a discriminator in an adversarial fashion. We wanted to fool the discriminator in order to generate indistinguishable features from \domainA{} or \domainB{}. We notice that this strategy allows us to reach good overall performances with a 49.32 mIoU on \cityscapes{} (second row).
Moreover, we thought that, as adversarial training provides a supervision on \domainB{}, using the NDA loss in combination with the adversarial loss could avoid producing degenerated features for \domainB{} while reaching a better overall alignment between \domainA{} and \domainB{}. However, we notice that the combination of the two losses leads us to slightly worse results than  adversarial training alone (rows 2 vs 3). Furthermore, since using an adversarial loss on the output space of \trnet{} lead us to good overall performances, we tested it in combination with the best input space alignment from \autoref{tab:atdtn1align}, \ie NDA loss applied when training $N_1$. However, the combination of these two methods achieves worse performance than using only the NDA loss on input space (rows 6 vs 7).
\rev{Finally, we also experimented a different alignment strategy for the \trnet{} output space. Instead of directly applying adversarial loss in $E_2$ feature space, we apply adversarial loss in $D_2$ output space while training \trnet{}. 
As discussed in \cite{Tsai_2018}, output space is easier to align than feature space for several reasons: i) the scene semantic structure is typically similar across domains ii) the feature space encode many information such as color, light, textures iii) the feature space has higher dimensions. By aligning $D_2$ output space we indirectly influence also $E_2$ features making them more domain aligned.
During training, we keep $D_2$ frozen and we update only \trnet{} weights. 
Also in this case, if compare this methodology with simply using $L_{NDA}$ alone (row 6 vs row 7), it achieves worse results.}

\section{Concluding Remarks}
We have introduced a framework to transfer knowledge between different tasks by learning an explicit mapping function between deep features.
This mapping function can be parametrized by a neural network and show interesting generalization capabilities across domains.
To further ameliorate performance we have proposed two novel feature alignment strategies.
At a domain level, we showed that the transfer function presented in our framework can be boosted by making its input space more homogeneous across domains with our simple yet effective NDA loss. At a task level, instead, we reported how deep features extracted for different tasks can be enriched and aligned with the introduction of a shared auxiliary task, which we implemented as edge detection in our experiments.
We reported good results in the challenging synthetic to real scenario while transferring knowledge between the semantic segmentation and monocular depth estimation tasks.

Our proposal is complementary to the whole domain adaptation literature and might be integrated with it.
While DA directly applied to the learned feature space does not seems effective (see \autoref{tab:atdte2align}) more modern techniques either try to align the prediction in the final label space \cite{Tsai_2018} or rely on self-ensembling for pseudo labeling \cite{choi2019self}.
We plan to incorporate these promising direction into our framework as part of future developments.

\ifCLASSOPTIONcaptionsoff
  \newpage
\fi



\bibliographystyle{IEEEtran}
\bibliography{egbib}

\begin{thebibliography}{10}
\providecommand{\url}[1]{#1}
\csname url@samestyle\endcsname
\providecommand{\newblock}{\relax}
\providecommand{\bibinfo}[2]{#2}
\providecommand{\BIBentrySTDinterwordspacing}{\spaceskip=0pt\relax}
\providecommand{\BIBentryALTinterwordstretchfactor}{4}
\providecommand{\BIBentryALTinterwordspacing}{\spaceskip=\fontdimen2\font plus
\BIBentryALTinterwordstretchfactor\fontdimen3\font minus
  \fontdimen4\font\relax}
\providecommand{\BIBforeignlanguage}[2]{{%
\expandafter\ifx\csname l@#1\endcsname\relax
\typeout{** WARNING: IEEEtran.bst: No hyphenation pattern has been}%
\typeout{** loaded for the language `#1'. Using the pattern for}%
\typeout{** the default language instead.}%
\else
\language=\csname l@#1\endcsname
\fi
#2}}
\providecommand{\BIBdecl}{\relax}
\BIBdecl

\bibitem{zamir2018taskonomy}
A.~R. Zamir, A.~Sax, W.~Shen, L.~J. Guibas, J.~Malik, and S.~Savarese,
  ``Taskonomy: Disentangling task transfer learning,'' in \emph{Proceedings of
  the IEEE Conference on Computer Vision and Pattern Recognition}, 2018, pp.
  3712--3722.

\bibitem{zamir2020robust}
A.~R. Zamir, A.~Sax, N.~Cheerla, R.~Suri, Z.~Cao, J.~Malik, and L.~J. Guibas,
  ``Robust learning through cross-task consistency,'' in \emph{Proceedings of
  the IEEE/CVF Conference on Computer Vision and Pattern Recognition}, 2020,
  pp. 11\,197--11\,206.

\bibitem{Wang_2018}
\BIBentryALTinterwordspacing
M.~Wang and W.~Deng, ``Deep visual domain adaptation: A survey,''
  \emph{Neurocomputing}, vol. 312, p. 135–153, Oct 2018. [Online]. Available:
  \url{http://dx.doi.org/10.1016/j.neucom.2018.05.083}
\BIBentrySTDinterwordspacing

\bibitem{ramirez2019learning}
P.~Z. Ramirez, A.~Tonioni, S.~Salti, and L.~D. Stefano, ``Learning across tasks
  and domains,'' in \emph{Proceedings of the IEEE International Conference on
  Computer Vision}, 2019, pp. 8110--8119.

\bibitem{ramirez2018exploiting}
P.~Z. Ramirez, A.~Tonioni, and L.~Di~Stefano, ``Exploiting semantics in
  adversarial training for image-level domain adaptation,'' in \emph{2018 IEEE
  International Conference on Image Processing, Applications and Systems
  (IPAS)}.\hskip 1em plus 0.5em minus 0.4em\relax IEEE, 2018, pp. 49--54.

\bibitem{soria2020dexined}
X.~Soria, E.~Riba, and A.~Sappa, ``Dense extreme inception network: Towards a
  robust cnn model for edge detection,'' in \emph{The IEEE Winter Conference on
  Applications of Computer Vision (WACV '20)}, 2020.

\bibitem{HED}
\BIBentryALTinterwordspacing
S.~Xie and Z.~Tu, ``Holistically-nested edge detection,'' \emph{CoRR}, vol.
  abs/1504.06375, 2015. [Online]. Available:
  \url{http://arxiv.org/abs/1504.06375}
\BIBentrySTDinterwordspacing

\bibitem{Wang_2019}
\BIBentryALTinterwordspacing
Y.~Wang, X.~Zhao, Y.~Li, and K.~Huang, ``Deep crisp boundaries: From boundaries
  to higher-level tasks,'' \emph{IEEE Transactions on Image Processing},
  vol.~28, no.~3, p. 1285–1298, Mar 2019. [Online]. Available:
  \url{http://dx.doi.org/10.1109/TIP.2018.2874279}
\BIBentrySTDinterwordspacing

\bibitem{Dosovitskiy17}
A.~Dosovitskiy, G.~Ros, F.~Codevilla, A.~Lopez, and V.~Koltun, ``{CARLA}: {An}
  open urban driving simulator,'' in \emph{Proceedings of the 1st Annual
  Conference on Robot Learning}, 2017, pp. 1--16.

\bibitem{Cordts_2016_CVPR}
M.~Cordts, M.~Omran, S.~Ramos, T.~Rehfeld, M.~Enzweiler, R.~Benenson,
  U.~Franke, S.~Roth, and B.~Schiele, ``The cityscapes dataset for semantic
  urban scene understanding,'' in \emph{The IEEE Conference on Computer Vision
  and Pattern Recognition (CVPR)}, June 2016.

\bibitem{zhuang2019comprehensive}
F.~Zhuang, Z.~Qi, K.~Duan, D.~Xi, Y.~Zhu, H.~Zhu, H.~Xiong, and Q.~He, ``A
  comprehensive survey on transfer learning,'' \emph{arXiv preprint
  arXiv:1911.02685}, 2019.

\bibitem{yosinski2014transferable}
J.~Yosinski, J.~Clune, Y.~Bengio, and H.~Lipson, ``How transferable are
  features in deep neural networks?'' in \emph{Advances in neural information
  processing systems}, 2014, pp. 3320--3328.

\bibitem{deng2009imagenet}
J.~Deng, W.~Dong, R.~Socher, L.-J. Li, K.~Li, and L.~Fei-Fei, ``Imagenet: A
  large-scale hierarchical image database,'' in \emph{2009 IEEE conference on
  computer vision and pattern recognition}.\hskip 1em plus 0.5em minus
  0.4em\relax Ieee, 2009, pp. 248--255.

\bibitem{redmon2016you}
J.~Redmon, S.~Divvala, R.~Girshick, and A.~Farhadi, ``You only look once:
  Unified, real-time object detection,'' in \emph{Proceedings of the IEEE
  conference on computer vision and pattern recognition}, 2016, pp. 779--788.

\bibitem{Ren_2017}
\BIBentryALTinterwordspacing
S.~Ren, K.~He, R.~Girshick, and J.~Sun, ``Faster r-cnn: Towards real-time
  object detection with region proposal networks,'' \emph{IEEE Transactions on
  Pattern Analysis and Machine Intelligence}, vol.~39, no.~6, p. 1137–1149,
  Jun 2017. [Online]. Available:
  \url{http://dx.doi.org/10.1109/TPAMI.2016.2577031}
\BIBentrySTDinterwordspacing

\bibitem{He_2017}
\BIBentryALTinterwordspacing
K.~He, G.~Gkioxari, P.~Dollar, and R.~Girshick, ``Mask r-cnn,'' \emph{2017 IEEE
  International Conference on Computer Vision (ICCV)}, Oct 2017. [Online].
  Available: \url{http://dx.doi.org/10.1109/ICCV.2017.322}
\BIBentrySTDinterwordspacing

\bibitem{liu2016ssd}
W.~Liu, D.~Anguelov, D.~Erhan, C.~Szegedy, S.~Reed, C.-Y. Fu, and A.~C. Berg,
  ``Ssd: Single shot multibox detector,'' in \emph{European conference on
  computer vision}.\hskip 1em plus 0.5em minus 0.4em\relax Springer, 2016, pp.
  21--37.

\bibitem{Pal_2019_CVPR}
A.~Pal and V.~N. Balasubramanian, ``Zero-shot task transfer,'' in \emph{The
  IEEE Conference on Computer Vision and Pattern Recognition (CVPR)}, June
  2019.

\bibitem{gong2012geodesic}
B.~Gong, Y.~Shi, F.~Sha, and K.~Grauman, ``Geodesic flow kernel for
  unsupervised domain adaptation,'' in \emph{Computer Vision and Pattern
  Recognition (CVPR), 2012 IEEE Conference on}.\hskip 1em plus 0.5em minus
  0.4em\relax IEEE, 2012, pp. 2066--2073.

\bibitem{long15Learning}
M.~Long, Y.~Cao, J.~Wang, and M.~Jordan, ``Learning transferable features with
  deep adaptation networks,'' in \emph{Proceedings of the 32nd International
  Conference on Machine Learning}, ser. Proceedings of Machine Learning
  Research, vol.~37.\hskip 1em plus 0.5em minus 0.4em\relax PMLR, 2015, pp.
  97--105.

\bibitem{ganin2015unsupervised}
Y.~Ganin and V.~Lempitsky, ``Unsupervised domain adaptation by
  backpropagation,'' in \emph{International Conference on Machine Learning},
  2015, pp. 1180--1189.

\bibitem{ganin2016domain}
Y.~Ganin, E.~Ustinova, H.~Ajakan, P.~Germain, H.~Larochelle, F.~Laviolette,
  M.~Marchand, and V.~Lempitsky, ``Domain-adversarial training of neural
  networks,'' \emph{The Journal of Machine Learning Research}, vol.~17, no.~1,
  pp. 2096--2030, 2016.

\bibitem{tzeng2017adversarial}
E.~Tzeng, J.~Hoffman, K.~Saenko, and T.~Darrell, ``Adversarial discriminative
  domain adaptation,'' in \emph{Computer Vision and Pattern Recognition
  (CVPR)}, 2017.

\bibitem{xu2019larger}
R.~Xu, G.~Li, J.~Yang, and L.~Lin, ``Larger norm more transferable: An adaptive
  feature norm approach for unsupervised domain adaptation,'' in
  \emph{Proceedings of the IEEE International Conference on Computer Vision},
  2019, pp. 1426--1435.

\bibitem{goodfellow2014generative}
I.~Goodfellow, J.~Pouget-Abadie, M.~Mirza, B.~Xu, D.~Warde-Farley, S.~Ozair,
  A.~Courville, and Y.~Bengio, ``Generative adversarial nets,'' in
  \emph{Advances in neural information processing systems}, 2014, pp.
  2672--2680.

\bibitem{Zhu_2017_ICCV}
J.-Y. Zhu, T.~Park, P.~Isola, and A.~A. Efros, ``Unpaired image-to-image
  translation using cycle-consistent adversarial networks,'' in \emph{The IEEE
  International Conference on Computer Vision (ICCV)}, Oct 2017.

\bibitem{Bousmalis_2017}
\BIBentryALTinterwordspacing
K.~Bousmalis, N.~Silberman, D.~Dohan, D.~Erhan, and D.~Krishnan, ``Unsupervised
  pixel-level domain adaptation with generative adversarial networks,''
  \emph{2017 IEEE Conference on Computer Vision and Pattern Recognition
  (CVPR)}, Jul 2017. [Online]. Available:
  \url{http://dx.doi.org/10.1109/CVPR.2017.18}
\BIBentrySTDinterwordspacing

\bibitem{Isola_2017_CVPR}
P.~Isola, J.-Y. Zhu, T.~Zhou, and A.~A. Efros, ``Image-to-image translation
  with conditional adversarial networks,'' in \emph{The IEEE Conference on
  Computer Vision and Pattern Recognition (CVPR)}, July 2017.

\bibitem{Tsai_2018}
\BIBentryALTinterwordspacing
Y.-H. Tsai, W.-C. Hung, S.~Schulter, K.~Sohn, M.-H. Yang, and M.~Chandraker,
  ``Learning to adapt structured output space for semantic segmentation,''
  \emph{2018 IEEE/CVF Conference on Computer Vision and Pattern Recognition},
  Jun 2018. [Online]. Available:
  \url{http://dx.doi.org/10.1109/CVPR.2018.00780}
\BIBentrySTDinterwordspacing

\bibitem{hong2018conditional}
W.~Hong, Z.~Wang, M.~Yang, and J.~Yuan, ``Conditional generative adversarial
  network for structured domain adaptation,'' in \emph{Proceedings of the IEEE
  Conference on Computer Vision and Pattern Recognition}, 2018, pp. 1335--1344.

\bibitem{pizzati2020domain}
F.~Pizzati, R.~d. Charette, M.~Zaccaria, and P.~Cerri, ``Domain bridge for
  unpaired image-to-image translation and unsupervised domain adaptation,'' in
  \emph{The IEEE Winter Conference on Applications of Computer Vision}, 2020,
  pp. 2990--2998.

\bibitem{hoffman2016fcns}
J.~Hoffman, D.~Wang, F.~Yu, and T.~Darrell, ``Fcns in the wild: Pixel-level
  adversarial and constraint-based adaptation,'' \emph{arXiv preprint
  arXiv:1612.02649}, 2016.

\bibitem{zhang2017curriculum}
Y.~Zhang, P.~David, and B.~Gong, ``Curriculum domain adaptation for semantic
  segmentation of urban scenes,'' in \emph{The IEEE International Conference on
  Computer Vision (ICCV)}, 2017.

\bibitem{chang2019all}
W.-L. Chang, H.-P. Wang, W.-H. Peng, and W.-C. Chiu, ``All about structure:
  Adapting structural information across domains for boosting semantic
  segmentation,'' in \emph{Proceedings of the IEEE Conference on Computer
  Vision and Pattern Recognition}, 2019, pp. 1900--1909.

\bibitem{hoffman2018cycada}
J.~Hoffman, E.~Tzeng, T.~Park, J.-Y. Zhu, P.~Isola, K.~Saenko, A.~Efros, and
  T.~Darrell, ``{C}y{CADA}: Cycle-consistent adversarial domain adaptation,''
  in \emph{Proceedings of the 35th International Conference on Machine
  Learning}, ser. Proceedings of Machine Learning Research, J.~Dy and
  A.~Krause, Eds., vol.~80.\hskip 1em plus 0.5em minus 0.4em\relax
  Stockholmsmässan, Stockholm Sweden: PMLR, 10--15 Jul 2018, pp. 1989--1998.

\bibitem{shrivastava2017learning}
A.~Shrivastava, T.~Pfister, O.~Tuzel, J.~Susskind, W.~Wang, and R.~Webb,
  ``Learning from simulated and unsupervised images through adversarial
  training,'' in \emph{The IEEE Conference on Computer Vision and Pattern
  Recognition (CVPR)}, 2017.

\bibitem{Zhang_2018}
\BIBentryALTinterwordspacing
Y.~Zhang, Z.~Qiu, T.~Yao, D.~Liu, and T.~Mei, ``Fully convolutional adaptation
  networks for semantic segmentation,'' \emph{2018 IEEE/CVF Conference on
  Computer Vision and Pattern Recognition}, Jun 2018. [Online]. Available:
  \url{http://dx.doi.org/10.1109/CVPR.2018.00712}
\BIBentrySTDinterwordspacing

\bibitem{Sankaranarayanan_2018}
\BIBentryALTinterwordspacing
S.~Sankaranarayanan, Y.~Balaji, A.~Jain, S.~N. Lim, and R.~Chellappa,
  ``Learning from synthetic data: Addressing domain shift for semantic
  segmentation,'' \emph{2018 IEEE/CVF Conference on Computer Vision and Pattern
  Recognition}, Jun 2018. [Online]. Available:
  \url{http://dx.doi.org/10.1109/CVPR.2018.00395}
\BIBentrySTDinterwordspacing

\bibitem{pan2020unsupervised}
F.~Pan, I.~Shin, F.~Rameau, S.~Lee, and I.~S. Kweon, ``Unsupervised
  intra-domain adaptation for semantic segmentation through self-supervision,''
  in \emph{Proceedings of the IEEE/CVF Conference on Computer Vision and
  Pattern Recognition}, 2020, pp. 3764--3773.

\bibitem{kim2020learning}
M.~Kim and H.~Byun, ``Learning texture invariant representation for domain
  adaptation of semantic segmentation,'' in \emph{Proceedings of the IEEE/CVF
  Conference on Computer Vision and Pattern Recognition}, 2020, pp.
  12\,975--12\,984.

\bibitem{Tonioni_2017_ICCV}
A.~Tonioni, M.~Poggi, S.~Mattoccia, and L.~Di~Stefano, ``Unsupervised
  adaptation for deep stereo,'' in \emph{The IEEE International Conference on
  Computer Vision (ICCV)}, Oct 2017.

\bibitem{Zheng_2018_ECCV}
C.~Zheng, T.-J. Cham, and J.~Cai, ``T2net: Synthetic-to-realistic translation
  for solving single-image depth estimation tasks,'' in \emph{The European
  Conference on Computer Vision (ECCV)}, September 2018.

\bibitem{Tonioni_2019_CVPR}
A.~Tonioni, F.~Tosi, M.~Poggi, S.~Mattoccia, and L.~D. Stefano, ``Real-time
  self-adaptive deep stereo,'' in \emph{The IEEE Conference on Computer Vision
  and Pattern Recognition (CVPR)}, June 2019.

\bibitem{lee2018spigan}
\BIBentryALTinterwordspacing
K.-H. Lee, G.~Ros, J.~Li, and A.~Gaidon, ``{SPIGAN}: Privileged adversarial
  learning from simulation,'' in \emph{International Conference on Learning
  Representations}, 2019. [Online]. Available:
  \url{https://openreview.net/forum?id=rkxoNnC5FQ}
\BIBentrySTDinterwordspacing

\bibitem{Kokkinos_2017}
\BIBentryALTinterwordspacing
I.~Kokkinos, ``Ubernet: Training a universal convolutional neural network for
  low-, mid-, and high-level vision using diverse datasets and limited
  memory,'' \emph{2017 IEEE Conference on Computer Vision and Pattern
  Recognition (CVPR)}, Jul 2017. [Online]. Available:
  \url{http://dx.doi.org/10.1109/CVPR.2017.579}
\BIBentrySTDinterwordspacing

\bibitem{ramirez2018geometry}
P.~Z. Ramirez, M.~Poggi, F.~Tosi, S.~Mattoccia, and L.~Di~Stefano, ``Geometry
  meets semantics for semi-supervised monocular depth estimation,'' in
  \emph{Asian Conference on Computer Vision}.\hskip 1em plus 0.5em minus
  0.4em\relax Springer, 2018, pp. 298--313.

\bibitem{Cipolla_2018}
\BIBentryALTinterwordspacing
R.~Cipolla, Y.~Gal, and A.~Kendall, ``Multi-task learning using uncertainty to
  weigh losses for scene geometry and semantics,'' \emph{2018 IEEE/CVF
  Conference on Computer Vision and Pattern Recognition}, 2018. [Online].
  Available: \url{http://dx.doi.org/10.1109/CVPR.2018.00781}
\BIBentrySTDinterwordspacing

\bibitem{tosi2020distilled}
F.~Tosi, F.~Aleotti, P.~Z. Ramirez, M.~Poggi, S.~Salti, L.~Di~Stefano, and
  S.~Mattoccia, ``Distilled semantics for comprehensive scene understanding
  from videos,'' in \emph{Proceedings of the IEEE Conference on Computer Vision
  and Pattern Recognition}, 2020.

\bibitem{tzeng2015simultaneous}
E.~Tzeng, J.~Hoffman, T.~Darrell, and K.~Saenko, ``Simultaneous deep transfer
  across domains and tasks,'' in \emph{Proceedings of the IEEE International
  Conference on Computer Vision}, 2015, pp. 4068--4076.

\bibitem{kundu2019adapt}
J.~N. Kundu, N.~Lakkakula, and R.~V. Babu, ``Um-adapt: Unsupervised multi-task
  adaptation using adversarial cross-task distillation,'' in \emph{Proceedings
  of the IEEE International Conference on Computer Vision}, 2019, pp.
  1436--1445.

\bibitem{Richter_2017}
\BIBentryALTinterwordspacing
S.~R. Richter, Z.~Hayder, and V.~Koltun, ``Playing for benchmarks,'' in
  \emph{{IEEE} International Conference on Computer Vision, {ICCV} 2017,
  Venice, Italy, October 22-29, 2017}, 2017, pp. 2232--2241. [Online].
  Available: \url{https://doi.org/10.1109/ICCV.2017.243}
\BIBentrySTDinterwordspacing

\bibitem{yu2020bdd100k}
F.~Yu, H.~Chen, X.~Wang, W.~Xian, Y.~Chen, F.~Liu, V.~Madhavan, and T.~Darrell,
  ``Bdd100k: A diverse driving dataset for heterogeneous multitask learning,''
  in \emph{Proceedings of the IEEE/CVF conference on computer vision and
  pattern recognition}, 2020, pp. 2636--2645.

\bibitem{Geiger2013IJRR}
A.~Geiger, P.~Lenz, C.~Stiller, and R.~Urtasun, ``Vision meets robotics: The
  kitti dataset,'' \emph{International Journal of Robotics Research (IJRR)},
  2013.

\bibitem{hirschmuller2005accurate}
H.~Hirschmuller, ``Accurate and efficient stereo processing by semi-global
  matching and mutual information,'' in \emph{Computer Vision and Pattern
  Recognition, 2005. CVPR 2005. IEEE Computer Society Conference on},
  vol.~2.\hskip 1em plus 0.5em minus 0.4em\relax IEEE, 2005, pp. 807--814.

\bibitem{Yu2017}
F.~Yu, V.~Koltun, and T.~Funkhouser, ``Dilated residual networks,'' in
  \emph{Computer Vision and Pattern Recognition (CVPR)}, 2017.

\bibitem{peng2018zero}
K.-C. Peng, Z.~Wu, and J.~Ernst, ``Zero-shot deep domain adaptation,'' in
  \emph{Proceedings of the European Conference on Computer Vision (ECCV)},
  2018, pp. 764--781.

\bibitem{eigen2014depth}
D.~Eigen, C.~Puhrsch, and R.~Fergus, ``Depth map prediction from a single image
  using a multi-scale deep network,'' in \emph{Advances in neural information
  processing systems}, 2014, pp. 2366--2374.

\bibitem{yang2020mind}
J.~Yang, H.~Zou, Y.~Zhou, Z.~Zeng, and L.~Xie, ``Mind the discriminability:
  Asymmetric adversarial domain adaptation,'' in \emph{European Conference on
  Computer Vision}.\hskip 1em plus 0.5em minus 0.4em\relax Springer, 2020, pp.
  589--606.

\bibitem{choi2019self}
J.~Choi, T.~Kim, and C.~Kim, ``Self-ensembling with gan-based data augmentation
  for domain adaptation in semantic segmentation,'' in \emph{Proceedings of the
  IEEE international conference on computer vision}, 2019, pp. 6830--6840.

\end{thebibliography}
%


%
\begin{IEEEbiography}[{\includegraphics[width=1in,height=1.25in,clip,keepaspectratio]{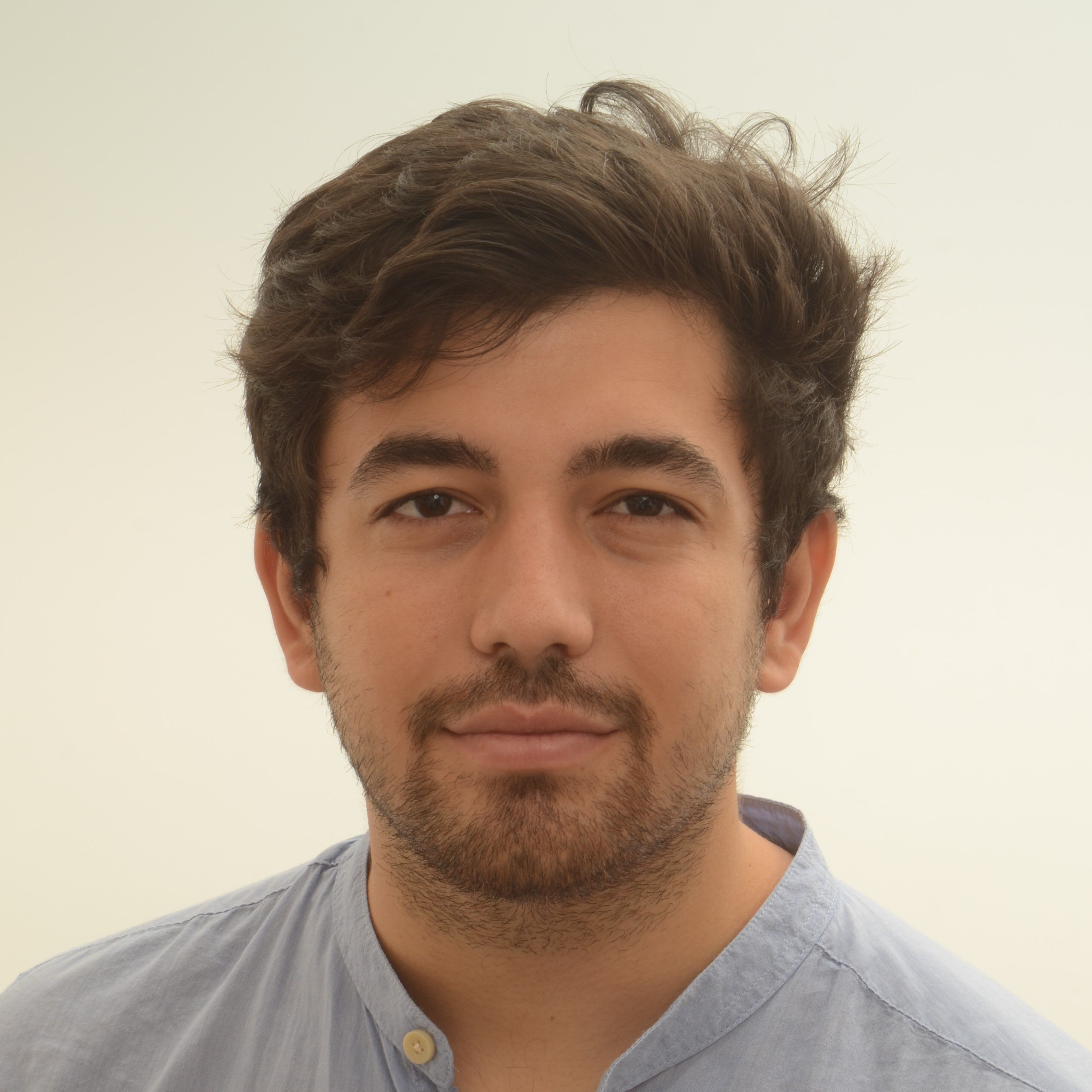}}]{Pierluigi Zama Ramirez} is a Post Doc at the Computer Vision Laboratory (CVLab), University of Bologna. His research interests include deep learning, semantic segmentation, depth estimation, optical flow and domain adaptation. He has authored more than 10 papers on these subjects.
\end{IEEEbiography}

\begin{IEEEbiography}[{\includegraphics[width=1in,height=1.25in,clip,keepaspectratio]{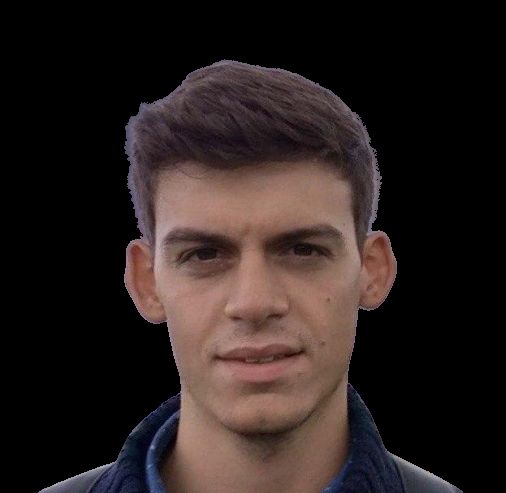}}]{Adriano Cardace} is a PhD student at the Computer Vision Laboratory (CVLab), University of Bologna. His research interests include deep learning for Computer Vision problems, especially semantic segmentation, domain adaptation and self-supervised learning.
\end{IEEEbiography}
\begin{IEEEbiography}[{\includegraphics[width=1in,height=1.25in,clip,keepaspectratio]{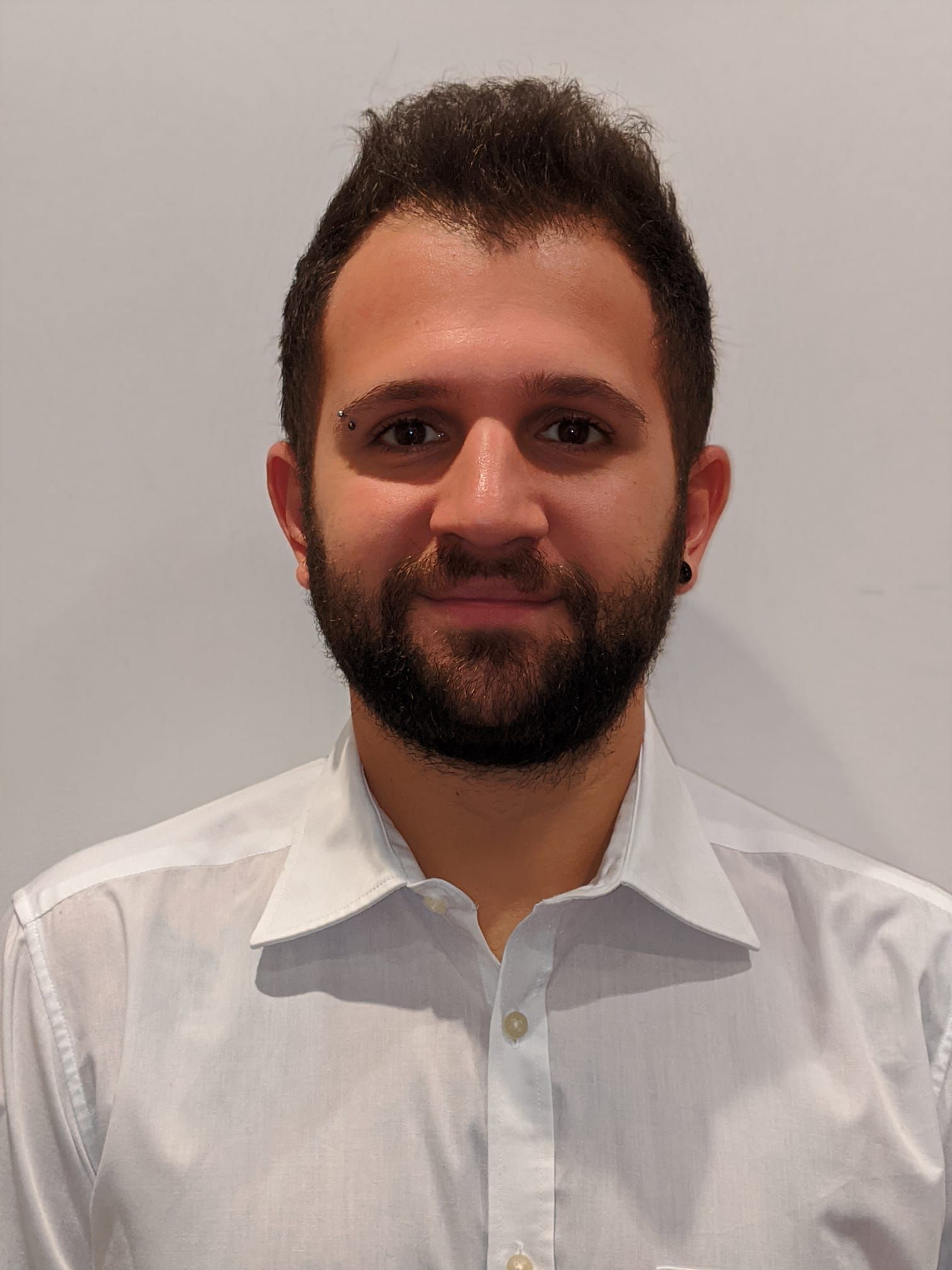}}]{Luca De Luigi} is a PhD student at the Computer Vision Laboratory (CVLab) at the University of Bologna. His research focuses on deep learning for computer vision problems, especially dealing with 3D geometry and implicit neural representations.
\end{IEEEbiography}

\begin{IEEEbiography}
[{\includegraphics[width=1in,height=1.25in,clip,keepaspectratio]{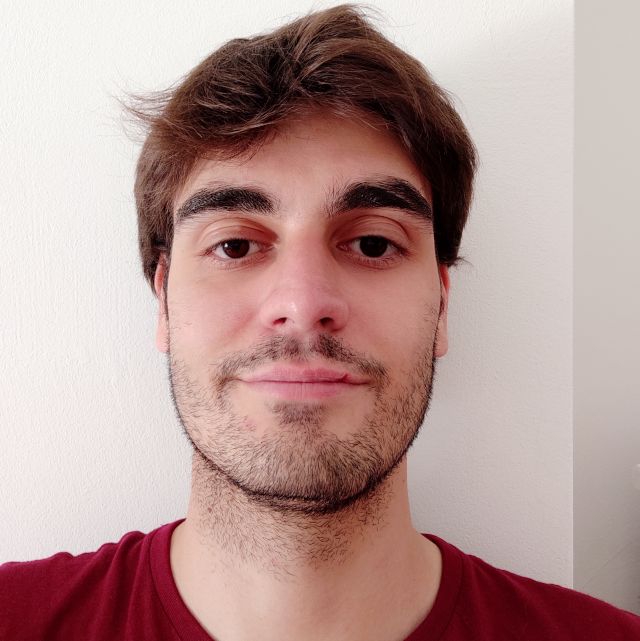}}]
{Alessio Tonioni}
received his PhD degree in Computer Science and Engineering from University of Bologna in 2019. 
Currently, he is a research scientist at Google Zurich.
His research interest concerns machine learning for depth estimation, domain adaptation and generalization. He has authored more than 15 papers on these subjects.
\end{IEEEbiography}

\begin{IEEEbiography}
[{\includegraphics[width=1in,height=1.25in,clip,keepaspectratio]{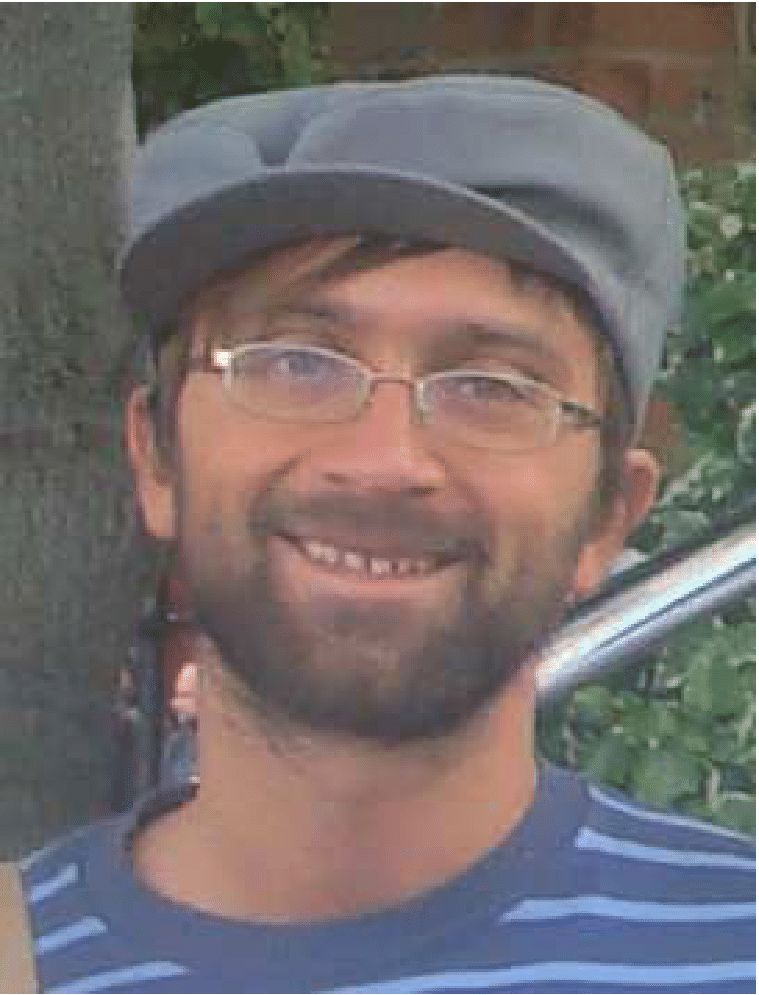}}]
{Samuele Salti} is currently assistant professor at the Department of Computer Science and Engineering (DISI) of the University of Bologna, Italy. Before joining the University of Bologna, he was leading the Data Science team at Verizon Connect, the world leading company in fleet management products and connected vehicles services. His main research interest is computer vision, in particular 3D computer vision, and machine/deep learning applied to computer vision problems. 
Dr. Salti has co-authored 42 publications in international conferences and journals and 8 international patents. In 2020, he co-founded the start-up eyecan.ai. He was awarded the best paper award runner-up at 3DIMPVT 2011, the top international conference on 3D computer vision,
and was nominated outstanding reviewer at CVPR 2020 and NeurIPS 2020. 
\end{IEEEbiography}

\begin{IEEEbiography}[{\includegraphics[width=1in,height=1.25in,clip,keepaspectratio]{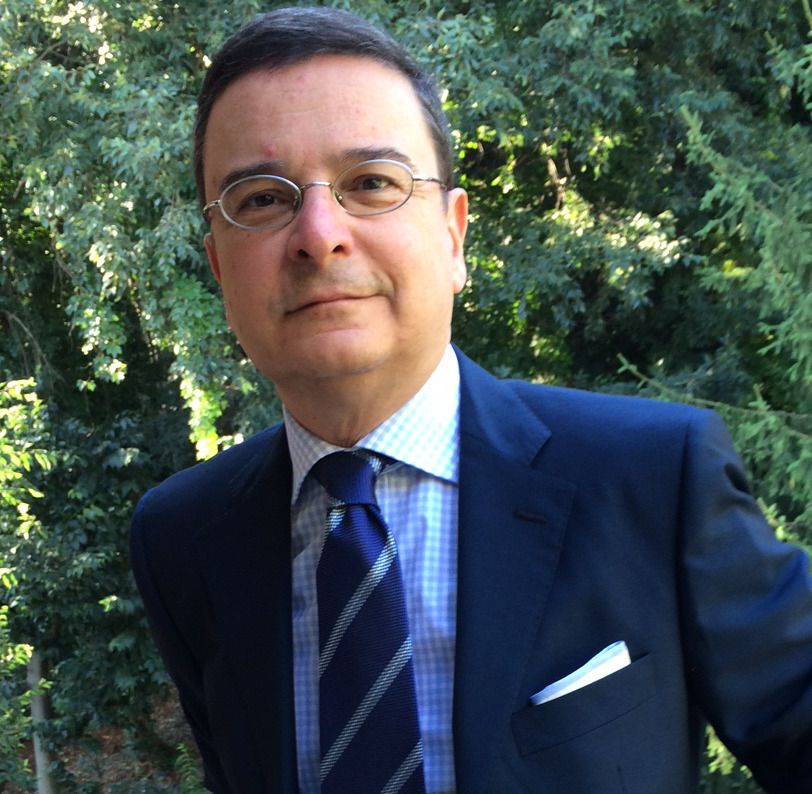}}]{Luigi Di Stefano}
received the PhD degree in
electronic engineering and computer science from
the University of Bologna, in 1994. He is currently
a full professor with the Department of Computer
Science and Engineering, University of Bologna,
where he founded and leads the Computer Vision
Laboratory (CVLab). His research interests
include image processing, computer vision and
machine/deep learning. He is the author of more
than 150 papers and several patents. He has
been scientific consultant for major companies in
the fields of computer vision and machine learning. He is a member of the
IEEE Computer Society, IEEE, and the IAPR-IC.
\end{IEEEbiography}







\end{document}